\documentclass{article}

\usepackage{microtype}
\usepackage{graphicx}
\usepackage{subfigure}
\usepackage{booktabs} 
\usepackage[utf8]{inputenc}
\usepackage{graphicx}
\usepackage{amsmath}
\usepackage{amsthm}
\usepackage{amssymb}
\newtheorem{theorem}{Theorem}
\newtheorem{theorem_sup}{Theorem}

\newtheorem{assumption}{Assumption}

\newtheorem{corollary}{Corollary}
\newtheorem{lemma_sup}{Lemma}
\newtheorem{corollary_sup}{Corollary}
\usepackage{mdframed}
\usepackage{lipsum}
\usepackage{url}

\newmdtheoremenv{theo}{Theorem}
\newsavebox{\savepar}
\newenvironment{bigboxit}{\begin{center}\begin{lrbox}{\savepar}

\begin{minipage}[h]{3.15in}

\normalfont
\begin{flushleft}}
{\end{flushleft}\end{minipage}\end{lrbox}\fbox{\usebox{\savepar}}
\end{center}}
\usepackage{algorithm}

\usepackage{algorithmic}

\usepackage{enumitem}
\setitemize{noitemsep,topsep=0pt,parsep=0pt,partopsep=0pt}
\usepackage{hyperref}
\usepackage{authblk}

\title{Invariant Risk Minimization Games}

\begin{document}
\author{Kartik Ahuja}
\author{Karthikeyan Shanmugam}
\author{Kush R. Varshney}
\author{Amit Dhurandhar}
\affil{IBM Research, Thomas J. Watson Research Center, Yorktown Heights, NY}
\date{}
\maketitle

\begin{abstract}

 The standard risk minimization paradigm of machine learning is brittle when operating in environments whose test distributions are different from the training distribution due to spurious correlations. Training on data from many environments and finding \emph{invariant} predictors reduces the effect of spurious features by concentrating models on features that have a causal relationship with the outcome. In this work, we pose such invariant risk minimization as finding the Nash equilibrium of an ensemble game among several environments. 
By doing so, we develop a simple training algorithm that uses best response dynamics and, in our experiments, yields similar or better empirical accuracy with much lower variance than the challenging bi-level optimization problem of \cite{arjovsky2019invariant}. One key theoretical contribution is showing that the set of Nash equilibria for the proposed game are equivalent to the set of invariant predictors for any finite number of environments, even with nonlinear classifiers and transformations. As a result, our method also retains the generalization guarantees to a large set of environments shown in \cite{arjovsky2019invariant}.  
The proposed algorithm adds to the collection of successful game-theoretic machine learning algorithms such as  generative adversarial networks.
\end{abstract}

\section{Introduction}
\label{sec:intro}

The annals of machine learning are rife with embarrassing examples of spurious correlations that fail to hold outside a  specific training (and identically distributed test) distribution.  In \cite{beery2018recognition} the authors trained a convolutional neural network (CNN) to classify camels from cows. The training dataset had one source of bias, i.e., most of the pictures of cows had green pastures, while most pictures of camels were in deserts. The CNN picked up the spurious correlation, i.e., it associated green pastures with cows and failed to classify pictures of cows on sandy beaches correctly. In another case, a neural network used a brake light indicator to continue applying brakes, which was a spurious correlation in the training data \cite{de2019causal}; the list of such examples goes on. 

To address the problem of models inheriting spurious correlations, the authors in \cite{arjovsky2019invariant} show that one can exploit the varying degrees of spurious correlation naturally present in data collected from multiple data sources to learn robust predictors.  The authors propose to  find a representation $\Phi$ such that the optimal classifier given $\Phi$ is invariant across training environments. This formulation leads to a challenging bi-level optimization, which the authors relax by fixing a simple linear classifier and learning a representation $\Phi$ such that the  classifier is ``approximately locally optimal'' in all the training environments. 

In this work, we take a very different approach. We create an \textit{ensemble} of classifiers with each environment controlling one component of the ensemble. Each environment uses the entire ensemble to make predictions. We let all the environments play a \textit{game} where each environment's action is to decide its contribution to the ensemble such that it minimizes its risk. 
 Remarkably, we establish that the set of predictors that solve the \textit{ensemble game} is equal to the set of invariant predictors across the  training environments; this result holds for a large class of non-linear classifiers.

 This brings us to the question: how do we solve the game?  We use classic best response dynamics \cite{fudenberg1998theory}, which has a very simple implementation. Each environment periodically takes its turn and moves its classifier in the direction that minimizes the risk specific to its environment.  
 Empirically, we establish that the invariant predictors found by our approach  lead to better or comparable performance with much lower standard deviation than \cite{arjovsky2019invariant} on several different datasets. A nice consequence of our approach is we do not restrict classifiers to be linear, which  was emphasized as an important direction for future work by  \cite{arjovsky2019invariant}.

 Broadly speaking, we believe that the game-theoretic perspective herein can open up a totally new paradigm to address the problem of invariance.
 
\section{Related Work}

\subsection{Invariance Principles in Causality}

The invariant risk minimization formulation of \cite{arjovsky2019invariant} is the most related work, and is motivated from the theory of causality and causal Bayesian networks (CBNs) \cite{pearl1995causal}. A variable $y$ is caused by a set of non-spurious actual causal factors $x_{\mathrm{Pa}(y)}$ if and only if in all environments where $y$ has not been intervened on, the conditional probability $P(y|x_{\mathrm{Pa}(y)})$ remains invariant. This is called the \textit{modularity condition} \cite{bareinboim2012local}.
Related and similar notions are the \emph{independent causal mechanism principle} \cite{scholkopf2012causal,janzing2010causal,janzing2012information} and the \emph{invariant causal prediction principle} \cite{peters2016causal,heinze2018invariant}. These principles imply that if all the environments (train and test) are modeled by interventions that do not affect the causal mechanism of target variable $y$, then a classifier conservatively trained on the transformation that involves the causal factors  ($\Phi(x)= x_{\mathrm{Pa}(y)}$) to predict $y$  is robust to unseen interventions.

In general, for finite sets of environments, there may be other invariant predictors. If one has information about the CBN structure, one can find invariant predictors that are maximally predictive  using conditional independence tests and other graph-theoretic tools \cite{magliacane2018domain,subbaswamy2019should}.

The above works 
select subsets of features, primarily using conditional independence tests, that make the optimal classifier trained on the selected features be invariant. 
In \cite{arjovsky2019invariant} the authors give an optimization-based reformulation of this invariance that facilitates searching over transformations in a continuous space, making their work widely applicable in areas such as computer vision where the causal features are latent (see Figure 6 in \cite{arjovsky2019invariant}).

\subsection{Sample Reweighting, Domain Adaptation, and Robust Optimization}

Statistical machine learning  has dealt with the distribution shift between the training distribution and test distribution in a number of ways. Conventional approaches are sample weighting, domain adaptation, and robust optimization. Importance weighting or more generally sample weighting attempts to match test and train distributions by reweighting samples \cite{shimodaira2000improving,sugiyama2008direct,gretton2009covariate,zhao2018metric}. It typically assumes that the probability of labels given all covariates does not shift, and in more general cases, requires access to test labels. Domain adaptation tries to find a representation $\Phi$ whose distribution is invariant across source and target domains \cite{ajakan2014domain,ben2007analysis,glorot2011domain,ganin2016domain}. Domain adaptation is known to have serious limitations even when the marginal distribution of labels shift across environments \cite{zhao2019learning,johansson2019support}. When only training data sources are given, robust optimization techniques find the worst case loss over all possible convex combinations of the training sources \cite{mohri2019agnostic,hoffman2018algorithms,lee2018minimax,duchi2016statistics}. This assumes that the test distribution is within the convex hull of training distributions, which is not true in many settings.

\section{Preliminaries}

\subsection{Game Theory Concepts}
We begin with some basic concepts from game theory \cite{fudenberg1991game} that we will use. Let $\Gamma = (N,\{S_i\}_{i\in N}, \{u_i\}_{i \in N})$ be the tuple representing a standard normal form game, where $N$ is the finite set of players.  Player $i \in N$ takes actions from a strategy set $S_i$. The utility of player $i$ is $u_i:S \rightarrow \mathbb{R}$, where we write the joint set $S = \Pi_{i\in N} S_i$. The joint strategy of all the players is given as $s \in S$,  the  strategy of player $i$ is $s_i$ and the strategy of the rest of players is $s_{-i} = (s_{i^{'}})_{i^{'} \not = i}$.  If the set $S$ is finite, then we call the game $\Gamma$ a \emph{finite game}. 
If the set $S$ is uncountably infinite, then the game $\Gamma$ is a \emph{continuous game}.


\textbf{Nash equilibrium in pure strategies.} A strategy $s^{*}$ is said to be a pure strategy Nash equilibrium (NE) if it satisfies
$$u_i(s_{i}^{*},s_{-i}^{*}) \geq u_i(k,s_{-i}^{*}), \forall k \in S_{i}, \forall i \in N$$

We continue the discussion on other relevant concepts in game theory in the Appendix Section. 
\subsection{Invariant Risk Minimization} 
We describe the invariant risk minimization (IRM) of \cite{arjovsky2019invariant}.
Consider datasets $\{(x_i^e, y_i^e)\}_{i=1}^{n_e}$ from multiple training environments $e\in \mathcal{E}_{tr}$.
The feature value $x_i^{e} \in \mathcal{X}$ and the corresponding labels $y_i^{e}\in \mathcal{Y}$, where $\mathcal{X} \subseteq \mathbb{R}^{n}$ and $\mathcal{Y} \subseteq \mathbb{R}^{k}$.\footnote{The setup applies to both continuous and categorical data. If any feature or label is categorical, we one-hot encode it.} Define a predictor $f:\mathcal{X} \rightarrow \mathbb{R}^{k}$. 

The goal of IRM is to use these multiple datasets to construct a predictor $f$ that performs well across many unseen environments $\mathcal{E}_{all}$. Define the risk achieved by $f$ in environment $e$ as $R^e(f) = \mathbb{E}_{X^e,Y^e}\big[\ell(f(X^e), Y^e)\big]$, where $\ell$ is the loss when $f(X)$ is the predicted value and $Y$ is the corresponding label. To assume that $f$ maps to real values is not restrictive; for instance, in a $k$-class classification problem, the output of the function $f$ is the score for each class, which can be converted into a hard label by selecting the class with the highest score.

\textbf{Invariant predictor:} We say that a data representation $\Phi:\mathcal{X}\rightarrow \mathcal{Z} \subseteq \mathbb{R}^{d}$ elicits an invariant predictor $w\circ \Phi$ across environments $e \in \mathcal{E}$ if there is a classifier $w:\mathcal{Z}\rightarrow \mathbb{R}^{k}$ that achieves the minimum risk for all the environments 
$w \in \arg\min_{\bar{w} \in \mathcal{H}_{w}} R^{e}(\bar{w}\circ \Phi)$. The set of all the mappings $\Phi$ is given as $\mathcal{H}_{\Phi}$ and the set of all the classifiers is given as $\mathcal{H}_{w}$. IRM may be phrased as the following constrained optimization problem \cite{arjovsky2019invariant}:
\begin{equation}
    \begin{split}
        & \min_{\Phi \in \mathcal{H}_{\Phi},w  \in \mathcal{H}_{w}} \sum_{e \in \mathcal{E}_{tr}}R^{e}(w\circ \Phi) \\ 
        & \text{s.t.}\;w \in \arg\min_{\bar{w} \in \mathcal{H}_{w}} R^{e}(\bar{w}\circ \Phi),\;\forall e \in \mathcal{E}_{tr}.
    \end{split}
    \label{eqn: IRM}
\end{equation}
If $(\Phi,w)$ satisfies the above constraints, then $w \circ \Phi$ is an invariant  predictor across the environments $\mathcal{E}_{tr}$.  

Define the set of representations and the corresponding classifiers, $(\Phi, w)$ that satisfy the constraints in the above optimization problem \eqref{eqn: IRM} as $\mathcal{S}^{\mathsf{IV}}$, where $\mathsf{IV}$ stands for invariant. Also, separately 
define the set of invariant predictors $w\circ \Phi$ as  $\hat{\mathcal{S}}^{\mathsf{IV}} = \{w\circ \Phi \; | (\Phi,w) \in \mathcal{S}^{\mathsf{IV}}\}$.

\textbf{Remark.} The sets $\mathcal{S}^{\mathsf{IV}}$, $\hat{\mathcal{S}}^{\mathsf{IV}}$ depend on the choice of classifier class $\mathcal{H}_w$ and representation class $\mathcal{H}_{\Phi}$. We avoid making this dependence explicit until later sections.

Members of $\mathcal{S}^{\mathsf{IV}}$ are equivalently expressed as the solutions to  
\begin{equation}
    R^{e}(w \circ \Phi)   \leq  R^{e}(\bar{w}\circ \Phi),\;\forall \bar{w} \in \mathcal{H}_w, \; \forall e \in \mathcal{E}_{tr}. 
    \label{eqn:IRM1}
\end{equation}
The main result of \cite{arjovsky2019invariant} states that if $\mathcal{H}_w$ and $\mathcal{H}_{\Phi}$ are from the class of linear models, i.e., $w(z) = \mathbf{w}^{t}z$, where $\mathbf{w} \in \mathbb{R}^{d}$, and $\Phi(x) = \mathbf{\Phi} x$ with $\mathbf{\Phi} \in \mathbb{R}^{d\times n}$, then under certain conditions on the data generation process and  training environments $\mathcal{E}_{tr}$, the solution to \eqref{eqn:IRM1} remains invariant in $\mathcal{E}_{all}$.

\section{Ensemble Invariant Risk Minimization Games}

\subsection{Game-Theoretic Reformulation}

Optimization problem \eqref{eqn: IRM} can be quite challenging to solve. We introduce an alternate characterization based on game theory to solve it. 
We endow each environment with its own classifier $w^{e} \in \mathcal{H}_w$. We use a simple ensemble to construct an overall classifier  $w^{av}:\mathcal{Z}\rightarrow \mathbb{R}^{k}$ defined as $w^{av}= \frac{1}{|\mathcal{E}_{tr}|}\sum_{q=1}^{|\mathcal{E}_{tr}|}w^{q}$, where for each $z\in \mathcal{Z}$, $w^{av}(z) = \frac{1}{|\mathcal{E}_{tr}|}\sum_{q=1}^{|\mathcal{E}_{tr}|}w^{q}(z)$.  (The $av$ stands for average.) Consider the example of binary classification with two environments $\{e_1, e_2\}$; $w^{e}= [w^{e}_1, w^{e}_2]$ is the classifier of environment $e$, where each component is the score for each class.  We define the component $j$ of the ensemble classifier $w^{av}$ as $w^{av}_j =\frac{w^{e_1}_j + w^{e_2}_j}{2}$.   These scores are input to a softmax; the final probability assigned to class $j$ for an input $z$ is $\frac{e^{w^{av}_j(z)}}{e^{w^{av}_1(z)} + e^{w^{av}_2(z)}}$.

We  require all the environments to use this ensemble $w^{av}$. We want to solve the following new optimization problem.

\begin{equation*}
    \begin{split}
        & \min_{\Phi \in \mathcal{H}_{\Phi}, w^{av} \in \mathcal{H}_w} \sum_{e \in \mathcal{E}_{tr}}R^{e}(w^{av}\circ \Phi) \\ 
        & \text{s.t.}\;w^e \in \arg\min_{\bar{w}^{e} \in \mathcal{H}_{w}} R^{e}\left(\frac{1}{|\mathcal{E}_{tr}|}\Big[\bar{w}^{e} + \sum_{q\not=e}w^{q}\Big]\circ \Phi\right),\;\forall e \in \mathcal{E}_{tr}
    \end{split}
    \label{eqn: IRM_ensemble}
\end{equation*}

We can equivalently restate the above 
as: 
\begin{bigboxit}
\begin{equation}
    \begin{split}
        & \min_{\Phi \in \mathcal{H}_{\Phi}, w^{av}} \sum_{e \in \mathcal{E}_{tr}}R^{e}(w^{av}\circ \Phi) \\ 
        &\text{s.t.}\; R^{e}\left(\frac{1}{|\mathcal{E}_{tr}|}\Big[w^{e} + \sum_{q\not=e}w^{q}\Big]\circ \Phi\right) \\
        &\leq R^{e}\left(\frac{1}{|\mathcal{E}_{tr}|}\Big[\bar{w}^{e} + \sum_{q\not=e}w^{q}\Big]\circ \Phi\right) \;\forall \bar{w}^{e} \in \mathcal{H}_{w} \;\forall e \in \mathcal{E}_{tr} 
    \end{split}
    \label{eqn: IRM_ensemble_inequality}
\end{equation}
\end{bigboxit}

What are the advantages of this formulation \eqref{eqn: IRM_ensemble_inequality}?
\begin{itemize}
    \item Using the ensemble automatically enforces invariance across environments.
        \item Each environment is free to select the classifier $w^{e}$ from the entire set $\mathcal{H}_w$, unlike in \eqref{eqn: IRM}, where all environments' choices are required to be the same. 
    \item The constraints in \eqref{eqn: IRM_ensemble_inequality} are equivalent to the set of pure NE of a game that we define next.
\end{itemize}

 The game is played between $|\mathcal{E}_{tr}|$ players, with each player corresponding to an environment $e$. The set of actions of the environment $e$ are $w^{e} \in \mathcal{H}_{w}$.  At the start of the game, a representation $\Phi$ is selected from the set $\mathcal{H}_{\Phi}$, which is observed by all the environments.  The utility function for an environment $e$ is defined as  $u_{e}[w^{e}, w^{-e}, \Phi] = - R^{e}(w^{av}, \Phi)$, where $w^{-e} = \{w^{q}\}_{q\not=e}$ is the set of choices of all environments but $e$. We call this game Ensemble Invariant Risk Minimization (EIRM) and express it as a tuple 

$$\Gamma^{\mathsf{EIRM}} = \Big(\mathcal{E}_{tr}, \mathcal{H}_{\Phi}, \{\mathcal{H}_{w}\}_{q=1}^{|\mathcal{E}_{tr}|}, \{u_{e}\}_{e\in \mathcal{E}_{tr}} \Big).$$
We represent a pure NE as a tuple $\Big(\Phi,\{ w^{q}\}_{q=1}^{|\mathcal{E}_{tr}|}\Big)$. Since each pure NE depends on $\Phi$, we include it as a part of the tuple.\footnote{We can also express each environment's action as a mapping from $\pi:\mathcal{H}_{\Phi} \rightarrow \mathcal{H}_{w}$ but we don't to avoid complicated notation.}
We define the set of pure NE as $\mathcal{S}^{\mathsf{EIRM}}$. We construct a set of all the ensemble predictors constructed from NE as\footnote{We don't double count  compositions leading to the same predictor.}    $$\hat{\mathcal{S}}^{\mathsf{EIRM}} =  \Big\{\Big[\frac{1}{|\mathcal{E}_{tr}|} \sum_{q=1}^{|\mathcal{E}_{tr}|} w^q \Big] \circ \Phi \; |\; (\Phi, \{w^{q}\}_{q=1}^{|\mathcal{E}_{t}|} ) \in \mathcal{S}^{\mathsf{EIRM}} \Big\}.$$  

Members of $ \mathcal{S}^{\mathsf{EIRM}} $ are equivalently expressed as the solutions to 
\begin{equation}
    \begin{split}
        u_{e}[w^{e}, w^{-e}, \Phi] \geq  u_{e}[\bar{w}^{e}, w^{-e}, \Phi], \;\forall w^{e} \in \mathcal{H}_{w}, \forall e \in \mathcal{E}_{tr}. 
    \end{split}
    \label{eqn:eirm_ne_ineq}
\end{equation} 
If we replace $u_{e}[w^{e}, w^{-e}, \Phi]$ with $- R^{e}(w^{av}, \Phi)$, we obtain the inequalities in \eqref{eqn: IRM_ensemble_inequality}. So far we have defined the game and given its relationship to the problem in \eqref{eqn: IRM_ensemble_inequality}.

\subsection{Equivalence Between NE and Invariant Predictors}

\emph{What is the relationship between the predictors obtained from NE  $\hat{\mathcal{S}}^{\mathsf{EIRM}}$ and invariant predictors $\hat{\mathcal{S}}^{\mathsf{IV}}$?}

Remarkably, these two sets are the same under very mild conditions. Before we show this result, we  establish a stronger result and this result will follow from it.

We use the set $\mathcal{S}^{\mathsf{EIRM}}$ to construct a new set. To each tuple $\left(\Phi, \{w^{q}\}_{q=1}^{|\mathcal{E}_{tr}|})\right) \in \mathcal{S}^{\mathsf{EIRM}}$ augment the ensemble classifier $w^{av} = \frac{1}{|\mathcal{E}_{tr}|}\sum_{q=1}^{|\mathcal{E}_{tr}|}w^{q}$ to get $\Big(\Phi, \{w^{q}\}_{q=1}^{|\mathcal{E}_{tr}|}, w^{av} \Big) $. We call the set of these new tuples $\tilde{\mathcal{S}}^{\mathsf{EIRM}}$.

We use the set $\mathcal{S}^{\mathsf{IV}}$ to construct a new set. Consider an element $(\Phi, w) \in \mathcal{S}^{\mathsf{IV}}$. We define a decomposition for $w$ in terms of the environment-specific classifiers as follows: $w = \frac{1}{|\mathcal{E}_{tr}|}\sum_{q=1}^{|\mathcal{E}_{tr}|}w^{q}$, where $w^{q} \in \mathcal{H}_w$. 
 $w^q = w, \forall q \in \mathcal{E}_{tr}$ is one trivial decomposition. We use each such decomposition and augment the tuple to obtain $\Big(\Phi, \{w^{q}\}_{q=1}^{|\mathcal{E}_{tr}|}, w \Big) $. We call this set of new tuples  $\tilde{\mathcal{S}}^{\mathsf{IV}}$.

Both the sets $\tilde{\mathcal{S}}^{\mathsf{IV}}$ and $\tilde{\mathcal{S}}^{\mathsf{EIRM}}$ consist of tuples of representation, set of environment specific classifiers, and the ensemble classifier. We ask an even more interesting question than the one above. Is the set of representations, environment specific classifiers, and the ensembles found by playing EIRM \eqref{eqn:eirm_ne_ineq} or solving IRM \eqref{eqn:IRM1} the same? If these two sets are equal, then equality between $\hat{\mathcal{S}}^{\mathsf{EIRM}}$   and $\hat{\mathcal{S}}^{\mathsf{IV}}$ follows trivially.

We state the only assumption we need. 
\begin{assumption}
\label{ass:affine closure}
\textbf{Affine closure:} The class of functions $\mathcal{H}_w$ is closed under the following operations.
\begin{itemize}
    \item Finite sum: If $w_1 \in \mathcal{H}_w$ and $w_2 \in \mathcal{H}_w$, then $w_1 + w_2 \in \mathcal{H}_w$, where for every $z\in \mathcal{Z}$, $(w_1 + w_2)(z) = w_1(z) + w_2(z) $
    \item Scalar multiplication: For any $c\in \mathbb{R}$ and $w \in \mathcal{H}_w$, $c w \in \mathcal{H}_w$, where for every $z\in \mathcal{Z}$, $(c w)(z) = c\times w(z) $
\end{itemize}
\end{assumption}
The addition of the functions and scalar multiplication are defined in a standard pointwise manner. Therefore, the class $\mathcal{H}_w$ also forms a vector space.

\textbf{Examples of functions that satisfy affine closure.} Linear classifiers, kernel based classifiers \cite{hofmann2008kernel} (functions in RKHS space), ensemble models with arbitrary number of weak learners \cite{freund1999short},  functions in $L^{p}$ space \cite{ash2000probability}, ReLU networks with arbitrary depth. We provide the justification for each of these functions in the Appendix Section. 
We now state the main result.

\begin{theorem}
\label{theorem1}
If Assumption \ref{ass:affine closure} holds, then
$\tilde{\mathcal{S}}^{\mathsf{IV}} = \tilde{\mathcal{S}}^{\mathsf{EIRM}}$
\end{theorem}

The proofs of all the results are in the Appendix Section.

\begin{corollary} 
\label{corollary1}
If Assumption \ref{ass:affine closure} holds, then $\hat{\mathcal{S}}^{\mathsf{IV}} = \hat{\mathcal{S}}^{\mathsf{EIRM}}$
\end{corollary}

\textbf{Significance of Theorem \ref{theorem1} and Corollary \ref{corollary1}}
\begin{itemize}

    \item From a computational standpoint, this equivalence permits tools from game theory to find NE of the EIRM game and, as a result, the invariant predictors. 
        \item From a theoretical standpoint, this equivalence permits to use game theory to analyze the solutions of the EIRM game and understand the invariant predictors.
    \item In Theorem 9 of \cite{arjovsky2019invariant}, it was shown for linear classifiers and linear representations that the invariant predictors generalize to a large set of unseen environments under certain conditions. Since our result holds for linear classifiers (but is even broader), the generalization result continues to hold for the predictors found by playing the EIRM game.

\end{itemize}

\textbf{Role of representation $\Phi$.} We investigate the  scenario when we fix $\Phi$ to the identity mapping; this will motivate one of our approaches. Define the set $\hat{\mathcal{S}}^{\mathsf{EIRM}}(\Phi)$ as the set of ensemble predictors arrived at by playing the EIRM game using a fixed representation representation $\Phi$.\footnote{$\cup_{\Phi}\hat{\mathcal{S}}^{\mathsf{EIRM}}(\Phi) = \hat{\mathcal{S}}^{\mathsf{EIRM}}$}  Similarly, we define a set $\hat{\mathcal{S}}^{\mathsf{IV}}(\Phi)$ as the set of invariant predictors derived using the representation $\Phi$. From Theorem \ref{theorem1}, it follows that $\hat{\mathcal{S}}^{\mathsf{EIRM}}(\Phi) = \hat{\mathcal{S}}^{\mathsf{IV}}(\Phi)$. We modify some of the earlier notations for results to follow.  The set of predictors that result from the EIRM game $\hat{\mathcal{S}}^{\mathsf{EIRM}}$ and the sets of  invariant predictors $\hat{\mathcal{S}}^{\mathsf{IV}}$ are defined for a family of maps $\Phi$ with co-domain $\mathcal{Z}$. We make the co-domain  $\mathcal{Z}$ explicit in the notation. We write  $\hat{\mathcal{S}}_{\mathcal{Z}}^{\mathsf{EIRM}}$ for $\hat{\mathcal{S}}^{\mathsf{EIRM}}$ and  $\hat{\mathcal{S}}_{\mathcal{Z}}^{\mathsf{IV}}$ for $\hat{\mathcal{S}}^{\mathsf{IV}}$.

\begin{assumption}\label{assum:phi}
$\Phi \in \mathcal{H}_{\Phi}$ satisfies the following
\begin{itemize}
    \item Bijective:  $\exists$ $\Phi^{-1}:\mathcal{Z} \rightarrow \mathcal{X}$ such that\;  $\forall x \in \mathcal{X}$, $\Big(\Phi^{-1} \; \circ  \;\Phi \Big)(x) = x$, and $\forall z \in \mathcal{Z}$ $\Big(\Phi\; \circ\; \Phi^{-1}\Big) (z) = z$.   
Both $\mathcal{X}$ and $\mathcal{Z}$ are subsets of $\mathbb{R}^{n}$
    \item $\Phi $ is differentiable and Lipschitz continuous.

\end{itemize}
\end{assumption}

We define $L^p(\mathcal{Z})$ as the set of functions $f:\mathcal{Z} \rightarrow \mathbb{R}$ s.t. $\int_{\mathcal{Z}} |f|^p d\mu <\infty$ 
\begin{assumption} \label{ass:w_l2}
$\mathcal{H}_{w} = L^p(\mathcal{Z})$.
\end{assumption}

Define a subset $\bar{\mathcal{S}}^{\mathsf{IV}}_{\mathcal{Z}} \subseteq \hat{\mathcal{S}}_{\mathcal{Z}}^{\mathsf{IV}}$ consisting of invariant predictors that are in $L^{p}(\mathcal{X})$, i.e., $  \bar{\mathcal{S}}^{\mathsf{IV}}_{\mathcal{Z}}= \{ u \;| \; u \in \hat{\mathcal{S}}^{\mathsf{IV}}_{\mathcal{Z}} \;\text{and}\; u \in L^{p}(\mathcal{X})\}$. Let  $\Phi= \mathsf{I}$, where $\mathsf{I}:\mathcal{X}\rightarrow \mathcal{X}$ is the identity mapping.  Following the above notation, the set of invariant predictors and the set of ensemble predictors obtained from NE are  $\hat{\mathcal{S}}^{\mathsf{IV}}_{\mathcal{X}}(\mathsf{I})$ and $\hat{\mathcal{S}}^{\mathsf{EIRM}}_{\mathcal{X}}(\mathsf{I})$ respectively.

\begin{theorem}
\label{thm:identity}
If Assumptions \ref{assum:phi} and \ref{ass:w_l2} are satisfied and $\bar{\mathcal{S}}_{\mathcal{Z}}^{\mathsf{IV}}$ is non-empty, then
 $\bar{\mathcal{S}}_{\mathcal{Z}}^{\mathsf{IV}} = \hat{\mathcal{S}}_{\mathcal{X}}^{\mathsf{IV}}(\mathsf{I}) = \hat{\mathcal{S}}_{\mathcal{X}}^{\mathsf{EIRM}}(\mathsf{I}) $
\end{theorem}

\textbf{Significance of Theorem \ref{thm:identity}.}
 If we fix the representation to identity and play the EIRM game, then it is sufficient to recover all the invariant predictors (with bounded $L^p$ norm) that can be obtained using all the representations $\Phi \in \mathcal{H}_{\Phi}$. Therefore, we can simply fix $\Phi=\mathsf{I}$ and use game-theoretic algorithms for learning equilibria.

\subsection{Existence of NE of $\Gamma^{\mathsf{EIRM}}$ and Invariant Predictors}
In this section, we first argue that there are many settings when  both invariant predictors and the NE exist. 

\textbf{Illustration through generative models.} We use a simplified version of the model described by \cite{peters2016causal}. In each environment $e$, the random variable $X^{e}=[X_1^{e},...,X_{n}^{e}]$ corresponds to the feature vector and $Y^{e}$ corresponds to the label. The data for each environment is generated by i.i.d. sampling $(X^{e},Y^{e})$ from the following generative model. Assume a subset $S^{*} \subset \{1,...,n\}$ is causal for the label $Y^{e}$. 
For all the environments $e$, $X^e$ has an arbitrary distribution and 
$$Y^{e} = g(X^{e}_{S^{*}}) + \epsilon^{e}$$
where $X^{e}_{S^{*}}$ is the vector $X^{e}$ with indices in $S^{*}$,   $g:\mathbb{R}^{|S^{*}|} \rightarrow \mathbb{R}$ is some underlying function and $\epsilon^{e} \sim F^{e}$, $\mathbb{E}[\epsilon_e]=0$, $\epsilon^{e} \perp  X^{e}_{S^{*}}$. Let $\ell$ be the squared error loss function. We fix the representation $\Phi^{*}(X^{e}) = X^{e}_{S^{*}}$. With $\Phi^{*}$ as the representation, the optimal classifier $w$  among all the functions is $g(X^{e}_{S^{*}})$ (this follows from the generative model).
If we assume that $g \in \mathcal{H}_w$, then for each environment $e$, $w^{e}_{*} = g $ is the optimal classifier in $\mathcal{H}_w$. Therefore, $w^{e}_{*}\circ \Phi^{*}  = g$ is the invariant predictor. If $\mathcal{H}_w$ satisfies affine closure, then any decomposition of $g$ is a pure NE of the EIRM game. 
We have illustrated existence of NE and invariant predictor when the data is generated as above and when the class $\mathcal{H}_w$ is sufficiently expressive to capture $g$. Next, we discuss the case when we do not know anything about the underlying data generation process. 

\begin{assumption}
\label{existence1}
\begin{itemize}
\label{assum:existence}
    \item  $\mathcal{H}_{w}$ is a class of linear models, where $w:\mathcal{Z}\rightarrow \mathbb{R}$ and $w(z) = \textbf{w}^{t}z$, where $z\in \mathcal{Z}$. We write $\mathcal{H}_{w}$ as the set of vectors $\textbf{w}$. $\mathcal{H}_{w}$ is a closed, bounded and convex set. The interior of $\mathcal{H}_{w}$ is non-empty. 
    \item  The loss function $\ell(\textbf{w}^{t}z, Y)$, where $Y\in \mathbb{R}$ is the label, is convex and continuous in $\textbf{w}$. For e.g., if loss is cross-entropy for binary classification or loss is mean squared error for regression, then this assumption is automatically satisfied. 
 
\end{itemize}
\end{assumption}

\begin{theorem}
\label{theorem3}
If Assumption \ref{existence1} is satisfied, then a pure strategy Nash equilibrium of the game $\Gamma^{\mathsf{EIRM}}$ exists.
If the weights of all the individual classifiers in the NE are in the interior of $\mathcal{H}_w$, then the corresponding ensemble predictor is an invariant predictor among all the linear models.  
\end{theorem}

 The family $\mathcal{H}_w$ of bounded linear functions does not satisfy affine closure, which is why existence of NE does not immediately imply the existence of invariant predictor (from Theorem \ref{theorem1}). However, if the solution is in the interior of $\mathcal{H}_w$ , then it is the globally optimal solution among all the linear functions, which in fact actually satisfy affine closure. As a result, in this case the invariant predictor also exists.

\textbf{Significance of Theorem \ref{theorem3}} 
Our approach is based on finding the NE. Therefore, it is important to understand when the solutions are guaranteed to exist. In the above theorem, we proved the result for linear models only, but there were no assumptions made on the representation class. In the Appendix Section, we show that for a large class of models, pure NE may not exist but mixed NE (a relaxation of pure NE) are guaranteed to exist. 
Following the sufficient condition for existence of invariant predictors, understanding what conditions cause the NEs to be in the interior or on the boundary of $\mathcal{H}_w$ can help further the theory of invariant prediction. 

\subsection{Algorithms for Finding NE of $\Gamma^{\mathsf{EIRM}}$}
There are different strategies in the literature to compute the equilibrium, such as best response dynamics (BRD) and fictitious play \cite{fudenberg1998theory}, but none of these strategies are guaranteed to arrive at equilibria in continuous games except for special classes of games  \cite{hofbauer2006best, barron2010best, mertikopoulos2019learning,bervoets2016learning}. BRD is one the most popular methods given its intuitive and natural structure. The training of GANs also follows an approximate BRD \cite{goodfellow2014generative}. BRD is not known to converge to equilibrium in GANs. Instead a modification of it proposed recently, \cite{hsieh2018finding} achieves mixed NE. 
Our game $\Gamma^{\mathsf{EIRM}}$ is a non-zero sum game with continuous actions unlike GANs. Since there are no known techniques that are guaranteed to compute the equilibrium (pure or mixed) for these games, we adopt the classic BRD approach. 

In our first approach, we use a fixed representation $\Phi$.  Recall in Theorem \ref{thm:identity}, we showed how just fixing $\Phi$ to identity can be a very effective approach. Hence, we can fix $\Phi$ to be identity mapping or we can select $\Phi$ as some other mapping such as approximation of the map for Gaussian kernel  \cite{rahimi2008random}.
Once we fix $\Phi$, the environments play according to best response dynamics as follows. 
\begin{itemize}
     \item Each environment takes its turn (in a periodic manner with each environment going once) and minimizes its respective objective. 
     \item Repeat this procedure until a certain criterion is achieved, e.g., maximum number of epochs or desired value of training accuracy. 
 \end{itemize} 
 
 The above approach does not give much room to optimize $\Phi$. We go back to the formulation in \eqref{eqn: IRM_ensemble_inequality} and use the upper level optimization objective as a way to guide search for $\Phi$. In this new approach, $\Phi$ is updated by the representation learner periodically using the objective in \eqref{eqn: IRM_ensemble_inequality} and between two updates of $\Phi$ the environments play according to best response dynamics as described above.

\begin{figure}[h]
\includegraphics[width=3.5in]{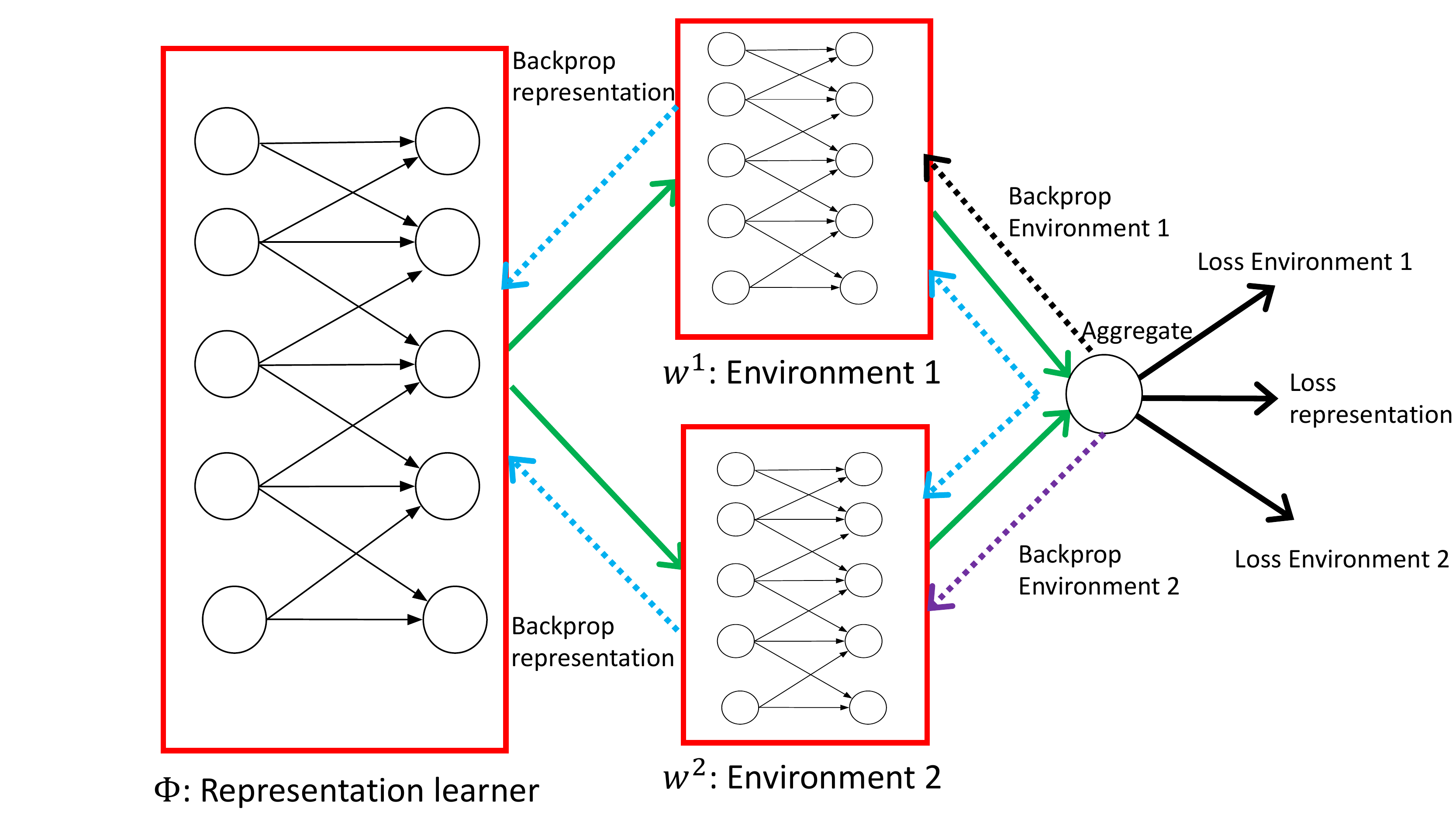}
\caption{Illustration of best response training with 2 environments and representation learner. Dotted lines  for backpropagation and solid lines for forward pass. }
\label{figure: illustrate backprop}
\end{figure}

We now make assumptions on $\mathcal{H}_w$ and $\mathcal{H}_{\Phi}$ and give a detailed algorithm (see Algorithm \ref{alg:brt}) that we use in experiments. We assume that $w^{e}$ is parametrized by family of neural networks $\theta_w \in \Theta_w$ and $\Phi$ is parametrized by family of neural networks $\theta_{\Phi} \in \Theta_{\Phi}$.   In the Algorithm \ref{alg:brt},  one of the variables $\mathsf{Fixed}$-$\mathsf{Phi}$ (for our first approach) or $\mathsf{Variable}$-$\mathsf{Phi}$ is set to true, and then accordingly $\Phi$ remains fixed or is updated periodically.  In Figure \ref{figure: illustrate backprop}, we also show an illustration of the best response training when there are two environments and one representation learner. 

\begin{algorithm}[tb]
  \caption{Best Response Training}
  \label{alg:brt}
\begin{algorithmic}
   \STATE {\bfseries Input:} Data for each environment and combined data
   \WHILE{$\mathsf{iter} \leq \mathsf{iter}_{\mathsf{max}}$}
      \IF{$\mathsf{Fixed}$-$\mathsf{Phi}$}
    \STATE $\Phi_{\mathsf{cur}} = \mathsf{I} $
  \ENDIF
   \IF{$\mathsf{Variable}$-$\mathsf{Phi}$}
  \STATE $\Phi_{\mathsf{nxt}} = \mathsf{SGD} \Big[\sum_{e}R^{e}(w^{av}_{\mathsf{cur}} \circ \Phi_{\mathsf{cur}})\Big]$, $\mathsf{SGD}[.]$:  update using stochastic gradient descent  \STATE $\Phi_{\mathsf{cur}} = \Phi_{\mathsf{nxt}} $
  \ENDIF
   \FOR{$p \in \{1,..K\}$}
   \FOR{$e\in \{1,.., |\mathcal{E}_{tr}|\}$}
   \STATE $w^{e}_{\mathsf{nxt}} = \mathsf{SGD} \Big[R^{e}(w^{av}_{\mathsf{cur}} \circ \Phi_{\mathsf{cur}}) \Big] $
      \STATE $ w^{e}_{\mathsf{cur}}= w^{e}_{\mathsf{nxt}} $
   \ENDFOR
\STATE $\mathsf{iter} = \mathsf{iter} +1$
\STATE $w^{av}_{\mathsf{cur}} = \frac{1}{|\mathcal{E}_{tr}|} \sum_{e}w^{e}_{\mathsf{cur}}$
   \ENDFOR
   \ENDWHILE 

\end{algorithmic}
\end{algorithm}

\section{Experiments} 

\subsection{Benchmarks}
The most important benchmark for comparison is \cite{arjovsky2019invariant}, which we refer to as IRM in the comparisons. We use the architecture described in their work (details in the Appendix Section). We also compare with
\begin{itemize}
    \item Variants of empirical risk minimization: ERM on entire training data (ERM), ERM on each environment separately (ERM $e$ refers to ERM trained on environment $e$), and ERM on data with no spurious correlations.
    \item Robust min-max training: In this method, we minimize the maximum loss across the multiple environments.
\end{itemize}
We have two approaches for EIRM games: one that uses a $\Phi$ fixed to the identity and the other that uses a variable $\Phi$, which we refer to as the F-IRM and V-IRM game, respectively. The details on architectures, hyperparameters, and optimizers used for all the methods are in the Appendix Section. The source-code is available at \url{https://github.com/IBM/IRM-games}.

\subsection{Datasets} In \cite{arjovsky2019invariant}, the comparisons were done on a colored digits MNIST dataset. We create the same dataset for our experiments. In addition, we also create two other datasets that are inspired from Colored MNIST: Colored Fashion MNIST and Colored Desprites. We also create another dataset: Structured Noise Fashion MNIST. In this dataset, instead of coloring the images to establish spurious correlations, we create small patches of noise at specific locations in the image, where the locations are correlated with the labels (detailed description of the datasets is in the Appendix Section). In all the comparisons, we averaged the performance of the different approaches over ten runs.

\textbf{Colored MNIST (Table 1)}
  Standard ERM based approaches, and robust training based approach achieve between 10-15 percent accuracy on the testing set.
F-IRM game achieves 59.9 $\pm$ 2.7 percent testing accuracy. This implies that the model is not using spurious correlation unlike the ERM based approaches, and robust training based approach, that is present in the color of the digit. F-IRM has a comparable mean and a much lower standard deviation than IRM, which achieves 62.75 $\pm$ 9.5 percent. ERM grayscale is ERM on uncolored data, which is why it is better than all.

\begin{table}[t]
\caption{Colored MNIST: Comparison   of methods in terms of training, testing accuracy (mean $\pm$ std deviation).  }
\label{sample-table}
\vskip 0.15in
\begin{center}
\begin{small}
\begin{sc}
\begin{tabular}{lrrr}
\toprule
Algorithm & Train accuracy & Test accuracy\\
\midrule
ERM   & 84.88 $\pm$ 0.16 & 10.45 $\pm$ 0.66\\
ERM 1 & 84.84 $\pm$ 0.21 & 10.86 $\pm$ 0.52 \\
ERM 2& 84.95 $\pm$ 0.20 & 10.05 $\pm$ 0.23\\
Robust min max   & 84.25 $\pm$ 0.43 & 15.24 $\pm$ 2.45      \\
F-IRM game & \textbf{63.37} $\pm$ \textbf{1.14}   & \textbf{59.91} $\pm$ \textbf{2.69}  \\ 
V-IRM game & \textbf{63.97} $\pm$ \textbf{1.03} & \textbf{49.06} $\pm$ \textbf{3.43}\\ 
IRM  & \textbf{59.27} $\pm$ \textbf{4.39} & \textbf{62.75} $\pm$ \textbf{9.59}\\ 
ERM grayscale  & 71.81 $\pm$ 0.47 & 71.36$\pm$ 0.65\\ 
Optimal  & 75  & 75 \\ 
\bottomrule
\end{tabular}
\end{sc}
\end{small}
\end{center}
\vskip -0.1in
\end{table}

\begin{table}[t]
\caption{Colored Fashion MNIST: Comparison   of methods in terms of training, testing accuracy (mean $\pm$ std deviation). }
\label{sample-table}
\vskip 0.15in
\begin{center}
\begin{small}
\begin{sc}
\begin{tabular}{lrrr}
\toprule
Algorithm & Train accuracy & Test accuracy\\
\midrule
ERM   & 83.17 $\pm$ 1.01 & 22.46 $\pm$ 0.68\\
ERM 1  & 81.33 $\pm$ 1.35 & 33.34 $\pm$ 8.85 \\
ERM 2  & 84.39 $\pm$ 1.89 & 13.16 $\pm$ 0.82\\
Robust min max   & 82.81 $\pm$ 0.11 & 29.22 $\pm$ 8.56      \\
F-IRM game & \textbf{62.31} $\pm$ \textbf{2.35}   & \textbf{69.25} $\pm$ \textbf{5.82}  \\ 
V-IRM game & \textbf{68.96} $\pm$ \textbf{0.95} & \textbf{70.19} $\pm$ \textbf{1.47}\\ 
IRM  & \textbf{75.01} $\pm$ \textbf{0.25} & \textbf{55.25} $\pm$ \textbf{12.42}\\ 
ERM grayscale & 74.79 $\pm$ 0.37 & 74.67$\pm$ 0.48\\ 
Optimal & 75  & 75 \\ 
\bottomrule
\end{tabular}
\end{sc}
\end{small}
\end{center}
\vskip -0.1in
\end{table}

\begin{table}[t]
\caption{Colored Desprites: Comparison   of methods in terms of training, testing accuracy (mean $\pm$ std deviation).}
\label{sample-table}
\vskip 0.15in
\begin{center}
\begin{small}
\begin{sc}
\begin{tabular}{lrrr}
\toprule
Algorithm & Train accuracy & Test accuracy\\
\midrule
ERM   & 85.01 $\pm$ 0.03 & 9.97 $\pm$ 0.05\\
ERM 1 & 81.33 $\pm$ 1.35 & 33.34 $\pm$ 8.85 \\
ERM 2& 84.39 $\pm$ 1.89 & 13.16 $\pm$ 0.82\\
Robust min max   & 84.94 $\pm$ 0.09 & 10.28 $\pm$ 0.33      \\
F-IRM game & \textbf{53.36} $\pm$ \textbf{1.40}   & \textbf{48.61} $\pm$ \textbf{3.06}  \\ 
V-IRM game & \textbf{56.31} $\pm$ \textbf{4.94} & \textbf{50.04} $\pm$ \textbf{0.15}\\ 
IRM  & \textbf{52.67} $\pm$ \textbf{2.40} & \textbf{51.82} $\pm$ \textbf{5.95}\\ 
ERM grayscale & 67.67 $\pm$ 0.58 & 66.97$\pm$ 0.69\\ 
Optimal & 75  & 75 \\ 
\bottomrule
\end{tabular}
\end{sc}
\end{small}
\end{center}
\vskip -0.1in
\end{table}

\textbf{Colored Fashion MNIST (Table 2)}
We observe that the V-IRM game performs the best both in terms of the mean and the standard deviation achieving 70.2 $\pm$ $1.5$ percent.

\textbf{Colored Desprites (Table 3)}
We observe that V-IRM game achieves 50.0 $\pm$ 0.2 percent while IRM achieves 51.8 $\pm$ 6 percent.

\begin{table}[t]
\caption{Structured Noise Fashion MNIST: Comparison   of methods in terms of training, testing accuracy (mean $\pm$ std deviation).}
\label{sample-table}
\vskip 0.15in
\begin{center}
\begin{small}
\begin{sc}
\begin{tabular}{lrrr}
\toprule
Algorithm & Train accuracy & Test accuracy\\
\midrule
ERM   & 83.49 $\pm$ 1.22 & 20.13 $\pm$ 8.06\\
ERM 1 & 81.80 $\pm$ 1.50 & 30.94 $\pm$ 1.01 \\
ERM 2 & 84.66 $\pm$ 0.40 & 11.98 $\pm$ 0.23\\
Robust  min max   & 82.78 $\pm$ 1.32& 25.59 $\pm$ 9.14      \\
F-IRM game & \textbf{51.54} $\pm$ \textbf{2.96}   & \textbf{62.03} $\pm$ \textbf{2.02}  \\ 
V-IRM game & \textbf{47.70} $\pm$ \textbf{1.69} & \textbf{61.46} $\pm$ \textbf{0.53}\\ 
IRM  & \textbf{52.57} $\pm$ \textbf{9.95} & \textbf{63.92} $\pm$ \textbf{10.95}\\ 
ERM no noise & 74.79 $\pm$ 0.37 & 74.67$\pm$ 0.48\\ 
Optimal & 75  & 75 \\ 
\bottomrule
\end{tabular}
\end{sc}
\end{small}
\end{center}
\vskip -0.1in
\end{table}

\textbf{Structured Noise Fashion MNIST (Table 4)}
We observe that F-IRM achieves 62.0 $\pm$ 2.0 percent and is comparable with IRM that achievs 63.9 $\pm$ 10.9 percent; again observe that we have a lower standard deviation.

\subsection{Analyzing the Experiments}

In this section, we use plots of F-IRM game played on Colored Fashion MNIST (plots for both F-IRM and V-IRM on all other datasets are similar and are in the Appendix Section).  In Figure \ref{fig2:analysis1}, we show the accuracy of the ensemble model on the entire data and the two environments separately. In the initial stages, the training accuracy increases and eventually it starts to oscillate. Best response dynamics can often oscillate \cite{herings2017best,fudenberg1998theory, barron2010best}.

Next, we demistify these oscillations and explain their importance.

\subsubsection{Explaining the mechanism of oscillations}
\begin{figure}[t]
\centering
  \includegraphics[ width=2.5in]{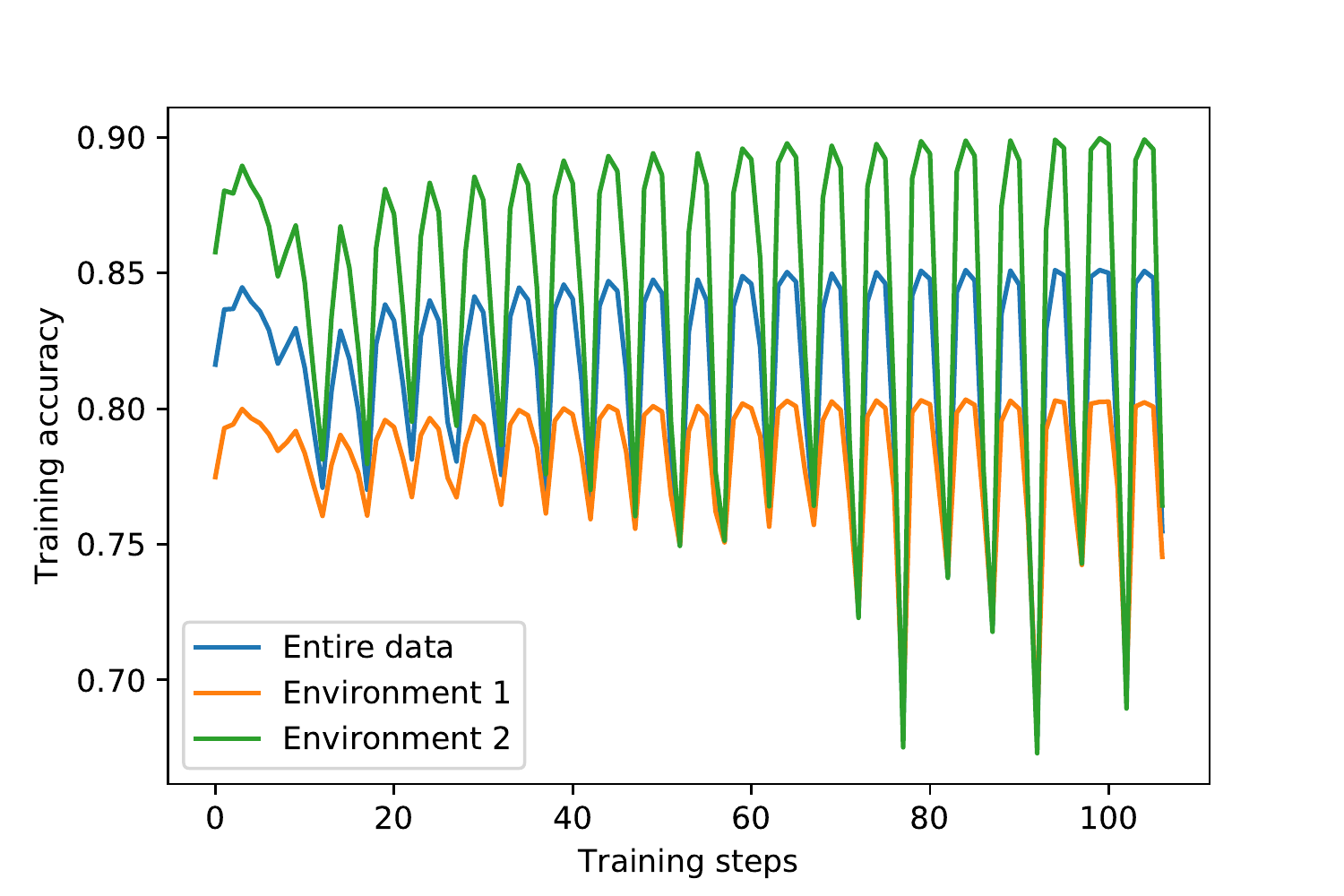}
  \caption{F-IRM, Colored Fashion MNIST: Comparing accuracy of ensemble }
  \label{fig2:analysis1}
\end{figure}

\begin{figure}[t]
  \centering
  \includegraphics[width=2.5in]{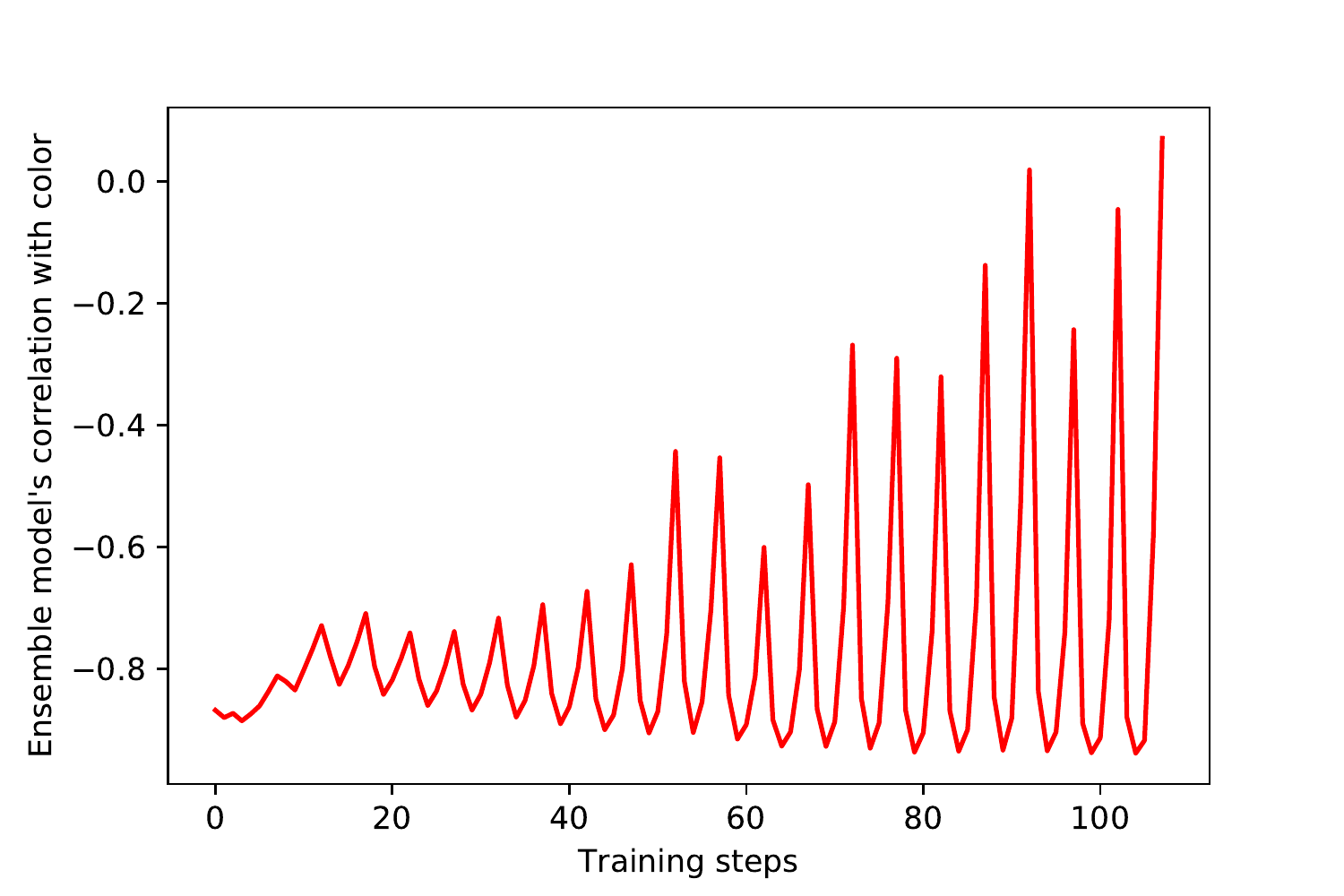}
  \caption{ F-IRM, Colored Fashion MNIST: Correlation of the ensemble model with color}
   \label{fig3:correlations1}
\end{figure}

The oscillation has two states. In the first state, the ensemble model performs well 88 $\%$ accuracy. In the second state, the accuracy dips to 75 $\%$.
 In Figure \ref{fig3:correlations1}, we plot the correlation between the ensemble model and the color. When the oscillations appear in training accuracy in Figure \ref{fig2:analysis1}, the correlation also start to oscillate in Figure \ref{fig3:correlations1}.  In the first state when the model performs well, the model is heavily correlated (negative correlation) with the color. In the second state, the model performs worse,  observe that the model now has much less correlation (close to zero) with the color. We ask two questions: (i) Why do the oscillations persist in the training accuracy plot (Figure \ref{fig2:analysis1}) and correlation plot (Figure \ref{fig3:correlations1})?, and (ii) How do the oscillations emerge?

\begin{figure}[t]
  \centering
  \includegraphics[width=2.5 in]{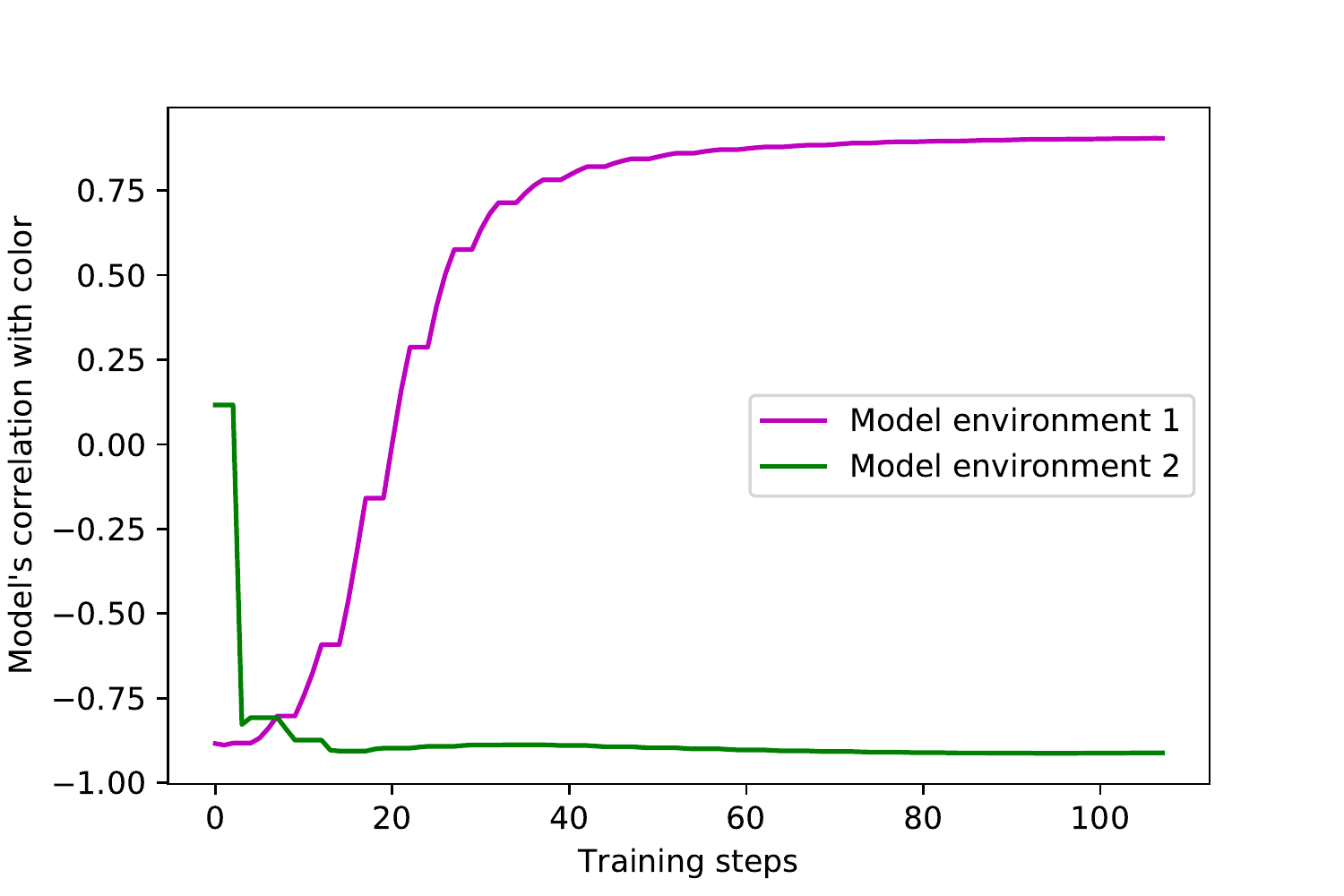}
  \caption{F-IRM, Colored Fashion MNIST: Correlations of the individual models with color}
     \label{fig5:correlations1}

\end{figure}

\textbf{Why do the oscillations persist?}
In our experiments there are two environments, the labels are binary, and we want to maximize the log-likelihood. Let $s_j$ be the score vector from environment $j$'s classifier, $p$ be the softmax of $s$ and $\tilde{y}$ be the one hot encoded vector of labels.  The gradient of the log-likelihood w.r.t.\ the scores given by each model for a certain instance $x$ (see derivation in the Appendix Section) is:
\begin{equation}
\begin{split}
\frac{\partial \log(p_y) } {\partial s_j} = \bar{y} - p = \tilde{e}.
    \end{split}
    \label{eqn:error_backprop1}
\end{equation}
where $\tilde{e}$ is the error vector. The error $\tilde{e}$ is determined by the both the models (both models impact $p$), it backpropagates and impacts individual weights. We argue next that the examples over which error occur are very different in the two states and that is the reason for oscillations.

 Consider the step when the correlation (absolute value) between the ensemble model and color is high. In this step, it is the turn of Model 1 to train. Observe that the accuracy of the model is high because the ensemble model is exploiting the spurious correlations with the color. We approximate this mathematically. The score from Model $j$ for Label 1  is $s_j^{1}-s_j^{0} \approx \beta_j^{t}\phi_j^{nc}(x) + \gamma_j \phi_j^{c}(x) $, where $\phi_j^{nc}$ are the features that are not correlated with the color, $\phi_j^{c}$ is the indicator of the color. From Figure \ref{fig5:correlations1}, $\gamma_1$ and $\gamma_2$ should have opposite signs, i.e. positive and negative  respectively. In the current step, $\gamma_2$ dominates $\gamma_1$, which is why the ensemble model has a heavy negative correlation. The errors \eqref{eqn:error_backprop1} that  backpropagate come from the examples for which exploiting spurious correlation with color does not work, i.e., the color is not indicative of the digit. During this step Model 1 is trained, backpropagation will change the weights such that $\gamma_1$ increases. As a result, the ensemble model's correlation with the color decreases (as we see in Figure \ref{fig3:correlations1}). In the next step, it is the turn of Model 2 to train. Model 2's environment has more examples than environment 1 where exploiting the color can help improve its accuracy. 
As a result, error from these examples  backpropagate and $\gamma_2$ decreases. This brings the ensemble model back to being negatively correlated with colors and  also the training accuracy back to where it was approximately. This cycle of push and pull between the models continues.

\textbf{How do these cycles emerge?}
The oscillations are weak at the beginning of the training. In the beginning, when Model 2 trains, the impact of the errors (from examples where spurious correlations can be exploited) on changing the weights are much stronger than when Model 1 trains, as the number of examples that benefit from spurious correlations is much larger in comparison. As the training proceeds, this impact decreases as many examples are classified correctly by using spurious correlations while the weights continue to accumulate for Model 1, thus giving rise to oscillations.

\textbf{How to terminate?} 
We terminate training when the oscillations are stable and when the ensemble model is in the lower accuracy state, which corresponds to the state with lower correlation with color. To ensure the oscillations are stable, we do not terminate until a certain number of steps have been completed (in our experiments we set this duration to be number of steps= (training data size)/(batch size)). To capture the model in a state of lower correlation with color, we set a threshold on accuracy (we decide the threshold by observing the accuracy plot); we terminate only when the training accuracy falls below this threshold.

\section{Conclusion}
We developed a new framework based on game-theoretic tools to learn invariant predictors. We work with data from multiple environments. In our framework, we set up an ensemble game; we construct an ensemble of classifiers with each environment controlling one portion of the ensemble.  Remarkably, the set of solutions to this game is exactly the same as the set of invariant predictors across training environments. The proposed framework performs comparably to the existing framework of \cite{arjovsky2019invariant} and also exhibits lower variance. We hope this framework opens new ways to address other problems pertaining to invariance in causal inference using tools from game theory.

\section{Appendix}
\subsection{Examples of hypothesis classes that satisfy affine closure} 
\begin{itemize}
\item \textbf{Linear classifiers}: The sum of linear functions (polynomial) leads to a linear function (polynomial), and so does scalar multiplication. Therefore,  linear classifiers satisfy affine closure.
\item \textbf{Reproducing Kernel Hilbert Space} (RKHS): RKHS is a Hilbert space, which is a vector spaces of functions. Therefore,  \textbf{kernel based classifiers} \cite{hofmann2008kernel} satisfy affine closure.

\item \textbf{Ensemble models:} Consider binary classification and boosting models \cite{freund1999short}. Let $\mathcal{H}_{\mathsf{weak}}$ be the set of weak learners $\omega:\mathcal{X} \rightarrow \mathbb{R}$. The final function that is input to a sigmoid is $w = \sum_{m=1}^{k} \theta_m \omega_m$, where each $\theta_m \in \mathbb{R}$. The set of functions spanned by the weak learners is defined as $\mathsf{Span}(\mathcal{H}_{\mathsf{weak}}) = \{\sum_{m=1}^{k} \theta_m \omega_m |  \forall m \in \{1,..,k\}, \theta_{m} \in \mathbb{R}, k \in \mathbb{N}\}$. $\mathsf{Span}(\mathcal{H}_{\mathsf{weak}})$ forms a vector space. Therefore,  ensemble models that may use arbitrary number of weak learners satisfy affine closure.

\item $L^{p}$ \textbf{spaces}.  The  set of functions $f:\mathcal{X}\rightarrow \mathbb{R}$ for which  $\|f\|_{p} = [\int_{\mathcal{X}} |f(x)|^{p}dx]^{\frac{1}{p}} < \infty$ is defined as $L^{p}(\mathcal{X})$.  $L^{p}(\mathcal{X})$ is a vector space \cite{ash2000probability}.

\end{itemize}

\textbf{ReLU networks with arbitrary depth}: Neural networks are known to be universal function approximators. 
Let us assume $\mathcal{X}$ to be a compact subset of $\mathbb{R}^{n}$. The output of a ReLU network is a continuous function on $\mathcal{X}$, which implies it is bounded and thus the function described by a ReLU network is in $L^{1}(\mathcal{X})$ space. It is clear that the set of functions parametrized by ReLU networks are a subset of  functions in $L^{1}(\mathcal{X})$ space.  In the other direction, from \cite{lu2017expressive}, we know that ReLU networks can come arbitrarily close to any function in $L^{1}$ sense. Since ReLU networks come arbitrarily close to the function and are not exactly equal we cannot argue that affine closure is satisfied. However, we argue later that since the networks can arbitrarily approximate any function in $L^{1}(\mathcal{X})$ it is sufficient to prove our results (our main result Theorem 1 and Corollary 1). 
\subsection{Theorems and Proofs}
In this section, we discuss the proofs to the lemmas, theorems, and corollaries in the paper. 
\begin{theorem_sup}If Assumption 1 holds, then
$\tilde{\mathcal{S}}^{\mathsf{IV}} = \tilde{\mathcal{S}}^{\mathsf{EIRM}}$
\end{theorem_sup}

\begin{proof}
In the first part, we want to show that  $\tilde{\mathcal{S}}^{\mathsf{IV}} \subseteq  \tilde{\mathcal{S}}^{\mathsf{EIRM}}$. We will use proof by contradiction.

Let us assume that there exists an element $(\Phi, \{w^{q}\}_{q=1}^{|\mathcal{E}_{tr}|}, w) \in \tilde{\mathcal{S}}^{\mathsf{IV}}$, which does not belong to $\tilde{\mathcal{S}}^{\mathsf{EIRM}}$. This implies that there exists at least one $e \in \mathcal{E}_{tr}$ in the ensemble game, which strictly prefers the action $\bar{w}^{e} \in \mathcal{H}_w$ to following its current action $w^{e}$. In other words, at least one of the inequalities in \eqref{eqn: IRM_ensemble_inequality} is not satisfied, which can be written as

\begin{equation}
  R^{e} \Bigg(\Big[\frac{\bar{w}^{e} + \sum_{q\not=e}w^{q}}{|\mathcal{E}_{tr}|}\Big]\circ \Phi\Bigg) <  R^{e}(w \circ\Phi) 
  \label{maintheorem:eqn1}
\end{equation}
The function $w^{'} = \frac{\bar{w}^{e} + \sum_{q\not=e}w^{q}}{|\mathcal{E}_{tr}|} \in \mathcal{H}_w$ (From Assumption 1). Therefore, $w^{'}$ is a strictly better classifier than $w$ with a fixed representation $\Phi$ for environment $e$, which contradicts the condition that 
$w \in \arg\min_{\bar{w} \in \mathcal{H}_w} R^e(\bar{w}\circ \Phi)$ (which follows from  $(\Phi, \{w^{q}\}_{q=1}^{|\mathcal{E}_{tr}|}, w) \in \tilde{\mathcal{S}}^{\mathsf{IV}}$ ). 

This proves the first part. 

In the second part, we want to show that $\tilde{\mathcal{S}}^{\mathsf{EIRM}} \subseteq \tilde{\mathcal{S}}^{\mathsf{IV}}$. Let us assume that there exists an element $(\Phi, \{w^{q}\}_{q=1}^{|\mathcal{E}_{tr}|}, w) \in \tilde{\mathcal{S}}^{\mathsf{EIRM}}$, which does not belong to $\tilde{\mathcal{S}}^{\mathsf{IV}}$. Following Assumption 1, $w$ lies in $\mathcal{H}_w$.
Since $(\Phi, \{w^{q}\}_{q=1}^{|\mathcal{E}_{tr}|}, w) \not\in \tilde{\mathcal{S}}^{\mathsf{IV}}$
 there exists at least one $e\in \mathcal{E}_{tr}$ and a classifier $w^{'} \in \mathcal{H}_w$ strictly better than $w$ for a fixed representation $\Phi$. If this were not the case, $w$ will be an invariant predictor w.r.t. $\Phi$ across $\mathcal{E}_{tr}$, which would contradict $(\Phi, \{w^{q}\}_{q=1}^{|\mathcal{E}_{tr}|}, w) \not\in \tilde{\mathcal{S}}^{\mathsf{IV}}$.  Therefore 
\begin{equation}
    R^{e}(w^{'} \circ \Phi) < R^{e} (w\circ \Phi)
    \label{maintheorem:eqn2}
\end{equation}
Let us construct a new auxiliary classifier $\tilde{w}^{e}$ as follows. 
$\tilde{w}^{e} = w^{'} |\mathcal{E}_{tr}| - \sum_{q\not=e} w^q$. 
It follows from Assumption 1 that $\tilde{w}^{e} \in \mathcal{H}_w$. Observe that the ensemble defined as  $\frac{\tilde{w}^{e} + \sum_{q\not=e}w^q}{|\mathcal{E}_{tr}|}$ simplifies to $w^{'}$.  This means that environment $e$ can deviate from $w^{e}$ to $\tilde{w}^{e} \in \mathcal{H}_w$ and strictly gain from this deviation.  This contradicts the fact that $\{w^{q}\}_{q=1}^{|\mathcal{E}_{tr}|}$ is a Nash equilibrium ($\{w^{q}\}_{q=1}^{|\mathcal{E}_{tr}|}$ is a Nash equilibrium because $(\Phi, \{w^{q}\}_{q=1}^{|\mathcal{E}_{tr}|}, w) \in \tilde{\mathcal{S}}^{\mathsf{EIRM}}$).

\end{proof}

\begin{corollary_sup}
If Assumption 1 holds, then $\hat{\mathcal{S}}^{\mathsf{IV}} = \hat{\mathcal{S}}^{\mathsf{EIRM}}$
\end{corollary_sup}
\begin{proof}
The proof follows straightaway from Theorem 1. For each  $w \circ \Phi \in \hat{\mathcal{S}}^{\mathsf{IV}}  $ we look at the corresponding tuple
 $(\Phi, \{w^{q}\}_{q=1}^{|\mathcal{E}_{tr}|}, w) \in \tilde{\mathcal{S}}^{\mathsf{IV}}$. From Theorem 1, $(\Phi, \{w^{q}\}_{q=1}^{|\mathcal{E}_{tr}|}, w) \in \tilde{\mathcal{S}}^{\mathsf{EIRM}}$. Therefore, $w \circ \Phi \in \hat{\mathcal{S}}^{\mathsf{EIRM}}$. The other side follows the same way.
\end{proof}

\subsubsection{Extending Theorem 1 and Corollary 1 to ReLU networks }

In the proof of Theorem 1, we used the affine closure property in \eqref{maintheorem:eqn1} and \eqref{maintheorem:eqn2}. However, in \eqref{maintheorem:eqn1} and \eqref{maintheorem:eqn2}, we only need to construct models that can achieve risk arbitrarily close to the models in the LHS of equations \eqref{maintheorem:eqn1} and \eqref{maintheorem:eqn2}. Let $\mathcal{H}_w$ the set of functions of ReLU networks with arbitrary depth defined on compact sets $\mathcal{X}$. These functions are in $L^{1}$ class as explained earlier. From \cite{lu2017expressive}, we can choose ReLU networks from $\mathcal{H}_w$ that approximate the classifiers in the LHS of \eqref{maintheorem:eqn1} and \eqref{maintheorem:eqn2} arbitrarily. We elaborate on this. Suppose the function to be approximated in the LHS is $f$. From \cite{lu2017expressive}, for each $\epsilon>0$, there exists a ReLU network $\hat{f}$ such that $\mathbb{E}_{X}[|f-\hat{f}|] \leq \epsilon$. The question is does $\mathbb{E}_{X}[|f-\hat{f}|] \leq \epsilon$ also ensure that the difference in risks is mitigated $|R^{e}(f,Y)-R^{e}(\hat{f}, Y)| \leq \tilde{\epsilon}$.  If the loss function $\ell$ is Lipschitz in the scores (e.g., cross-entropy loss, hinge loss), then  if the functions are arbitrarily close the risks will also be arbitrarily close. We show this below.
\begin{equation}
    \begin{split}
        & |R^{e}(f,Y)-R^{e}(\hat{f}, Y)| \\
      =&   \;|\mathbb{E}^{e}[\ell(f(X),Y)-\ell(\hat{f}(X),Y)]| \\
      \leq &\; \mathbb{E}^{e}[|\ell(f(X),Y)-\ell(\hat{f}(X),Y)|]  \\
      \leq &\; \mathbb{E}^{e}[L|f(X) -\hat{f}(X)|]
    \end{split}
\end{equation}
where $L$ is the Lipschitz constant for $\ell$.

Below we  illustrate an example of Lipschitz continuous loss $\ell$. Consider cross entropy for binary classification (labels $Y=0$ and $Y=1$). Suppose $f(x)=s$ is the score assigned to class $1$, it is converted into probability as $e^{s}/(1+e^{s})$. The cross-entropy loss is simplified as 
\begin{equation}
    \ell(s,Y) = Ys -\log(1+e^{s})
\end{equation}
Observe $\frac{\partial \ell(s,Y)}{\partial s} = Y - \frac{1}{1+e^s}$ and $|\frac{\partial \ell(s,Y)}{\partial s}| \leq 1$. Therefore, $\ell(s,Y) $ is Lispchitz continuous in $s$.
\begin{lemma_sup}
If Assumptions 2 and 3 are satisfied, then for any $w^{'}\in \mathcal{H}_w$ and $\Phi \in \mathcal{H}_{\Phi}$,    $w^{'} \circ \Phi^{-1} \in L^p(\mathcal{Z})$.
\end{lemma_sup}
\begin{proof}
To show $w^{'} \circ \Phi^{-1} \in L^{p}(\mathcal{Z}) $ let us first express the integral $\int_{\mathcal{Z}} |w^{'}(\Phi^{-1}(z))|^p dz$ by using substitution rules \cite{rudin1987real}. We can use the substitution rule because both $\mathcal{X}$ and $\mathcal{Z}$ are $n$ dimensional, the function $\Phi$ is bijective, differentiable and Lipshitz continuous (From Asumption 2 and 3). Substitute $z= \Phi(x)$. Then, $ \int_{\mathcal{Z}} |w^{'}(\Phi^{-1}(z))|^p  dz = \int_{\Phi^{-1}(\mathcal{Z})} |w^{'}(x)|^p \mathsf{det}( J(\Phi(x)) ) dx $  . Here $J(\Phi(x))$ is the Jacobian of the transformation $\Phi$. Since $\Phi$ is a Lipschitz continuous map, its determinant is also bounded. We show this as follows.

Lipschitz continuity implies that for any $x,x' \in {\cal X}$, $ \lVert \Phi(x) - \Phi(x') \rVert \leq  \gamma \lVert  x- x' \rVert$ where $\gamma$ is the Lipschitz constant. In particular, since $\Phi(\cdot)$ is differentiable (Assumption 2), this means that the length of any partial derivative vector $ \lVert \frac{\delta \Phi(x)}{\delta x_i} \rVert \leq \gamma$ for any coordinate $i \in [n]$. Now, we apply the Hadamard inequality \cite{garling2007inequalities}
for the determinant of the square matrix $J(\Phi(x))$:

$\mathsf{det}(J(\Phi(x))) \leq \prod \limits_{i \in [n]} \lVert \frac{\delta \Phi(x)}{\delta x_i} \rVert \leq \gamma^n $. Therefore,
\begin{align}
\int_{\mathcal{Z}} |w^{'}(\Phi^{-1}(z))|^p  dz & = \int_{\Phi^{-1}(\mathcal{Z})} |w^{'}(x)|^p \mathsf{det}( J(\Phi(x)) ) dx\nonumber \\
\hfill &\leq \gamma^n \int_{\Phi^{-1}(\mathcal{Z})}|w^{'}(x)|^p dx \nonumber \\
 \hfill &\leq \gamma^n \int_{{\cal X}} |w^{'}(x)|^p dx
\end{align}

Since, $w \in L^p({\cal X})$ (Assumption 3) we have that  $w^{'} \circ \Phi^{-1} \in L^p({\cal Z})$  from the above inequality. 
\end{proof} 

\begin{theorem_sup}
\label{thm:identityx}
If Assumptions 2 and 3 are satisfied and $\bar{\mathcal{S}}_{\mathcal{Z}}^{\mathsf{IV}}$ is not empty, then
 $\bar{\mathcal{S}}_{\mathcal{Z}}^{\mathsf{IV}} = \hat{\mathcal{S}}_{\mathcal{X}}^{\mathsf{IV}}(\mathsf{I}) = \hat{\mathcal{S}}_{\mathcal{X}}^{\mathsf{EIRM}}(\mathsf{I}) $
\end{theorem_sup}

\begin{proof}

In the first part, we want to show that $ \bar{\mathcal{S}}_{\mathcal{Z}}^{\mathsf{IV}} \subseteq \hat{\mathcal{S}}_{\mathcal{X}}^{\mathsf{IV}}(\mathsf{I})  $.  We will use proof by contradiction.

 Suppose $(w\circ\Phi) \in \bar{\mathcal{S}}_{\mathcal{Z}}^{\mathsf{IV}} $ but not in $\hat{\mathcal{S}}_{\mathcal{X}}^{\mathsf{IV}}(\mathsf{I})$. 
First note that $w \circ \Phi  \in L^{p}(\mathcal{X})$ (From definition of the set $\bar{\mathcal{S}}_{\mathcal{Z}}^{\mathsf{IV}}$).  
This implies that there must exist  an environment $e$ and a classifier $w^{'}:\mathcal{X}\rightarrow \mathcal{Y}$ which is better than   $(w\circ\Phi)$. Therefore, we can state that
\begin{equation}
    R^e(w^{'}) < R^{e}(w \circ \Phi) 
\end{equation}

Define a classifier $\tilde{w} = w^{'} \circ \Phi^{-1}$. From Lemma 1 it follows $\tilde{w} \in L^p(\mathcal{Z})$. 
Define the risk achieved by this classifier as $R^{e}(\tilde{w}\circ \Phi)$. We simplify this as follows. \begin{equation}
\begin{split}
&R^{e}(\tilde{w}\circ \Phi) = R^{e}((w^{'}\circ\Phi^{-1}) \circ \Phi) = \\ & R^{e}(w^{'}\circ(\Phi^{-1} \circ \Phi)) = R^{e}(w^{'} \circ \mathsf{I}) = R^{e}(w')
\end{split}
\label{eqn2:thm2}
\end{equation}
Therefore, the risk of $\tilde{w} \circ \Phi$ is better than the risk achieved by $w \circ \Phi$. This contradicts that  $w\circ \Phi$ is an invariant predictor. We show this as follows. Since $w \circ \Phi$ is an invariant predictor with $\Phi$ as the representation it implies $w \in \arg\min_{\bar{w}} R^{e}(\bar{w} \circ \Phi)$. However, $\tilde{w}$ is clearly  better than $w$ with $\Phi$ as the representation \eqref{eqn2:thm2} , which leads to a contradiction. This proves the first part.

The second side $ \hat{\mathcal{S}}_{\mathcal{X}}^{\mathsf{IV}}(\mathsf{I})  \subseteq \bar{\mathcal{S}}_{\mathcal{Z}}^{\mathsf{IV}}  $. Suppose $w \in \hat{\mathcal{S}}_{\mathcal{X}}^{\mathsf{IV}}(\mathsf{I}) $ but not in $\bar{\mathcal{S}}_{\mathcal{Z}}^{\mathsf{IV}}$. Select any $\Phi :\mathcal{X}\rightarrow \mathcal{Z}$ from the set of representations for which invariant predictors exist in the set $\bar{\mathcal{S}}_{\mathcal{Z}}^{\mathsf{IV}}$ (recall that we assumed $\bar{\mathcal{S}}_{\mathcal{Z}}^{\mathsf{IV}}$ is not empty). Define a predictor $\tilde{w} = w \circ \Phi^{-1}$. Since $w \in L^{p}(\mathcal{X})$, from Lemma 1 we know that  $\tilde{w}$ is in $L^{p}(\mathcal{Z})$. There should exist an environment $e$ for which $\tilde{w}$ is not the optimal classifier given $\Phi$ otherwise $w$ will be in the set $\bar{\mathcal{S}}_{\mathcal{Z}}^{\mathsf{IV}}$, which would be a contradiction. $\Phi$ is a representation for which an invariant predictor exists, let $w^{'}$ be the classifier and $w'\circ \Phi$ be the invariant predictor in $\bar{\mathcal{S}}_{\mathcal{Z}}^{\mathsf{IV}}$.
$\exists$ an environment $e$ for which  $w^{'}$   is strictly better than $\tilde{w}$ given $\Phi$. We write this condition as 
\begin{equation}
R^{e}(w^{'}\circ \Phi) < R^{e}(\tilde{w}\circ \Phi) = R^{e}(w)
\label{thm_identity_eq_ss}
\end{equation}

$w^{'} \circ \Phi  \in \bar{\mathcal{S}}_{\mathcal{Z}}^{\mathsf{IV}}$ and from the definition of the set it follows that $w^{'} \circ \Phi \in L^{p}(\mathcal{X})$.  Also, $w^{'} \circ \Phi$ is better than $w$ from \eqref{thm_identity_eq_ss}. However, $w$ is an invariant predictor with $\Phi= \mathsf{I}$, which leads to contradiction.

From Theorem 2 it follows that $\hat{\mathcal{S}}_{\mathcal{X}}^{\mathsf{EIRM}}(\mathsf{I}) = \hat{\mathcal{S}}^{\mathsf{IV}}_{\mathcal{X}}(\mathsf{I})$. 
This completes the proof.  
\end{proof}

\textbf{When $\Phi=\mathsf{I}$, can the game recover the solution that focuses on causal parents?} 
We will consider the following data generation process.
The data for each environment is generated by i.i.d. sampling $(X^{e},Y^{e})$ from the following generative model. Assume a subset $S^{*} \subset \{1,...,n\}$ is causal for the label $Y^{e}$. 
For all the environments $e$, $X^e$ has an arbitrary distribution and 
$$Y^{e} \leftarrow g(X^{e}_{S^{*}}) + \epsilon^{e}$$
where $X^{e}_{S^{*}}$ is the vector $X^{e}$ with indices in $S^{*}$,   $g:[-u,u]^{|S^{*}|} \rightarrow \mathbb{R}$ is some underlying function and $\epsilon^{e} \sim F^{e}$, $\mathbb{E}[\epsilon_e]=0$, $\epsilon^{e} \perp  X^{e}_{S^{*}}$. We assume  $g \in  L^{p}([-u,u]^{|S^{*}|}) $ and $\mathcal{H}_w= L^{p}([-u,u]^{|S^{*}|})$.
Let $\ell$ be the squared error loss function. We fix the representation $\Phi^{*}(X^{e}) = X^{e}_{S^{*}}$. With $\Phi^{*}$ as the representation, the optimal classifier $w$  among all the functions is $g(X^{e}_{S^{*}})$ (this follows from the generative model).
For each environment $e$, $w^{e}_{*} = g $ is the optimal classifier in $\mathcal{H}_w$. Therefore, $w^{e}_{*}\circ \Phi^{*}  = g$ is the invariant predictor. Since $\mathcal{H}_{w}$ is affine closed $\frac{1}{|\mathcal{E}_{tr}|}\sum_{e} w^{e}_{*}\circ\Phi^{*}$ is an invariant predictor obtained from the EIRM game. Define a function $\tilde{g}(X^{e}) = g(X^{e}_{S^{*}}) $. Since $g \in L^{p}([-u,u]^{|S^{*}|})$, $\tilde{g} \in L^{p}([-u,u]^{n}) $. We claim that $\Phi=\mathsf{I}$ elicits $\tilde{g}\circ \mathsf{I}$ as an invariant predictor. Suppose this was not the case then for some environment $e$, there exists $\hat{g} \in L^{p}([-u,u]^{n})$ which achieves a lower risk than $\tilde{g}$, i.e. $R^{e}(\hat{g} ) < R^{e}(\tilde{g} )$. Consider $$\min_{\bar{g} \in  L^{p}([-u,u]^{n})}\mathbb{E}[(Y^e-\bar{g})^2]$$

We simplify the objective as follows
 $$\mathbb{E}[(Y^e-\bar{g})^2] =\mathbb{E}[(g-\bar{g})^2 + (\epsilon^e)^2 + 2(g-\bar{g})\epsilon^e ]  = \mathbb{E}[(g-\bar{g})^2 + (\epsilon^e)^2 ] \geq \mathbb{E}[(\epsilon^e)^2 ]$$
 
$\bar{g}=\tilde{g}$ is an optimal solution since $\mathbb{E}[(Y^e-\tilde{g})^2] = \mathbb{E}[(\epsilon^e)^2 ]$. This contradicts that $\mathbb{E}[(Y^e-\hat{g})^2] < \mathbb{E}[(Y^e-\tilde{g})^2] $. 

Therefore, to conclude even when $\Phi = \mathsf{I}$ the EIRM game will recover the invariant predictor that focuses on the causal parents of $Y$.

\textbf{When $\Phi=\mathsf{I}$, can the game recover the solution when causal parents are not directly observed?}

We consider a similar generative process as described above except, we now assume that the causal features are not directly observed
$$Y^{e} \leftarrow g(Z^e_{S^{*}}) + \epsilon^{e}$$

where  $Z^{e}_{S^{*}}$ is the vector $Z^{e}$ with indices in $S^{*}$, $g:[-u,u]^{|S^{*}|} \rightarrow \mathbb{R}$, 
$g \in L^{p}([-u,u]^{|S^{*}|})$. We assume that we do not observe $Z^e$ directly and instead observe $X^e \leftarrow f(Z^e)$, where $f$ is an invertible map. In addition, we assume that $f$ satisfies the Assumption 2. Let $\Phi^{*}=  f^{-1}$ and define $P_{S^{*}}$ as the projection function that projects the input onto indices in $S^{*}$.  Observe that $g \circ P_{S^{*}} \circ f^{-1}(X^e) = g(Z^e_{S^{*}})$. Fix  $w_{e}^{*} = g \circ P_{S^{*}}$. Therefore  $ w_{*}^{e} \circ \Phi^{*}$ is an invariant predictor. Observe that $g \circ P_{S^{*}} \in \mathcal{H}_{w}  = L^{p}([-u,u]^{n})$. Since $\mathcal{H}_{w}$ is affine closed $\frac{1}{|\mathcal{E}_{tr}|}\sum_{e} w^{e}_{*}\circ\Phi^{*}$ is an invariant predictor obtained from the EIRM game. 

What happens when $\Phi=\mathsf{I}$? Is  $(g \circ P_{S^{*}}\circ  f^{-1} ) \circ \mathsf{I}$ an invariant predictor? Note that  $g \circ P_{S^{*}} \circ f^{-1} \in L^{p}([-u,u]^{n})$ (To see why this is the case, use the following observations. $g \circ P_{S^{*}} \in L^{p}([-u,u]^{n})$, $f$ satisfies Assumption 2, and use Lemma 1). From the generative model it is clear that there cannot be another classifier that is strictly better than 
$g \circ P_{S^{*}} \circ f^{-1}$ for any environment. Therefore, $g \circ P_{S^{*}} \circ f^{-1}$ is indeed an invariant predictor. Since $g \circ P_{S^{*}}\circ f^{-1} \in L^{p}([-u,u]^{n})$ and $L^{p}([-u,u]^{n})$ is affine closed, $g \circ P_{S^{*}}\circ f^{-1}$ is also a solution obtained from the EIRM game with $\Phi= \mathsf{I}$.

\begin{theorem_sup}
If Assumption 4 is satisfied, then a pure strategy Nash equilibrium of the game $\Gamma^{\mathsf{EIRM}}$ exists.
If the weights of all the individuals in the NE are in the interior of $\mathcal{H}_w$, then the corresponding ensemble predictor is an invariant predictor among all linear models.  
\end{theorem_sup}

\begin{proof} We will use the classic result from \cite{debreu1952social}, which shows the sufficient conditions for the existence of pure Nash equilibrium in continuous action games. We provide this result in the next section Theorem 5, where we continue the discussion on concepts in game theory. Informally speaking, the result states that if the game is concave with compact and convex action sets, then the pure Nash equilibrium exists.  

The set of actions of each environment $\mathcal{H}_w$ is a closed bounded and convex subset (following the Assumption 4).  
Recall the definition of the utility of a player $e$ in the EIRM game is given as 

\begin{equation}
\begin{split}
&u_{e}[w^{e}, w^{-e}, \Phi] = -R^{e}( w^{av} \circ \Phi ) = \\ & =-\mathbb{E}^{e}[\ell((w^{av} \circ \Phi)(x), Y) ] 
\end{split}
\label{thm4:utility}
\end{equation}

Following Assumption 4, we simplify the inner term in the expectation as follows. 
\begin{equation}
    \begin{split}
    &    \ell((w^{av} \circ \Phi)(x), Y)  = \ell(\Phi(x)^{t}[\frac{1}{|\mathcal{E}_{tr}|}\sum_{q=1}^{|\mathcal{E}_{tr}|}\textbf{w}^{q}] ,Y)  
    \end{split}
    \label{eqn:comp}
\end{equation}
$\ell(\Phi(x)^{t} \textbf{w}, Y) = h_Y(\textbf{w}) $. $h_{Y}(\textbf{w})$ is a convex function of $\textbf{w}$ (From Assumption 4). 
 Define $g: \mathbb{R}^{d} \times \mathbb{R}^{d}... \times \mathbb{R}^d \rightarrow \mathbb{R}^d$  as  $g(\textbf{w}^{1},...,\textbf{w}^{|\mathcal{E}_{tr}|}) = \frac{1}{|\mathcal{E}_{tr}|}\sum_{k}\textbf{w}^{k}$.
 Note that $g$ is an affine mapping.
 The function in \eqref{eqn:comp} can be expressed as $h_Y(g(\textbf{w}^{1},...\textbf{w}^{|\mathcal{E}_{tr}|}))$. The composition of a convex function with an affine function is also convex \cite{boyd2004convex}. We use this to conclude that the composition $h_Y(g(\textbf{w}^{1},...\textbf{w}^{|\mathcal{E}_{tr}|}))$ is a convex function in $\textbf{w}^{1},...\textbf{w}^{|\mathcal{E}_{tr}|}$. 
 We express \eqref{thm4:utility} in terms of $h$ and $g$ as 
 \begin{equation}
 \begin{split}
     u_{e}[w^{e}, w^{-e}, \Phi] =
 -\mathbb{E}^{e}[h_Y(g(\textbf{w}^{1},...\textbf{w}^{|\mathcal{E}_{tr}|}))]     
\end{split}
 \end{equation}

 Each term inside the expectation above is concave. Therefore, $u_{e}$ is concave in $w^{e}$ (follows directly from Jensen's inequality applied to $u_e$). 
$h_Y$ is a continuous function in $\textbf{w}$ (from Assumption 4) and $g$ is a continuous function as well, the composition of the two continuous functions is also continuous. As a result $u_e$ is continuous. Therefore, the EIRM game above satisfies the  assumptions in Theorem 5 (\cite{debreu1952social}, which implies that  a pure NE exists. This proves the first part of the theorem. We now discuss the second part of the  which provides a simple condition for the existence of invariant predictor. 

Say the weights  that comprise one of the NE are given as $\{w^{q}_{*}\}_{q=1}^{|\mathcal{E}_{tr}|}$. This set of weights satisfy
\begin{equation}
    w^{e}_{*}= \arg\min_{w^{e} \in \mathcal{H}_w} -u_{e}(w^{e},w_{*}^{-e}, \Phi)
\end{equation}

From Assumption 4,  $w^{e}_{*}$ is in the interior of $\mathcal{H}_w$. Therefore,  we can construct a ball around it in which it is the smallest point, which implies it is  a local minima of $-u_{e}(w^{e},w_{*}^{-e}, \Phi)$. Since local minima is also the global minima for convex functions; it follows that the solution would be equivalent to searching over the space of all the linear functions, i.e.
\begin{equation}
    w^{e}_{*}= \arg\min_{w^{e} \in \mathbb{R}^{d}} -u_{e}(w^{e},w_{*}^{-e}, \Phi)
\end{equation}
The above argument holds for all the environments because each solution $w_{*}^{e}$ is in the interior. Therefore, we can transform the EIRM game from the current restricted space $\mathcal{H}_w$ to the space of all the linear functions. The space of the linear functions satisfy affine closure property unlike the space of bounded linear functions $\mathcal{H}_w$. From Theorem 1 it follows that the ensemble classifier $\frac{1}{|\mathcal{E}_{tr}|}\sum_{q=1}^{|\mathcal{E}_{tr}|}w_{*}^{q}$ composed with $\Phi$ will be an invariant predictor. 
\end{proof}

In Theorem 3 we assumed that the model and the representation are both linear functions. We now discuss the existence under a more general class of models. 

\textbf{Assumption 5}
$\mathcal{H}_w$ is a family of functions parametrized by  $\theta \in \Theta$.  We assume that $\Theta$ is compact. We assume $w_{\theta} \in \mathcal{H}_w$, where $w_{\theta}:\mathbb{R}^{d} \rightarrow \mathbb{R}$ is  continuous in its inputs.

Consider a multilayer perceptron (MLP) with say ReLU activation. Each weight in the network belongs $[w_{min},w_{max}]$. This family of neural networks satisfies the Assumption 5 above.

Suppose that each environment is looking to solve for a probability distribution over the parameters of the neural network written as vector $w^{e}$ given as $p_{w^{e}}$.  We rewrite the expected loss of the environments as follows. 
$$\bar{u}_{e}(p_{w^{e}}, p_{w^{-e}}, p_{\Phi}) = \mathbb{E}_{\Pi_{e}p_{w^{e}} \times p_{\Phi}}\Big[u_{e}(w^{e},w^{-e}, \Phi)\Big] $$. 
We use $\bar{u}_e$ as the utility of each environment in the EIRM game.
\begin{theorem_sup}
If Assumption 5 is satisfied, then a mixed strategy Nash equilibrium of $\Gamma^{\mathsf{EIRM}}$ is guaranteed to exist.
\end{theorem_sup}

\begin{proof}
The proof is a direct consequence of the existence result \cite{glicksberg1952further}, which we restate in Theorem 7.
\end{proof}
The main message of the above theorem is that we relax the requirement of having a deterministic classifier, then we are guaranteed to have a solution for general models as well.

\subsection{Game Theory Concepts Continued}
This section is a continuation to the Section 3.1 on Game Theory Concepts. We discuss some classic results on the existence of NE.
Let us now  consider continuous action games. We make the following assumption. 

\textbf{Assumption NE 1}
For each $i$:
\begin{itemize}
    \item  $S_i$ is a compact, convex subset of $\mathbb{R}^{n_i}$
    \item $u_{i}(s_i,s_{-i})$ is continuous in $s_{-i}$
    \item $u_{i}(s_i,s_{-i})$ is continuous and concave in $s_i$ .
\end{itemize}

\begin{theorem_sup} \cite{debreu1952social}
If Assumption NE 1 is satisfied for game $\Gamma$, then a pure strategy Nash equilibrium exists.
\end{theorem_sup}
We  extend the definition of pure strategy NE to mixed strategies (discussion on mixed strategies given in the next section, where we continue the discussion on concepts in game theory), where instead of choosing an action deterministically, each player chooses a probability distribution over the set of actions. We assume that each set $S_i$ is a compact subset of $\mathbb{R}^{n_i}$. Define the set of Lesbegue measures over $S_i$ as $\Delta(S_i)$.
Each player $i$, draws a probability distribution  $\theta_i$ from $\Delta(S_i)$. The joint strategy played by all the players is the product of their individual distributions written as $\Pi_{k\in N}\theta_k$

\textbf{Nash equilibrium in mixed strategies.} A strategy $\theta^{*} = \Pi_{k\in N}  \theta_k^{*}$  is said to be a mixed strategy Nash Equilibrium (NE) if it satisfies
$$\mathbb{E}_{\theta^{*}}\Big[u_i(S_{i},S_{-i}^{*})\Big] \geq \mathbb{E}_{\theta_{-i}^{*}}\Big[u_i(k,S_{-i})\Big], \forall k \in S_{i}, \forall i$$
where $\theta_{-i}^{*} = \Pi_{k \not =i}  \theta_{k}^{*}$.

\begin{theorem_sup}
\cite{nash1950equilibrium}
Every finite game has a mixed strategy Nash equilibrium.
\end{theorem_sup}

Next, we relax some of the above assumptions. 

\textbf{Assumption NE 2}
For each $i$
\begin{itemize}
    \item  $S_i$ is a non empty, compact subset of $\mathbb{R}^{n_i}$
    \item $u_i(s_i, s_{-i})$ is continuous in $s_i$ and $s_{-i}$
\end{itemize}

\begin{theorem_sup}
\cite{glicksberg1952further} If Assumption NE 2 is satisfied, then the game  has a mixed strategy Nash equilibrium.
\end{theorem_sup}

\subsection{Deriving the expression for backpropagation}
For instance $x$, the predicted score from Environment 1,2 (Model 1,2) for class $k$ is given as $w_1^k \circ x $, $w_2^k \circ x$ respectively, where $w_j^k$ is the score output by neural network $j$ for class $k$.  The overall score is given as $w_1^k\circ x + w_2^k \circ x$. We take the softmax to get the overall probability for class $k$ as 

\begin{equation}
    p_{k}  = \frac{\exp{\Big[w_1^k \circ x + w_2^k \circ x\Big]}}{\sum_j\exp{\Big[ w_1^j \circ x + w_2^j \circ x\Big]}}
\end{equation}
The softmax vector is $p = [p_0,p_1]$. Denote $w_{j}^{k}\circ x  = s_j^k$. 
The log-likelihood for instance $x$ with label $y$ is given as 

\begin{equation}
\begin{split}
    & \log[p_y]  \\ 
    &    =  w_1^y \circ x + w_2^y \circ x - \log\Big(\sum_j\exp{\Big[ w_1^j \circ x + w_2^j \circ x\Big]}\Big)\\  & =s_1^y + s_2^y - \log\Big(\sum_j\exp{\Big[ s_1^j + s_2^j\Big]}\Big) 
    \end{split} 
\end{equation}

The gradient of log-likelihood w.r.t score of each model is given as 

\begin{equation}
\begin{split}
  \frac{\partial \log[p_y] } {\partial s_j^k} &= I(k=y) - \frac{\exp{\Big[ s_1^k + s_2^k\Big]}}{\sum_j \exp{\Big[ s_1^j + s_2^j\Big]} } \\
    & = I(k=y) - p_k 
    \end{split}
    \label{eqn:error_backprop0}
\end{equation}
We convert $y$ into a one hot encoded vector $\bar{y}$ and simplify the above expression as 
\begin{equation}
\begin{split}
\frac{\partial \log[p_u] } {\partial s_j} = \bar{y} - p = \tilde{e}
    \end{split}
    \label{eqn:error_backprop}
\end{equation}
\subsection{Computing Environment}
The experiments were done on 2.3 GHZ Intel Core i9 processor with 32 GB memory (2400 MHz DDR4). 
\subsection{Description of the Datasets}

\subsubsection{Colored MNIST Digits}
We use the exact same environment as in \cite{arjovsky2019invariant}. \cite{arjovsky2019invariant} propose to create an environment for training to classify digits in MNIST digits data \footnote{\url{https://www.tensorflow.org/api_docs/python/tf/keras/datasets/mnist/load_data}}, where the images in MNIST are now colored in such a way that the colors spuriously correlate with the labels. The task is to classify whether the digit is less than 5 (not including 5) or more than 5. There are three environments (two training containing 30,000 points each, one test containing 10,000 points)  We add noise to the preliminary label ($\tilde{y}=0$ if digit is between 0-4 and $\tilde{y}=1$ if the digit is between 5-9) by flipping it with 25 percent probability to construct the final label. We sample the color id $z$ by flipping the final labels with probability $p_e$, where $p_e$ is $0.2$ in the first environment, $0.1$ in the second environment, and $0.9$ in the third environment. The third environment is the testing environment. We color the digit red if $z = 1$ or green if $z = 0$.

\subsubsection{Colored Fashion MNIST}
We modify the fashion MNIST dataset \footnote{\url{https://www.tensorflow.org/api_docs/python/tf/keras/datasets/fashion_mnist/load_data}} in  a manner similar to the MNIST digits dataset. Fashion MNIST data has images from different categories:  ``t-shirt'', ``trouser'', ``pullover'', ``dress'', ``coat'', ``sandal'', ``shirt'', ``sneaker'', ``bag'', ``ankle boots''. We add colors to the images in such a way that the colors correlate with the labels. The task is to classify whether the image is that of foot wear or a clothing item.
There are three environments (two training, one test)  We add noise to the preliminary label ($\tilde{y}=0$: ``t-shirt'', ``trouser'', ``pullover'', ``dress'', ``coat'', ``shirt'' and $\tilde{y}=1$: ``sandle'', ``sneaker'', ``ankle boots'') by flipping it with 25 percent probability to construct the final label. We sample the color id $z$ by flipping the noisy label with probability $p_e$, where $p_e$ is $0.2$ in the first environment, $0.1$ in the second environment, and $0.9$ in the third environment, which is the test environment. We color the object red if $z = 1$ or green if $z = 0$.

\subsubsection{Colored Desprites Dataset}
We modify the Desprites dataset \footnote{https://github.com/deepmind/dsprites-dataset} in a manner similar to the MNIST digits dataset. The task is to classify if the image is a circle or a square.  We take the preliminary binary labels $\tilde{y}=0$ for a circle and $\tilde{y}=1$ for a square. We add noise to the preliminary label by flipping it with 25 percent probability to construct the final label. We sample the color id $z$ by flipping the noisy label with probability $p_e$, where $p_e$ is $0.2$ in the first environment, $0.1$ in the second environment, and $0.9$ in the third environment, which is the test environment. We color the object red if $z = 1$ or green if $z = 0$.

\subsubsection{Structured Noise in Fashion MNIST}
In the previous three experiments, we used color in the images to create correlations. In this experiment, we use a different mechanism to create correlations in Fashion MNIST dataset. We add a small square (3$\times$ 3), in the top left corner of some images and an even smalller square (2 $\times$ 2) in the bottom right corner of other images.  The location of the box is correlated with labels. The preliminary labels are the same as in the other experiment with Fashion MNIST. 
There are three environments (two training, one test).  We add noise to the preliminary label by flipping it with 25 percent probability to construct the final label. We sample the location id $z$ by flipping the noisy label with probability $p_e$, where $p_e$ is $0.2$ in the first environment, $0.1$ in the second environment, and $0.9$ in the third environment, which is the test environment. We place the square in the top left if $z = 1$ or bottom right if $z = 0$.

\subsubsection{Architecture, Hyperparameter and Training Details}

\textbf{Architecture for 2 player EIRM game with fixed $\Phi$}

In the game with fixed $\Phi$, we used the following architecture for the two models.
The model used is a simple multilayer perceptron with following parameters. 

\begin{itemize}
    \item Input layer: Input batch $(\mathsf{batch}, \mathsf{len}, \mathsf{wid},\mathsf{depth})$ $\rightarrow$ Flatten
    \item Layer 1: Fully connected layer, output size = 390, activation = ELU, L2-regularizer = 1.25e-3, Dropout = 0.75
    \item Layer 2: Fully connected layer, output size = 390, activation = ELU, L2-regularizer = 1.25e-3, Dropout = 0.75
    \item Output layer: Fully connected layer, output size = 2
\end{itemize}
We use the above architecture across all the experiments. The shape of the input in the above architecture depends on  the dimensions of the data that are input.

\textbf{Architecture for 2 player EIRM game with variable $\Phi$}

In the game with variable $\Phi$, we used the following architecture. 

The architecture for the representation learner is
\begin{itemize}
    \item Input layer: Input batch $(\mathsf{batch},  \mathsf{len}, \mathsf{wid},\mathsf{depth})$ $\rightarrow$ Flatten
    \item Layer 1: Fully connected layer, output size = 390, activation = ELU, L2-regularizer = 1.25e-3, Dropout = 0.75
    \item Output layer: Fully connected layer, output size = 390, activation = ELU, L2-regularizer = 1.25e-3, Dropout = 0.75
\end{itemize}

The output from the representation learner above is fed into two MLPs one for each environment (we use the same architecture for both environments).

\begin{itemize}
    \item Layer 1: Fully connected layer, output size = 390, activation = ELU, L2-regularizer = 1.25e-3, Dropout = 0.75
    \item Layer 2: Fully connected layer, output size = 390, activation = ELU, L2-regularizer = 1.25e-3, Dropout = 0.75
    \item Output layer: Fully connected layer, output size = 2
\end{itemize}
We use the above architecture across all the experiments. The shape of the input in the above architecture depends on  the dimensions of the data that are input.

\textbf{Optimizer and other hyperparameters}
 We used  Adam optimizer for training with learning rate set to 2.5e-4. We optimize the cross-entropy loss function. We set the batch size to 256. We terminate the algorithm according to the rules we explained in the paper. Thus the number of training steps can vary across different trials. There is a warm start phase for all the methods; we set the warm start phase to be equal to the number of steps in one epoch, where one epoch is the (training data size/ batch size). For the setup with fixed $\Phi$, we set the period to be 2, i.e. in one step first model trains and in the other step the second model trains and this cycle repeats throughout the training. For the setup with variable $\Phi$, we let the two environments and representation learner take turns to update their respective models, environment 1 trains in one step, environment 2 trains in the next step, representation learner trains, and this cycle continues.  

\textbf{Architecture for IRM \cite{arjovsky2019invariant}}

We used the same architecture that they described in the github repository. \footnote{\url{https://github.com/facebookresearch/InvariantRiskMinimization}}. We describe their architecture below. 
\begin{itemize}
    \item Input layer: Input batch $(\mathsf{batch},  \mathsf{len}, \mathsf{wid},\mathsf{depth})$ $\rightarrow$ Flatten
    \item Fully connected layer, output size = 390, activation = ReLU, L2-regularizer = 1.1e-3
    \item Fully connected layer, output size = 390, activation = ReLU, L2-regularizer = 1.1e-3
    \item Output layer: Fully connected layer, output size= 2
\end{itemize}
\textbf{Optimizer, hyperparameters and some remarks}
 We used  Adam optimizer for training with learning rate set to 4.89e-4. We optimize the cross-entropy loss function. We set the batch size to 256. The total number of steps is set to 500. The penalty weight is set to 91257. The penalty term is only used after 190 steps. The code from \cite{arjovsky2019invariant} uses a normalization trick to the loss to avoid gradient explosion. We found that this strategy was not useful in all settings. Therefore, we carried out experiments for both the cases (with and without normalization of loss) and report the case for which the accuracy is higher.

\subsection{Figures Continued} 

\begin{figure}
\centering

  \includegraphics[width=2.75in]{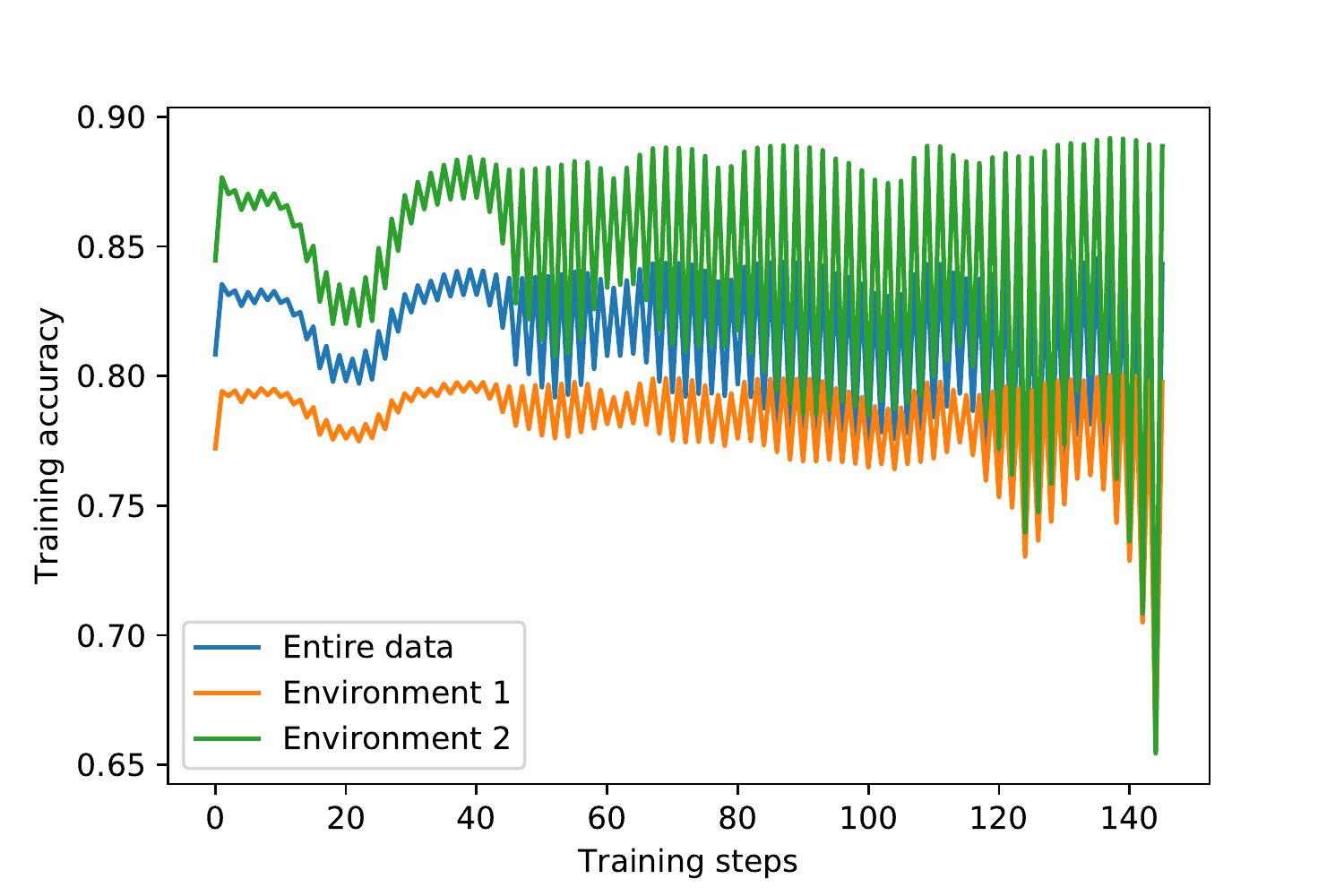}
  \caption{F-IRM, Colored Fashion MNIST: Comparing accuracy  of ensemble}

\label{figs5}
\end{figure}

\begin{figure}
\centering

  \includegraphics[width=2.75 in]{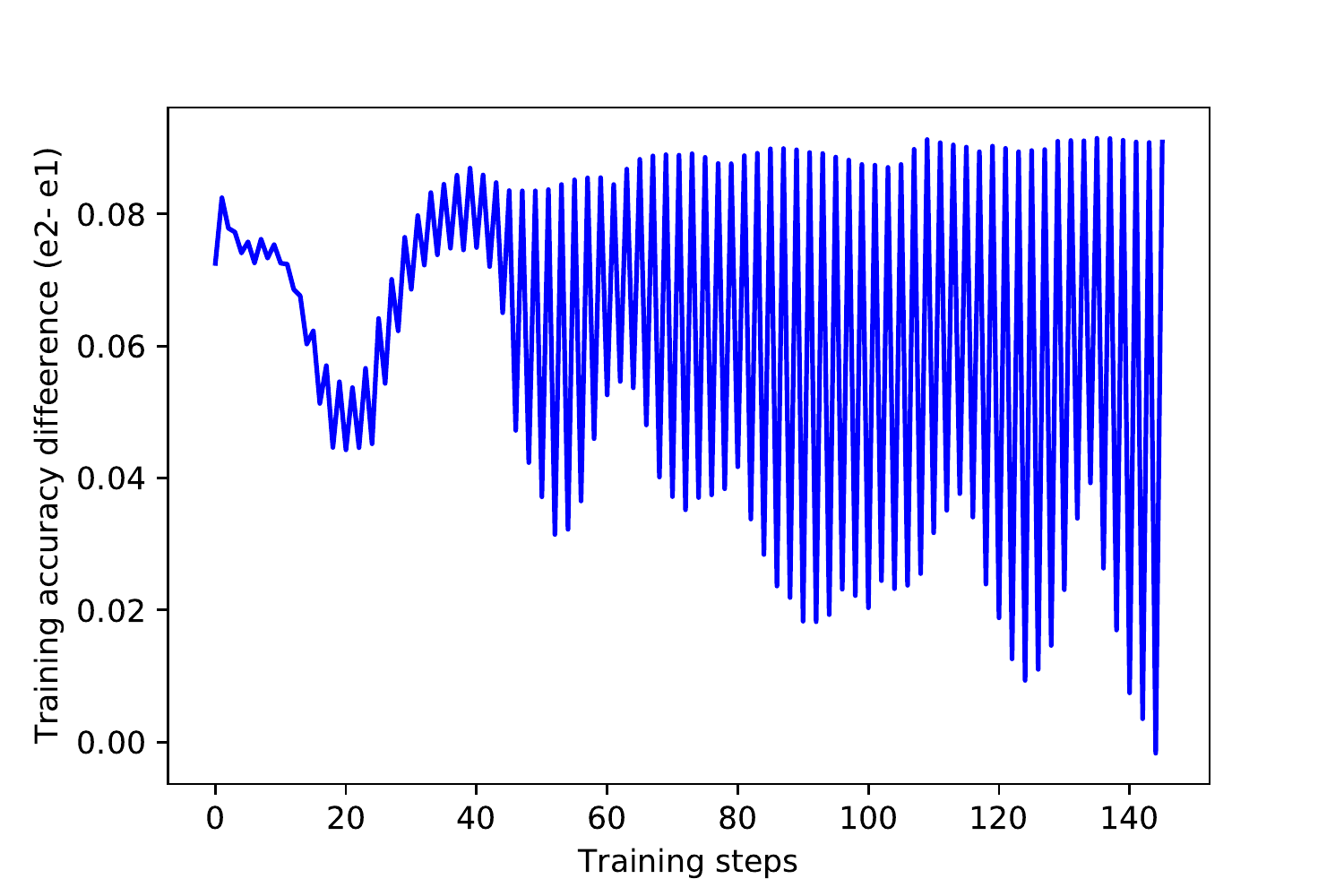}
  \caption{F-IRM, Colored Fashion MNIST: Difference in accuracy of the ensemble model between the two environments }

\label{figs6}
\end{figure}

\begin{figure}
\centering

  \includegraphics[width=2.5in]{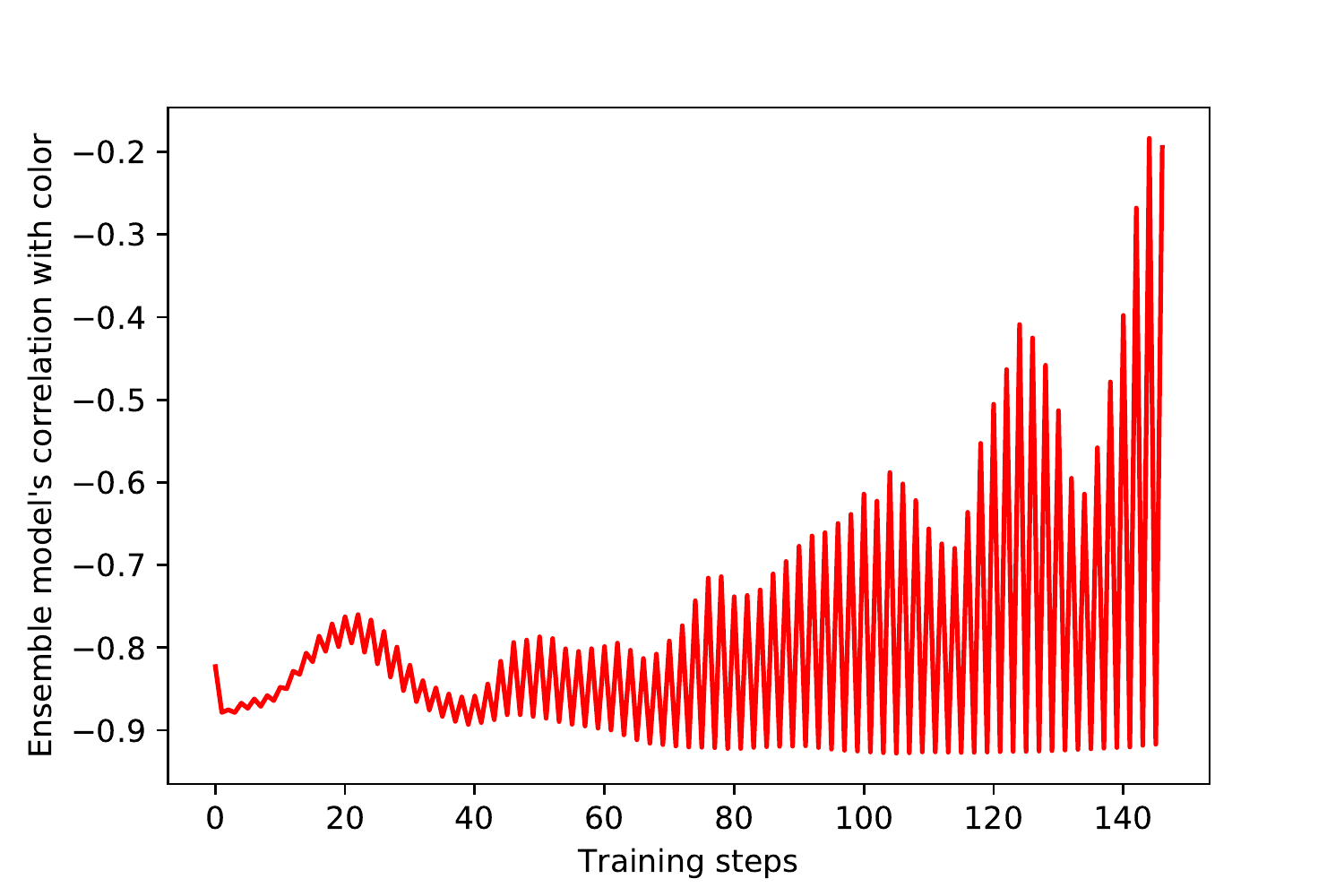}
  \caption{F-IRM, Colored Fashion MNIST: Ensemble's correlation with color}

\label{figs7}
\end{figure}

\begin{figure}
\centering

  \includegraphics[width=2.5 in]{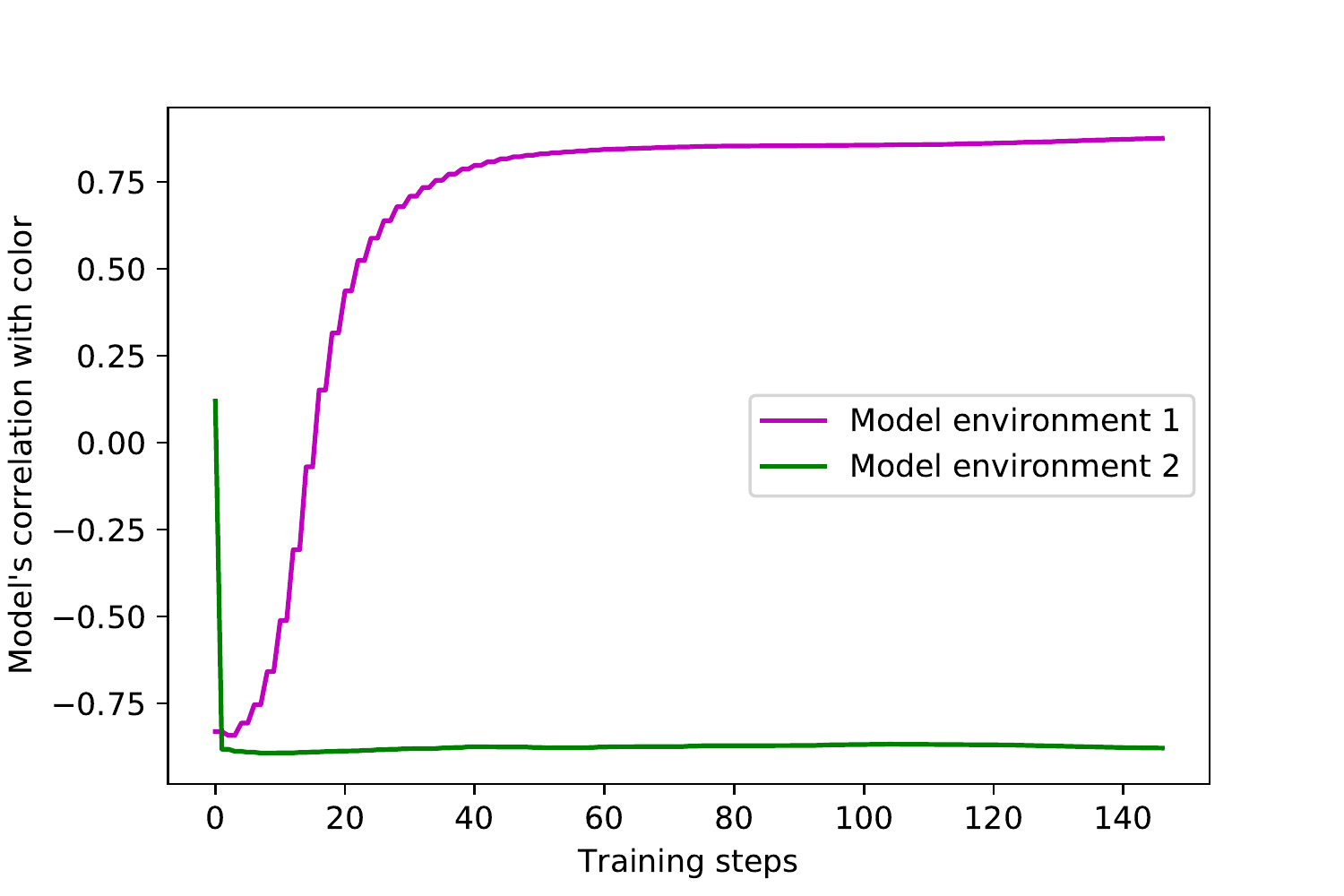}
  \caption{F-IRM, Colored Fashion MNIST: Compare individual model correlations}

\label{figs8}
\end{figure}

\begin{figure}
\centering
  \includegraphics[width=2.75in]{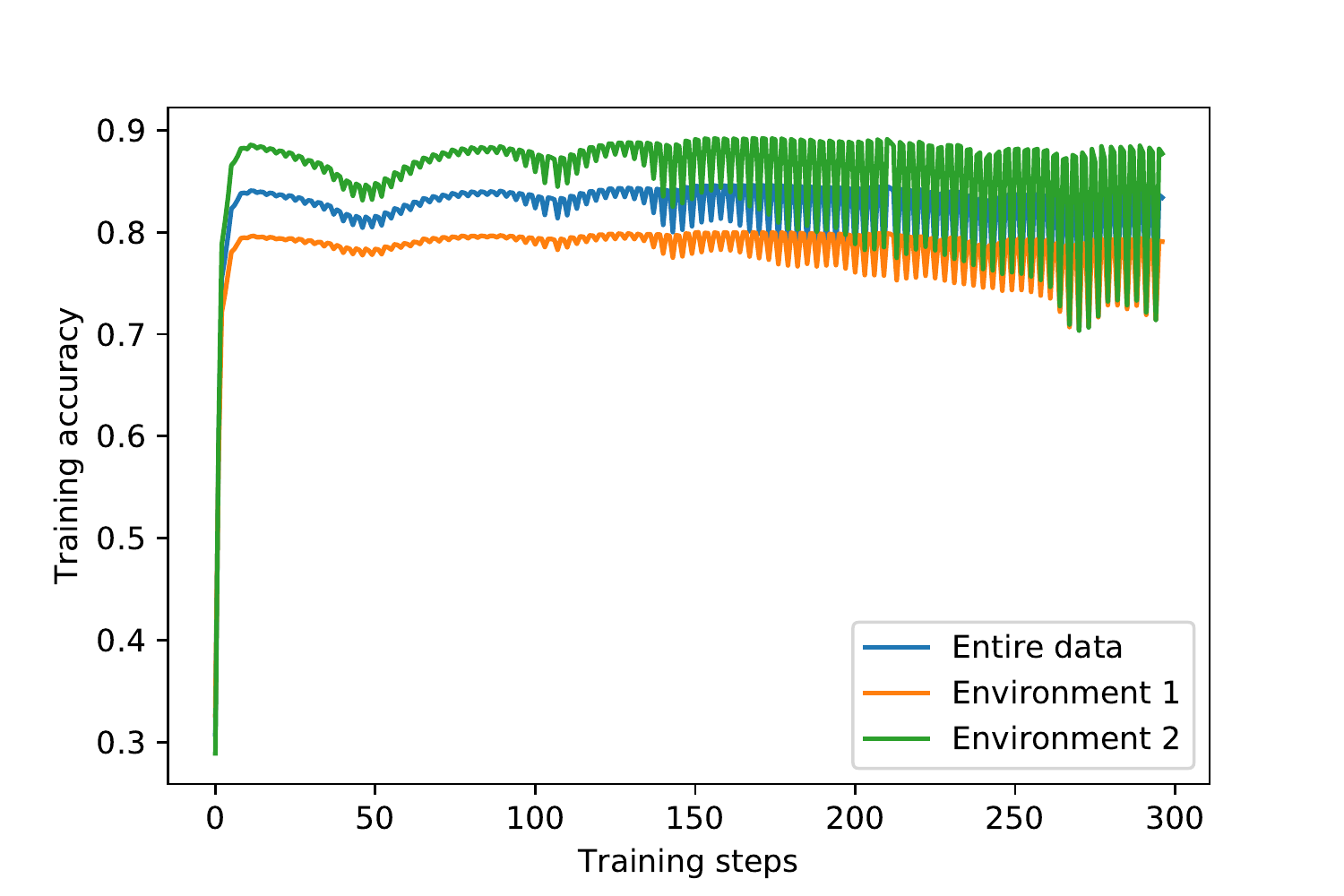}
  \caption{V-IRM Colored Fashion MNIST: Comparing accuracy of ensemble}
\label{figs9}
\end{figure}

\begin{figure}
\centering

  \includegraphics[width=2.75 in]{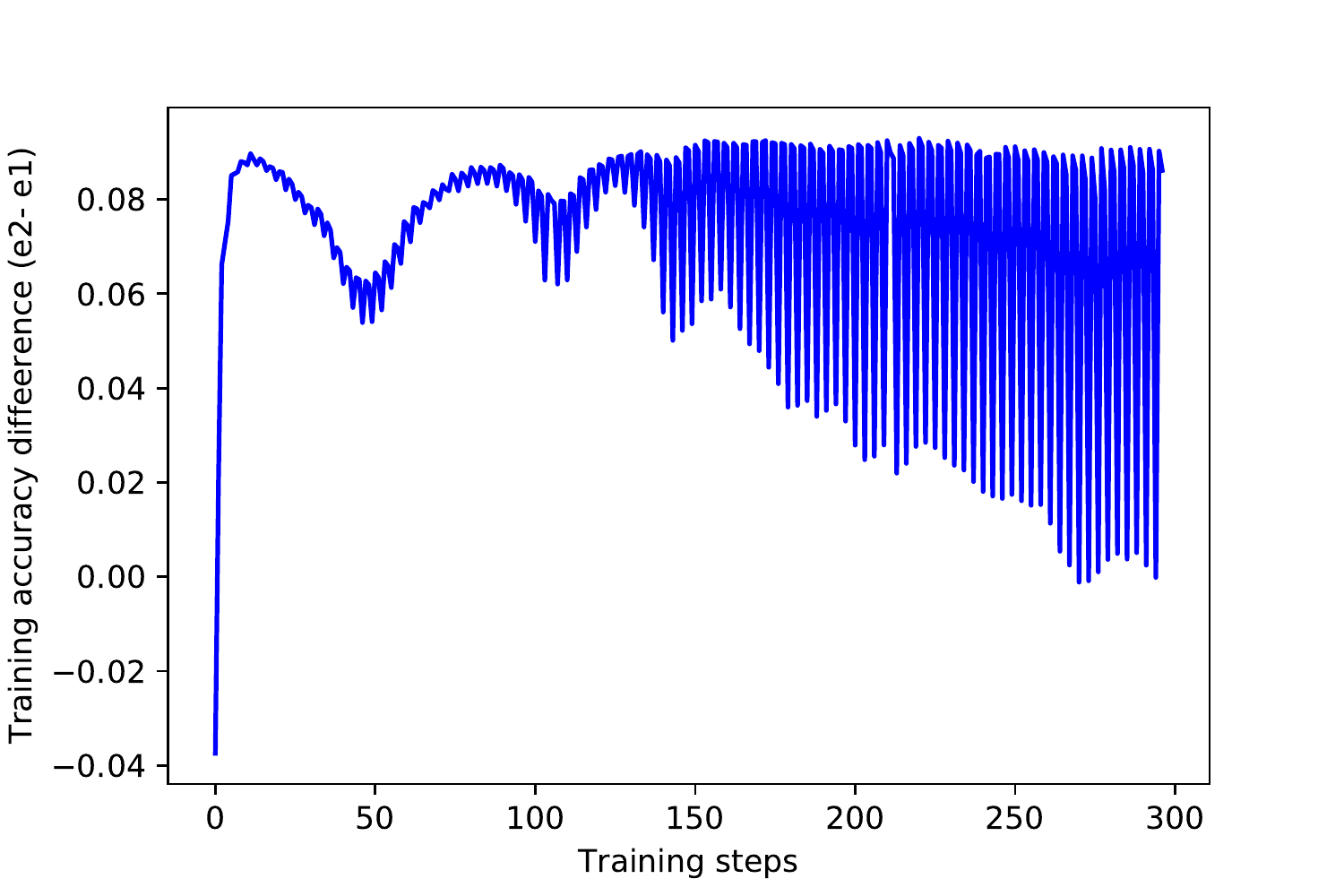}
  \caption{V-IRM Colored Fashion MNIST: Difference in accuracy of the ensemble model between the two environments}
\label{figs10}
\end{figure}

\begin{figure}
\centering

  \includegraphics[width=2.5in]{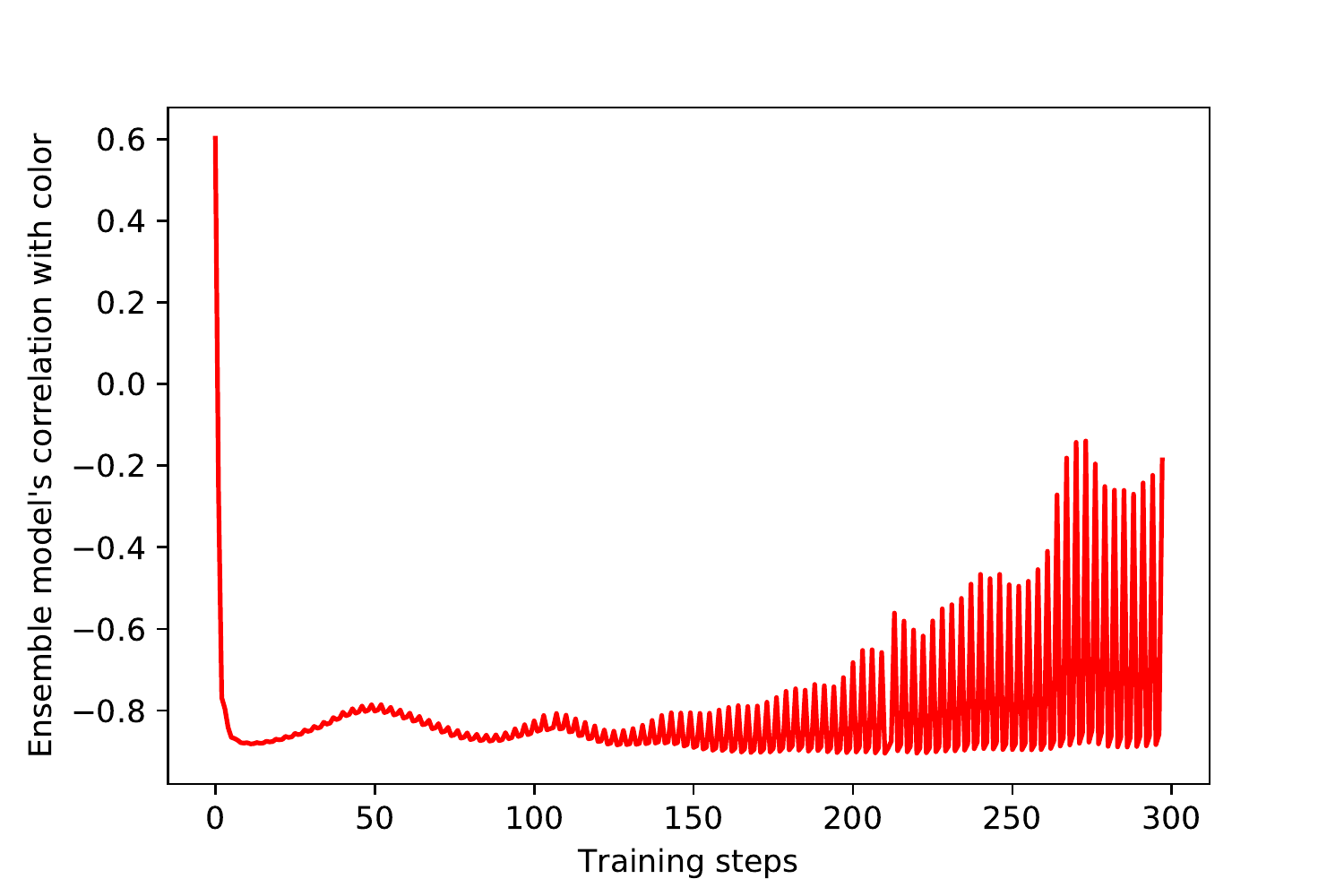}
  \caption{V-IRM Colored Fashion MNIST:  Ensemble's correlation with color}

\label{figs11}
\end{figure}

\begin{figure}
\centering

  \includegraphics[width=2.5 in]{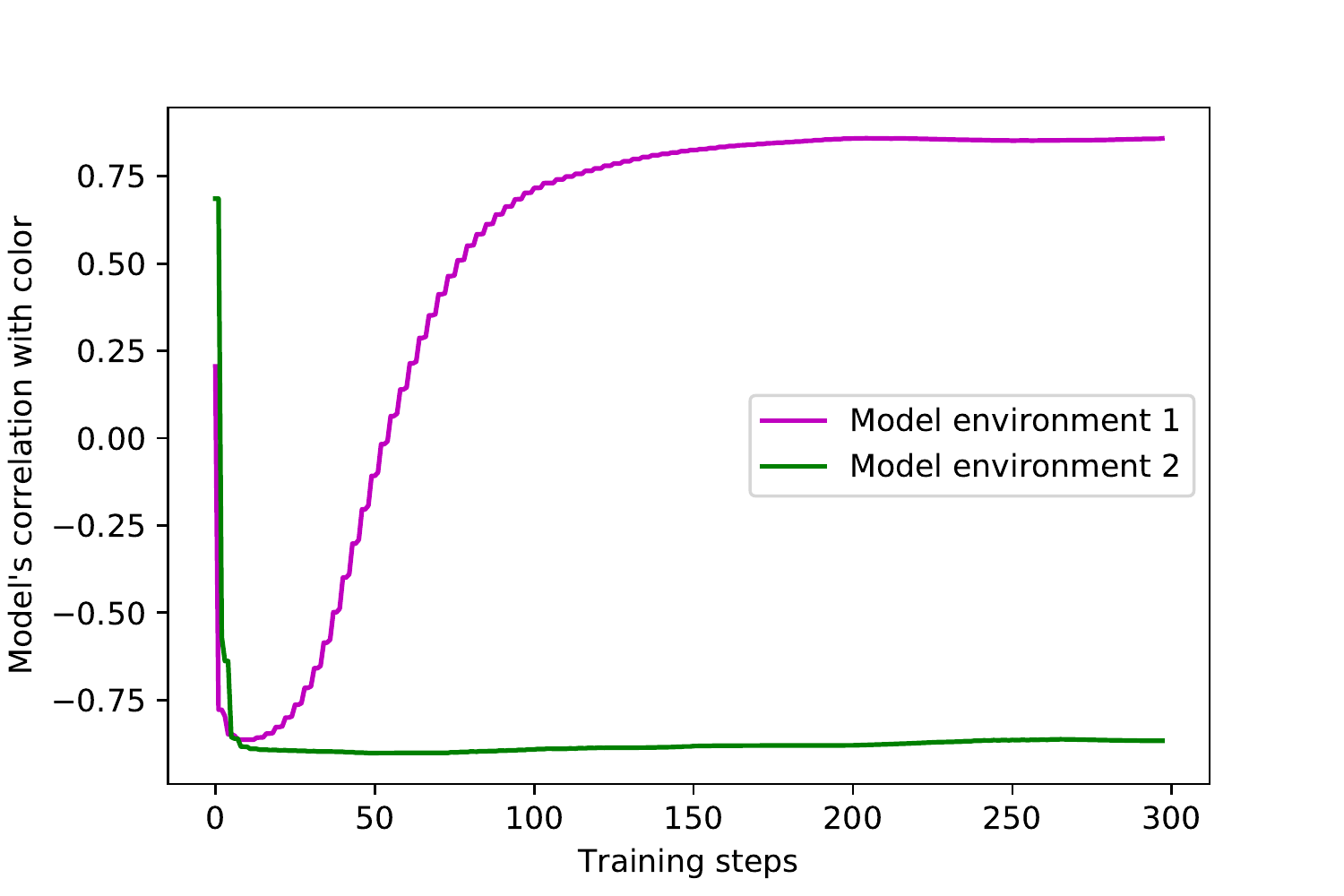}
  \caption{V-IRM Colored Fashion MNIST:  Compare individual model correlations. }

\label{figs12}
\end{figure}

\begin{figure}
\centering

  \includegraphics[width=2.75in]{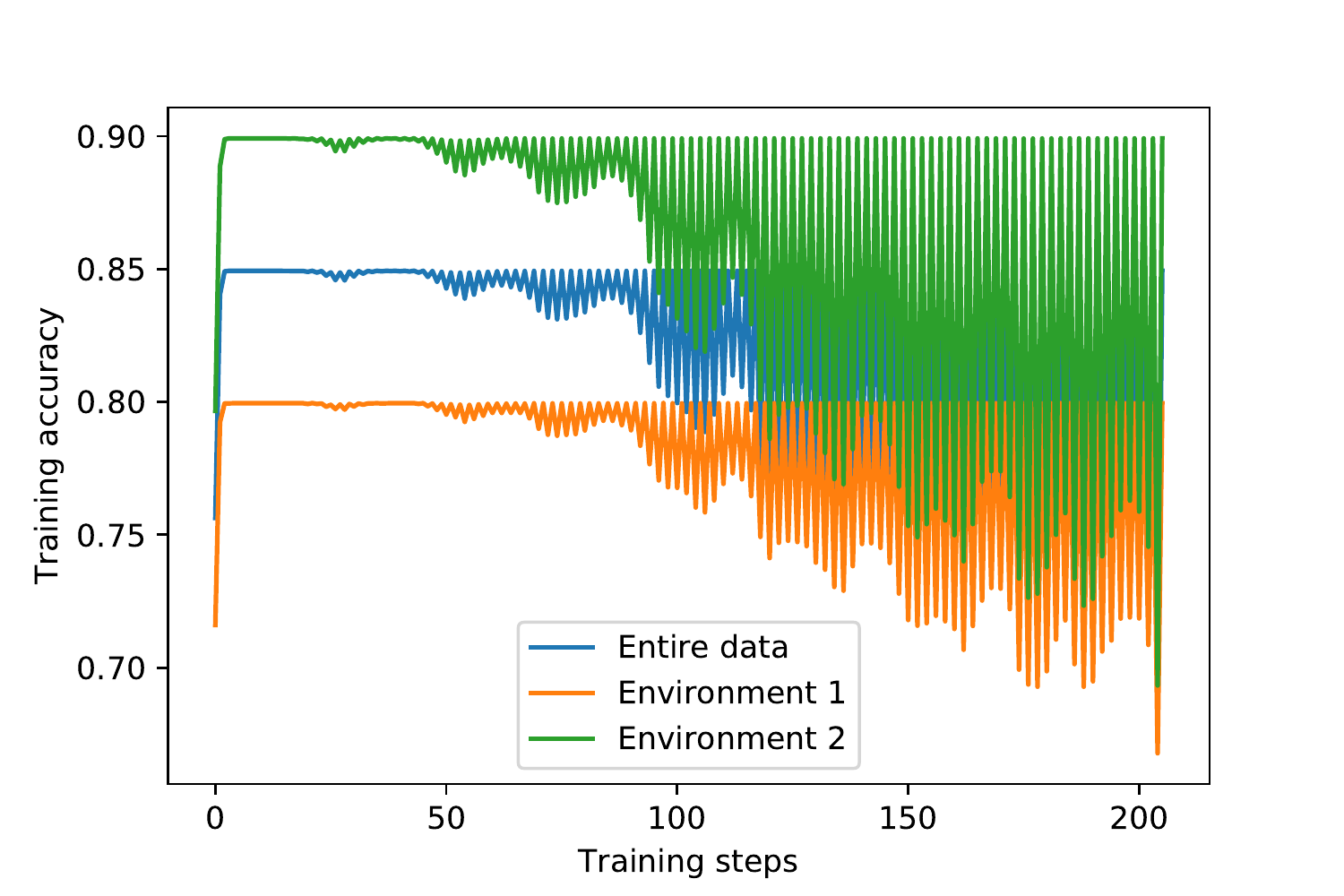}
  \caption{F-IRM Colored Digits MNIST: Comparing accuracy of ensemble}

\label{figs13}
\end{figure}

\begin{figure}
\centering

  \includegraphics[width=2.75 in]{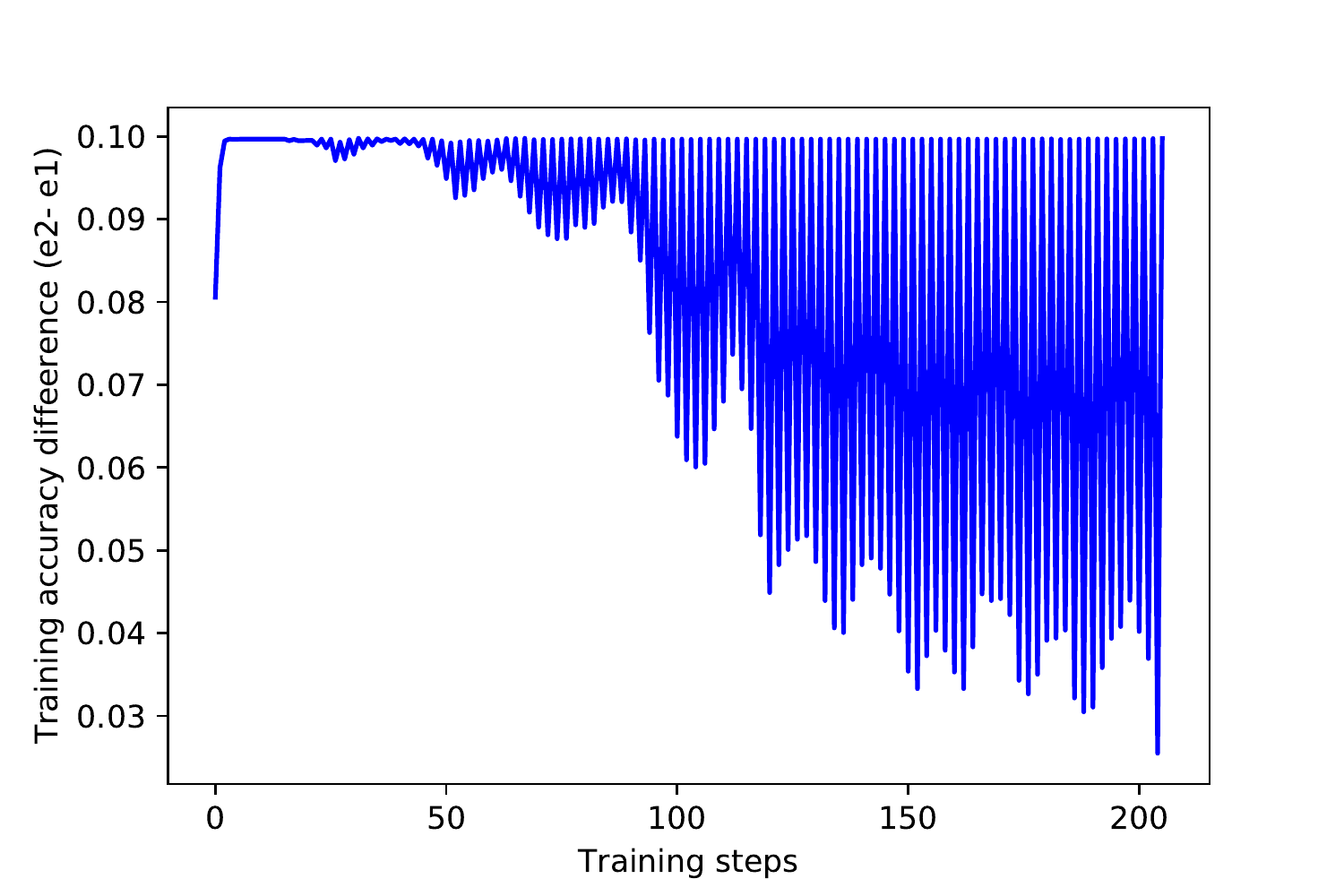}
  \caption{F-IRM Colored Digits MNIST: Difference in accuracy of the ensemble model between the two environments}

\label{figs14}
\end{figure}

\begin{figure}
\centering
  \includegraphics[width=2.5in]{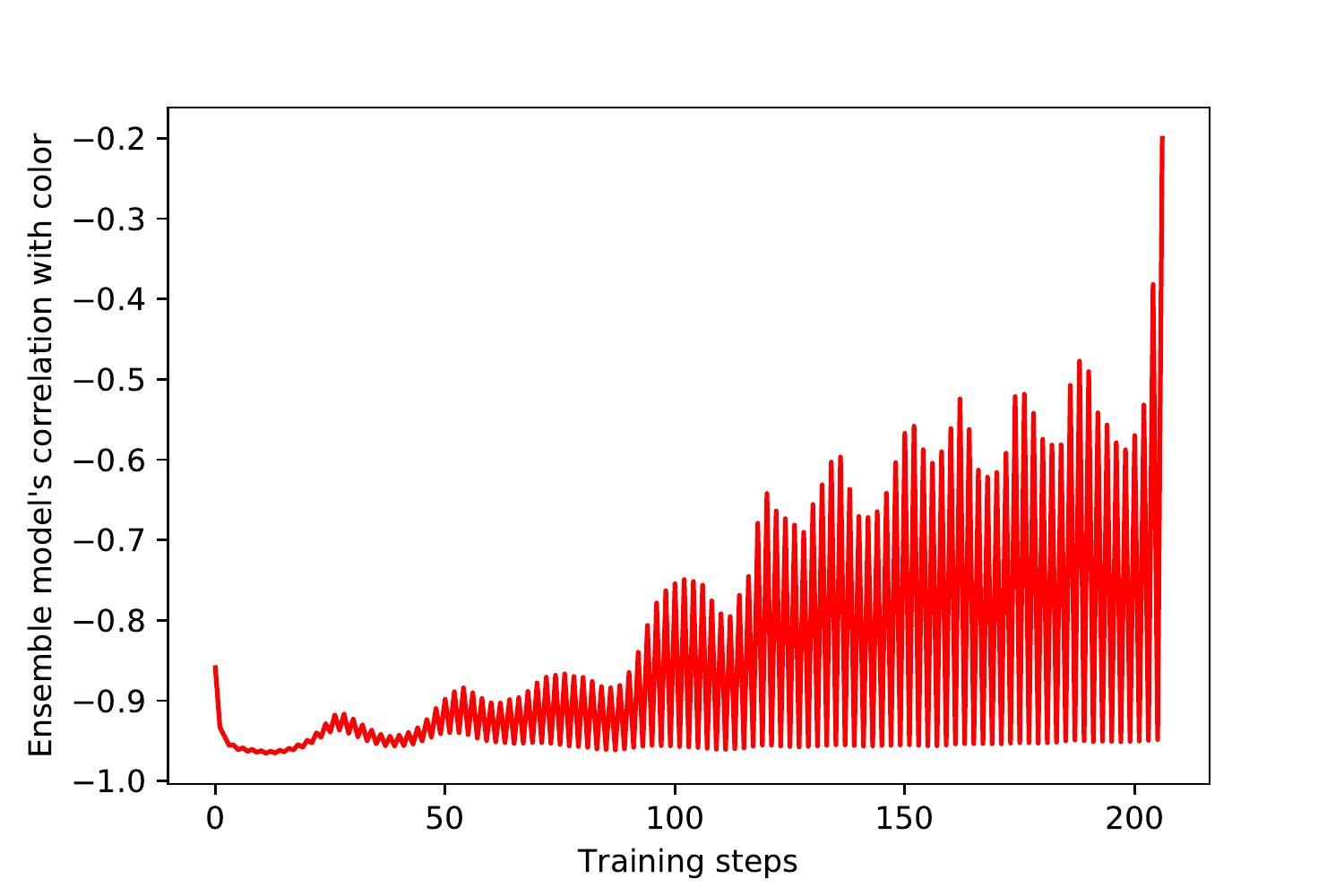}
  \caption{F-IRM Colored Digits MNIST: Ensemble's correlation with color}
\label{figs15}
\end{figure}

\begin{figure}
\centering
  \includegraphics[width=2.5 in]{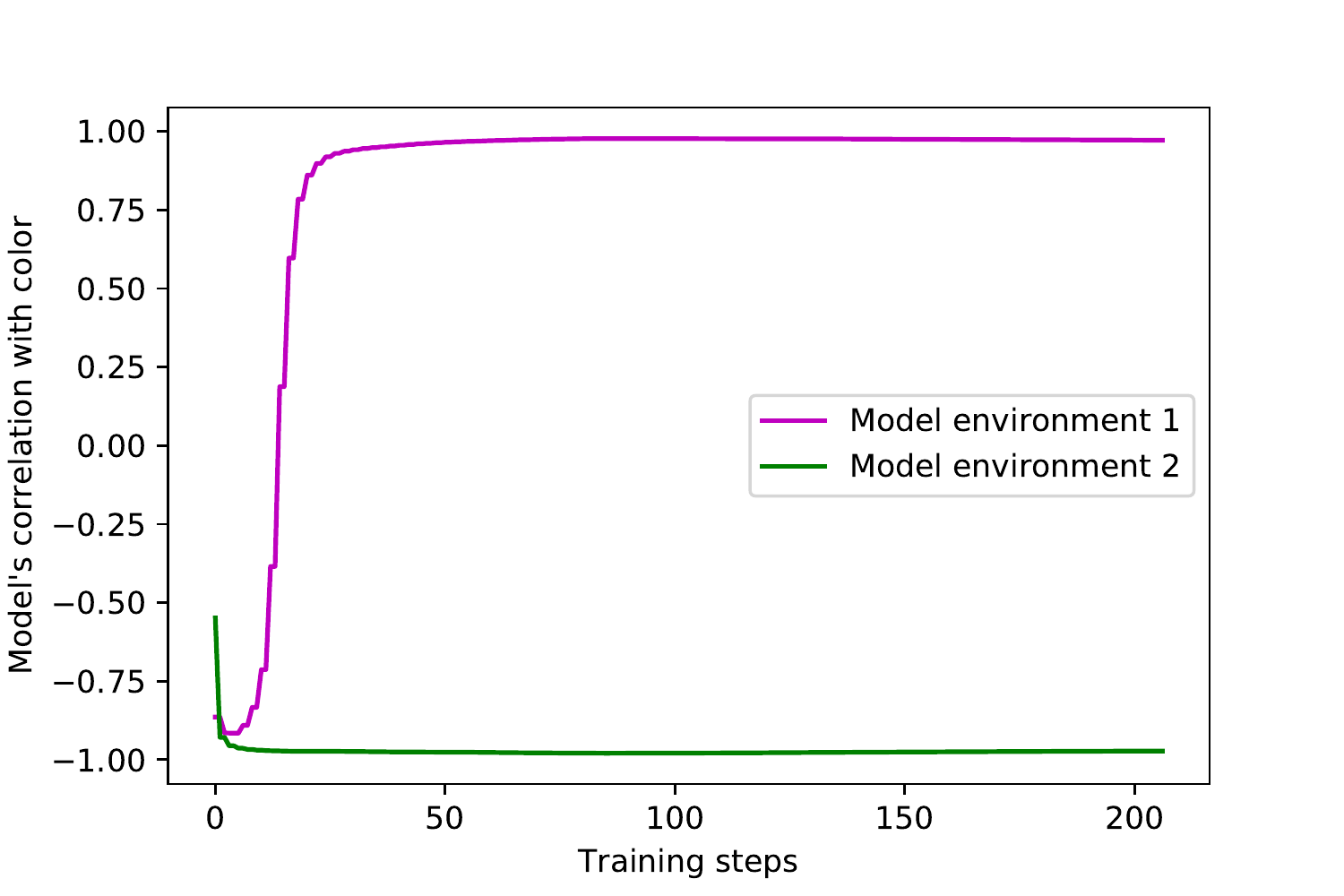}
  \caption{F-IRM Colored Digits MNIST:  Compare individual model correlations. }
\label{figs16}
\end{figure}

\begin{figure}
\centering
  \includegraphics[width=2.75in]{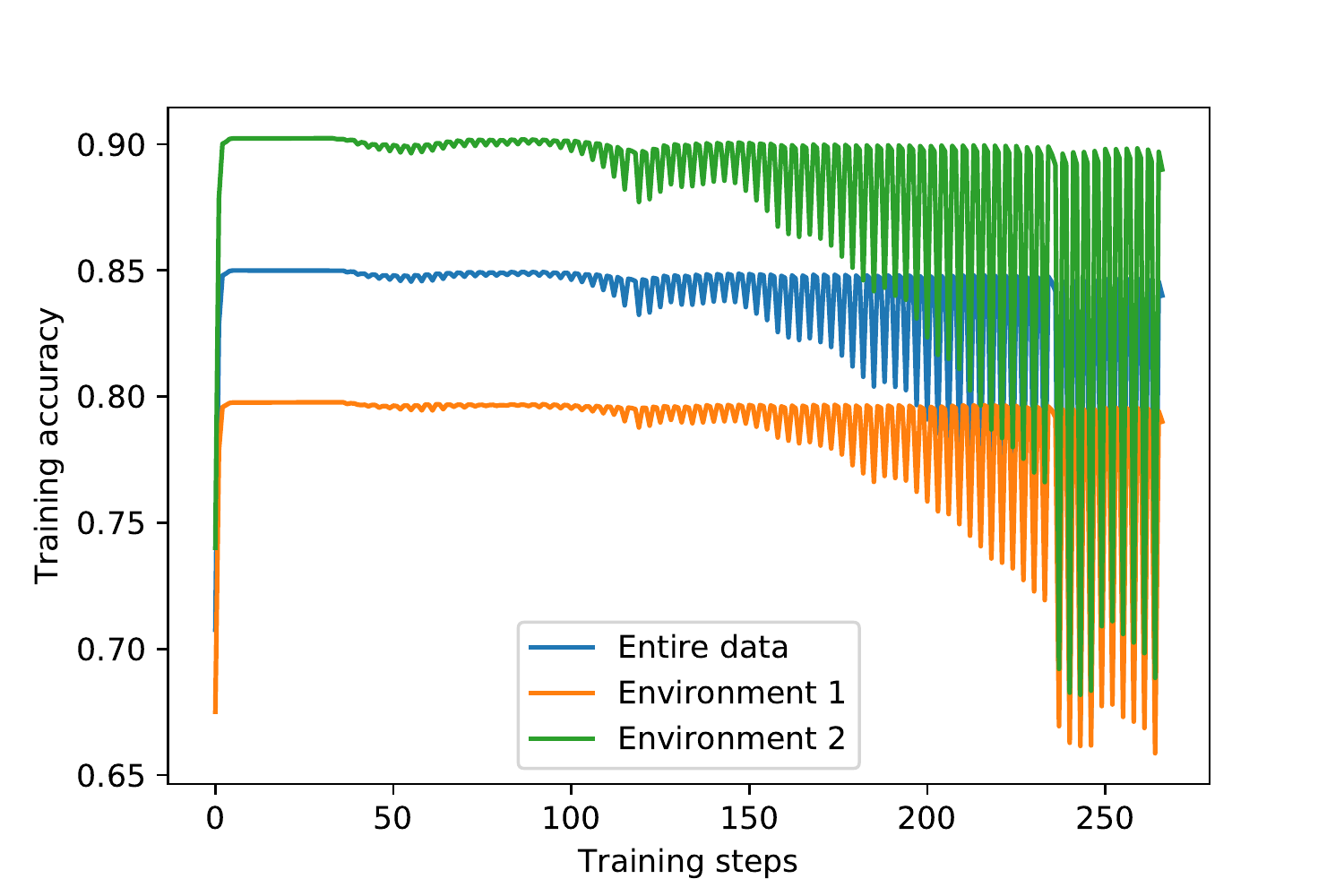}
  \caption{V-IRM Colored Digits MNIST:  Comparing accuracy of ensemble  }
\label{figs17}
\end{figure}

\begin{figure}
\centering
  \includegraphics[width=2.75 in]{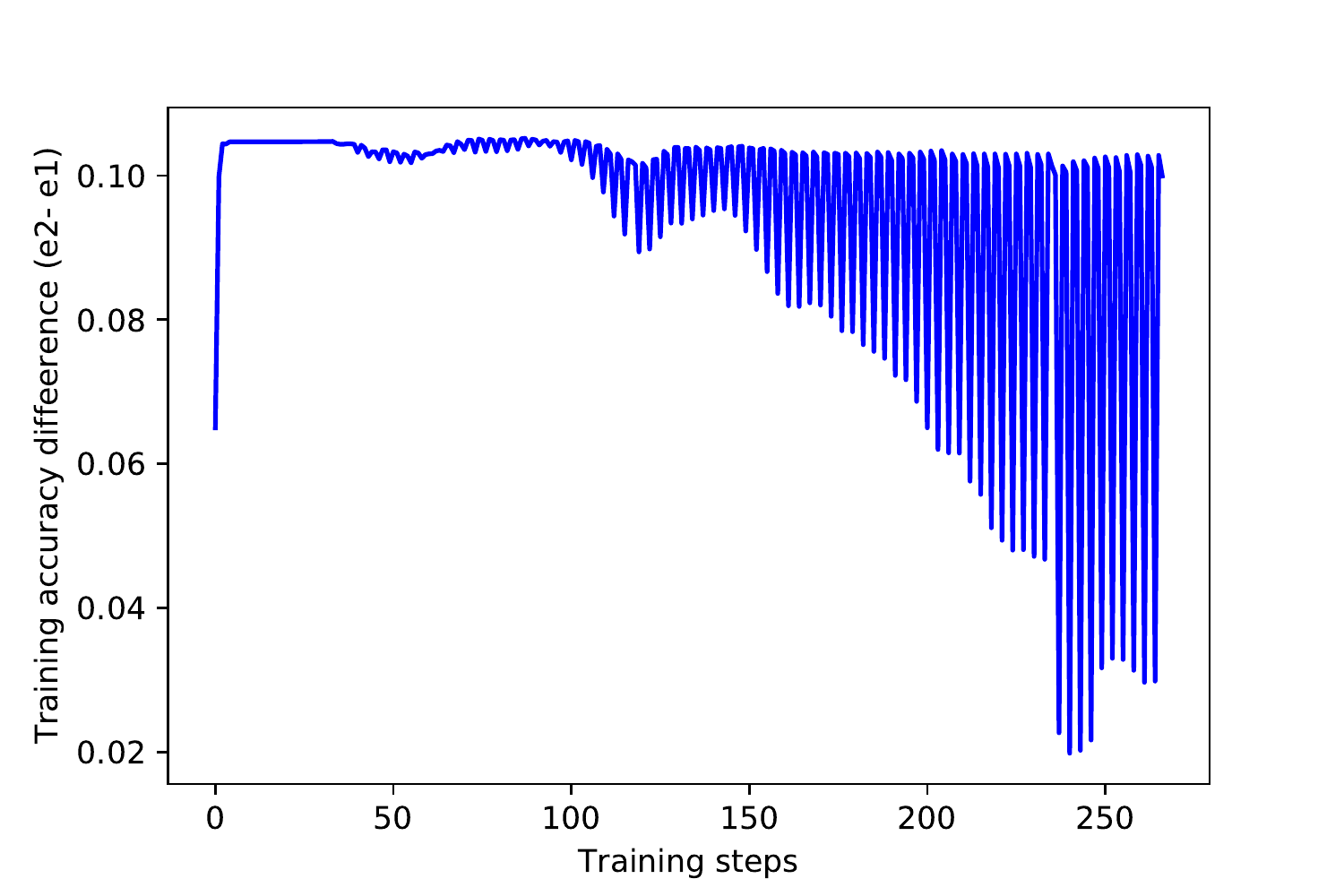}
  \caption{V-IRM Colored Digits MNIST: Difference in accuracy of the ensemble model between the two environments}

\label{figs18}
\end{figure}

\begin{figure}
\centering

  \includegraphics[width=2.5in]{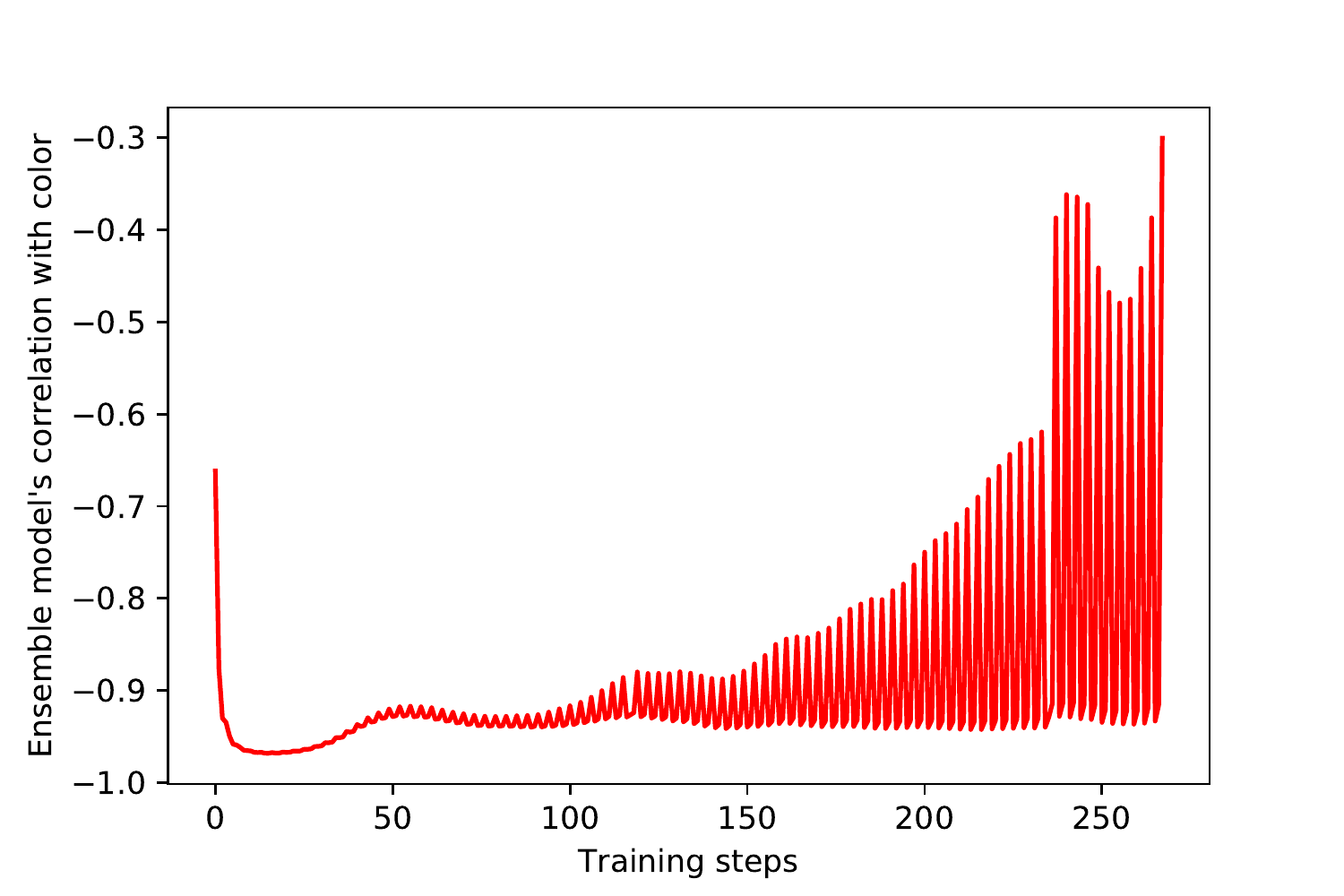}
  \caption{V-IRM Colored Digits MNIST: Ensemble's correlation with color}

\label{figs19}
\end{figure}

\begin{figure}
\centering

  \includegraphics[width=2.5 in]{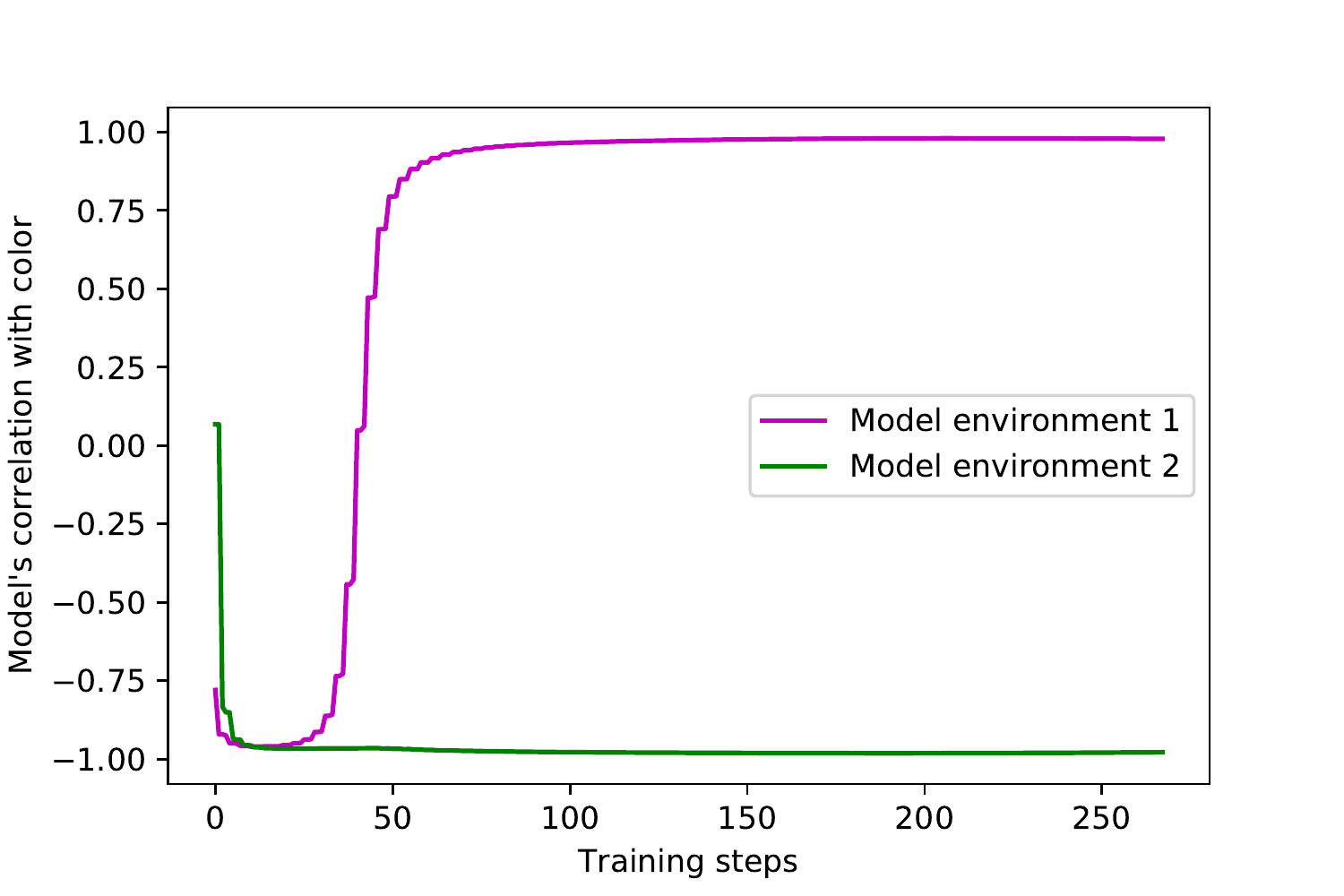}
  \caption{V-IRM Colored Digits MNIST: Compare individual model correlations }

\label{figs20}
\end{figure}

\begin{figure}
\centering

  \includegraphics[width=2.75in]{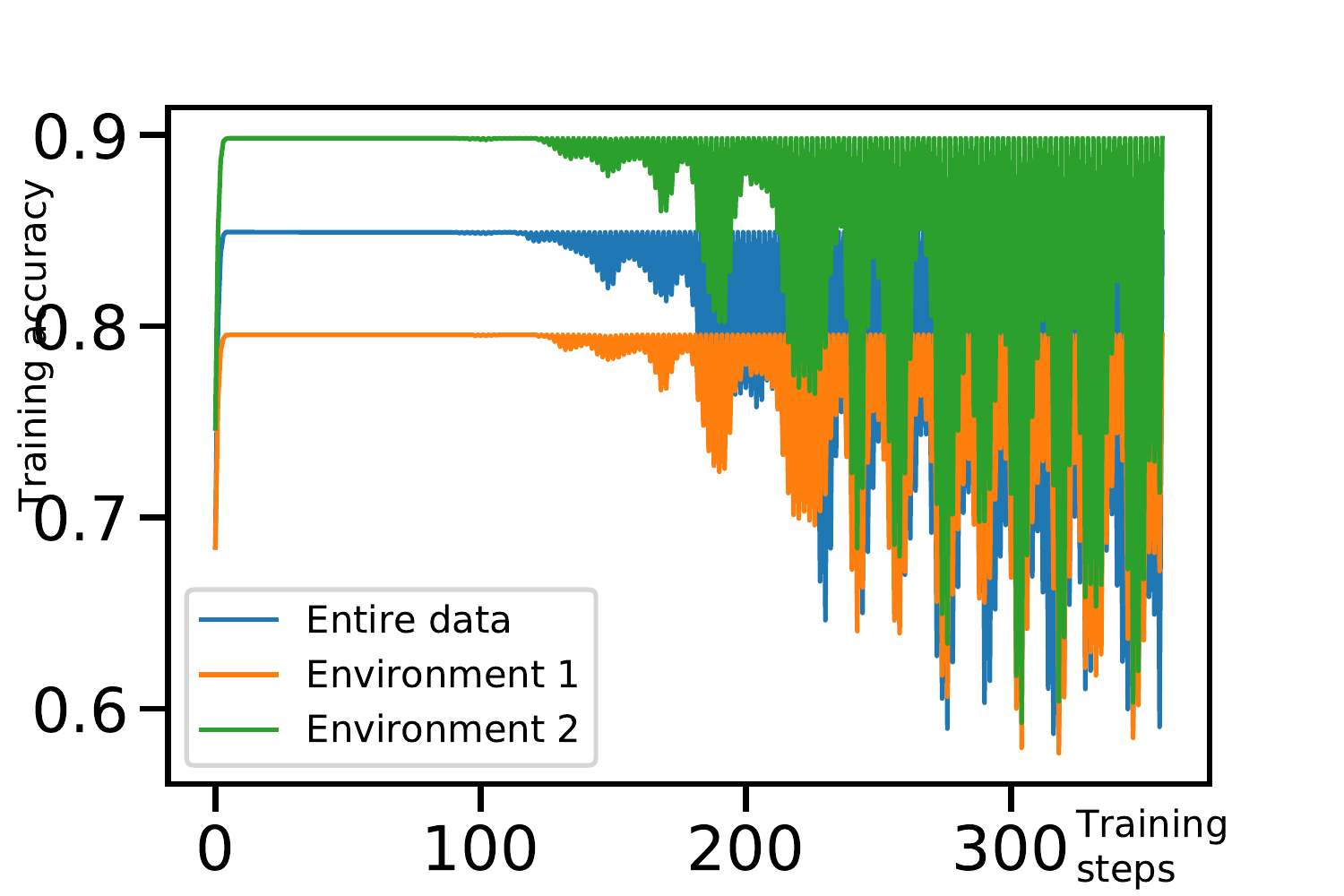}
  \caption{F-IRM Colored Desprites: Comparing accuracy of ensemble}

\label{figs21}
\end{figure}

\begin{figure}
\centering

  \includegraphics[width=2.75 in]{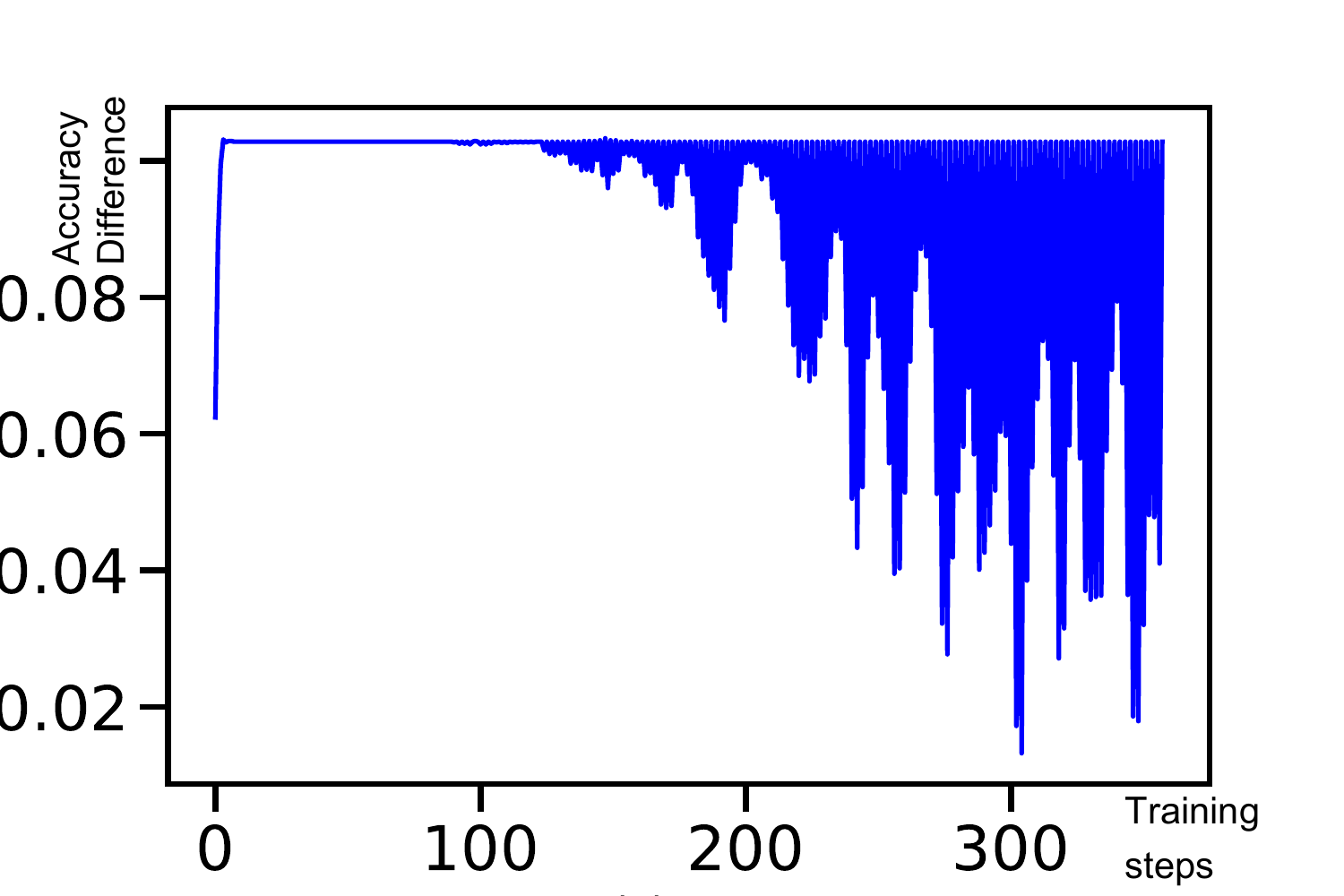}
  \caption{F-IRM Colored Desprites: Difference in accuracy of the ensemble model between the two environments}

\label{figs22}
\end{figure}

\begin{figure}
\centering

  \includegraphics[width=2.5in]{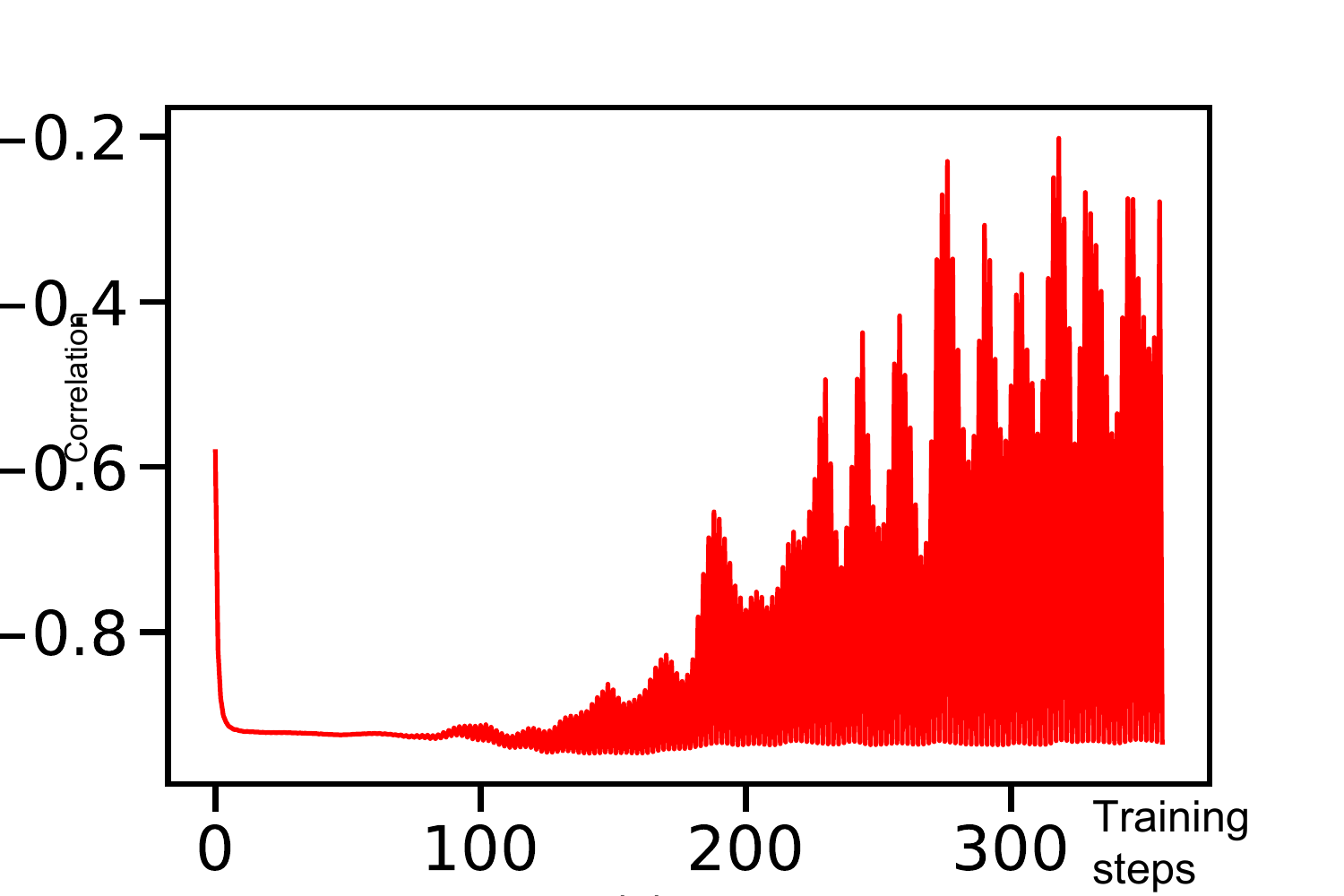}
  \caption{F-IRM Colored Desprites: Ensemble's correlation with color}

\label{figs23}
\end{figure}

\begin{figure}
\centering

  \includegraphics[width=2.5 in]{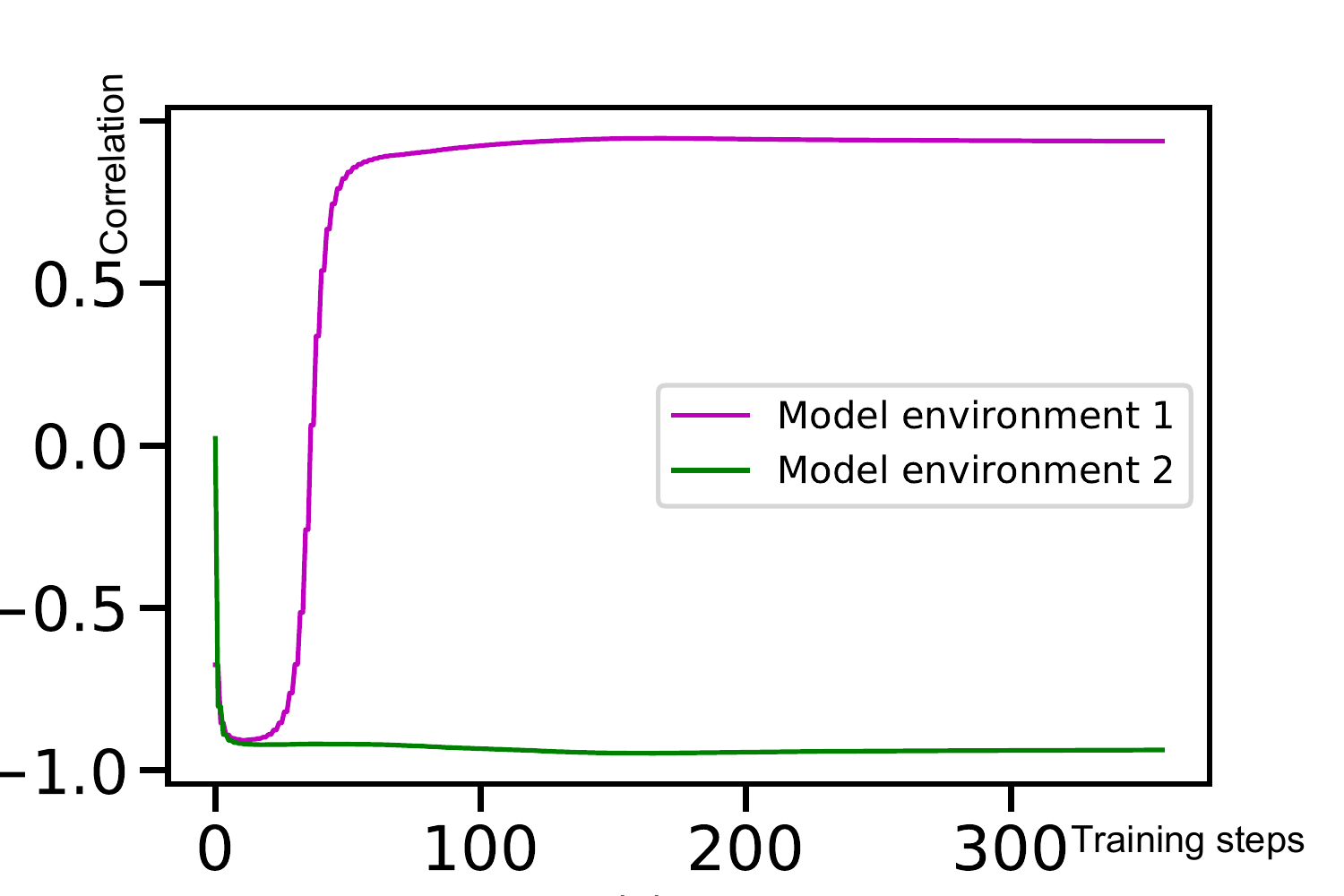}
  \caption{F-IRM Colored Desprites: Compare individual model correlations  }

\label{figs24}
\end{figure}

\begin{figure}
\centering

  \includegraphics[ width=2.75in]{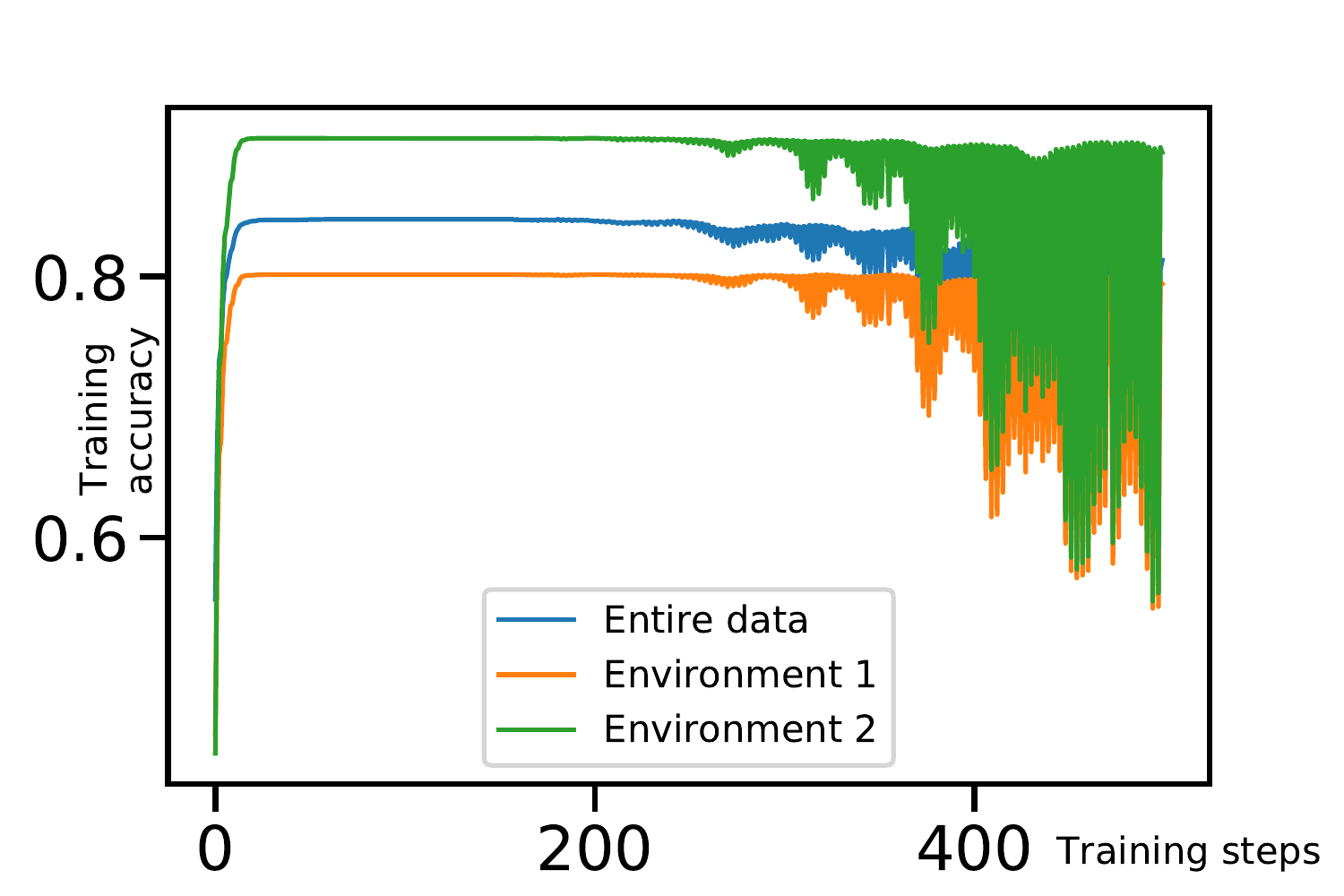}
  \caption{V-IRM Colored Desprites: Comparing accuracy of ensemble}

\label{figs25}
\end{figure}

\begin{figure}
\centering

  \includegraphics[width=2.75 in]{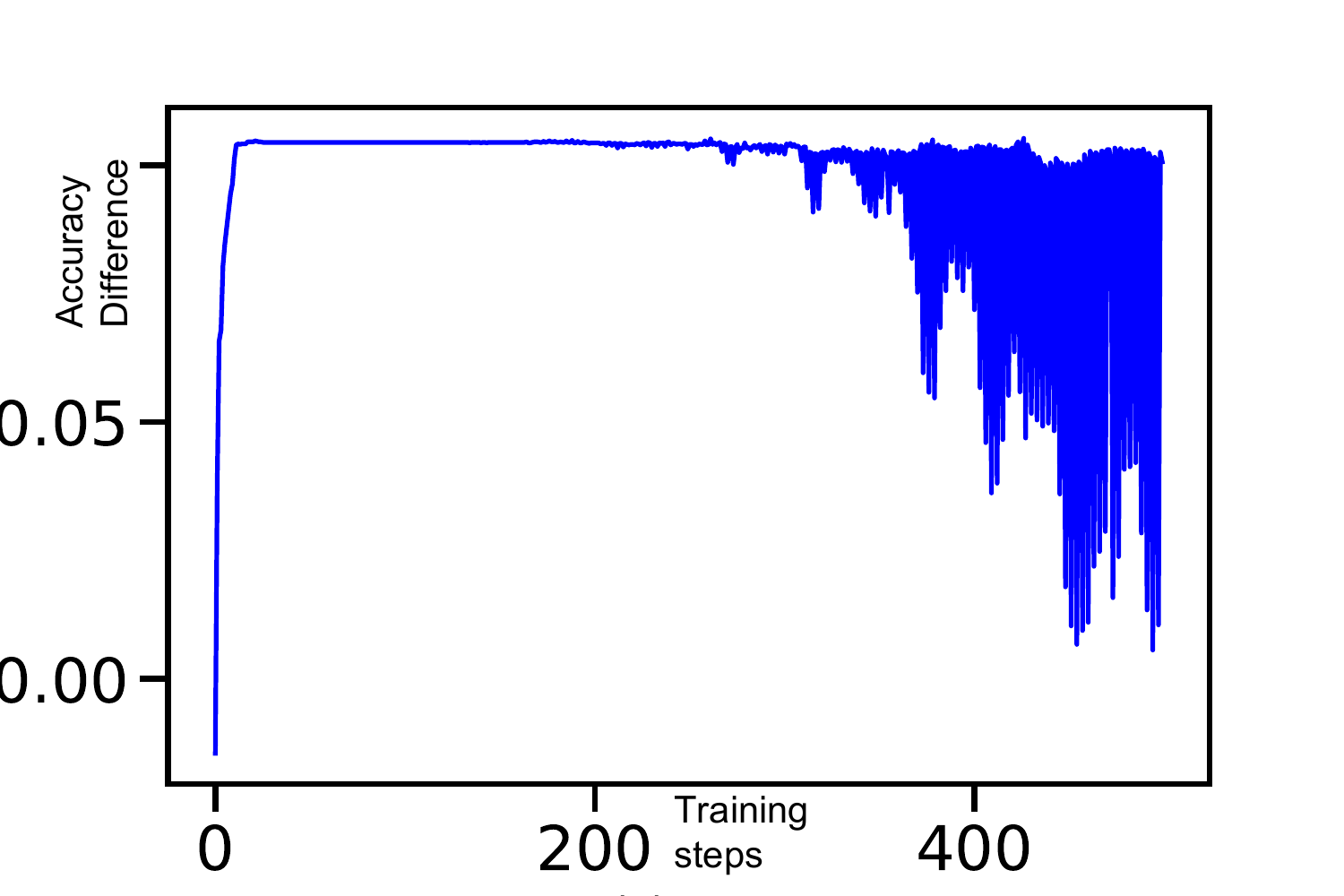}
  \caption{V-IRM Colored Desprites: Difference in accuracy of the ensemble model between the two environments}

\label{figs26}
\end{figure}

\begin{figure}
\centering
  \includegraphics[width=2.5in]{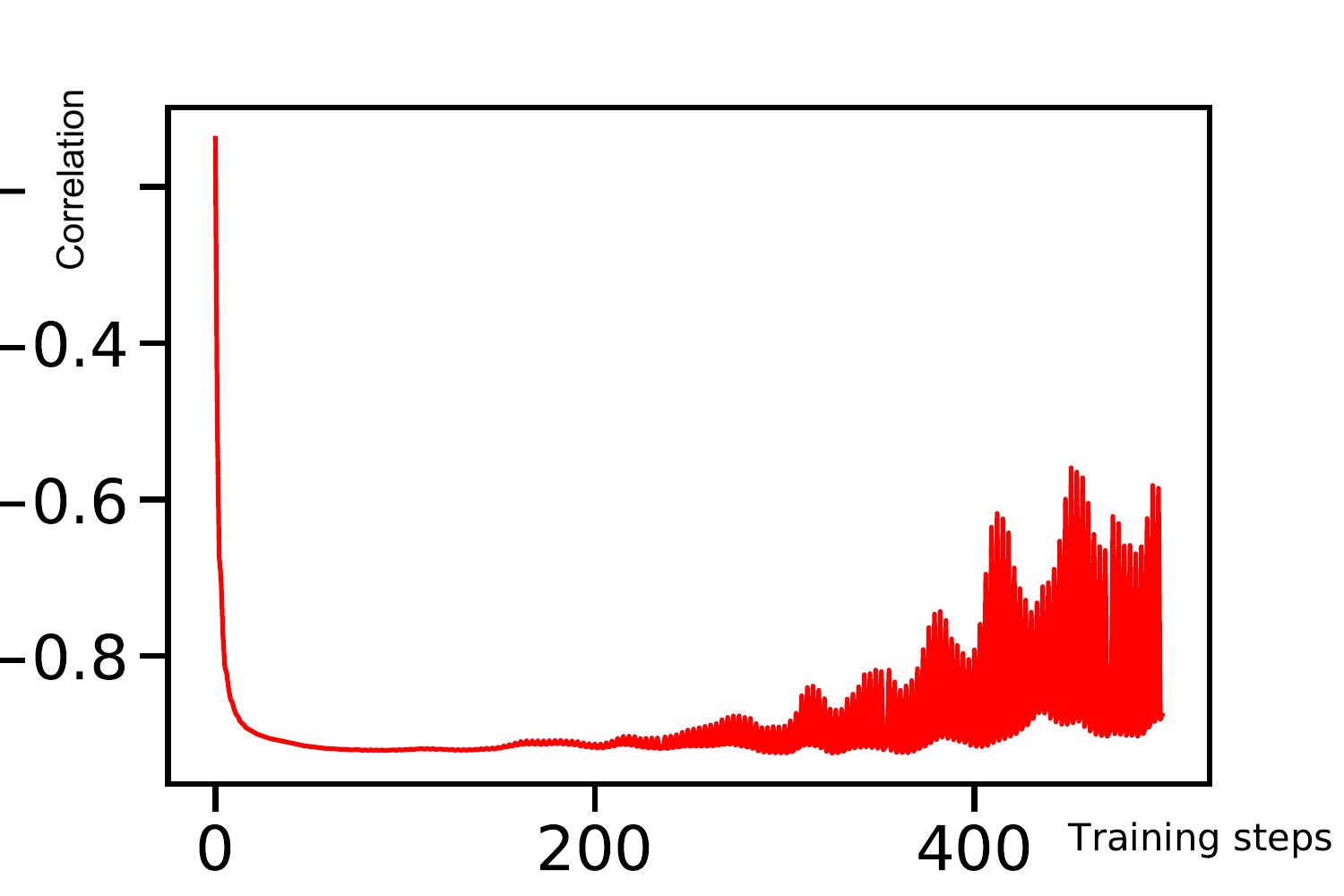}
  \caption{V-IRM Colored Desprites: Correlation of the ensemble model with color}
\label{figs27}
\end{figure}

\begin{figure}
\centering
  \includegraphics[width=2.5 in]{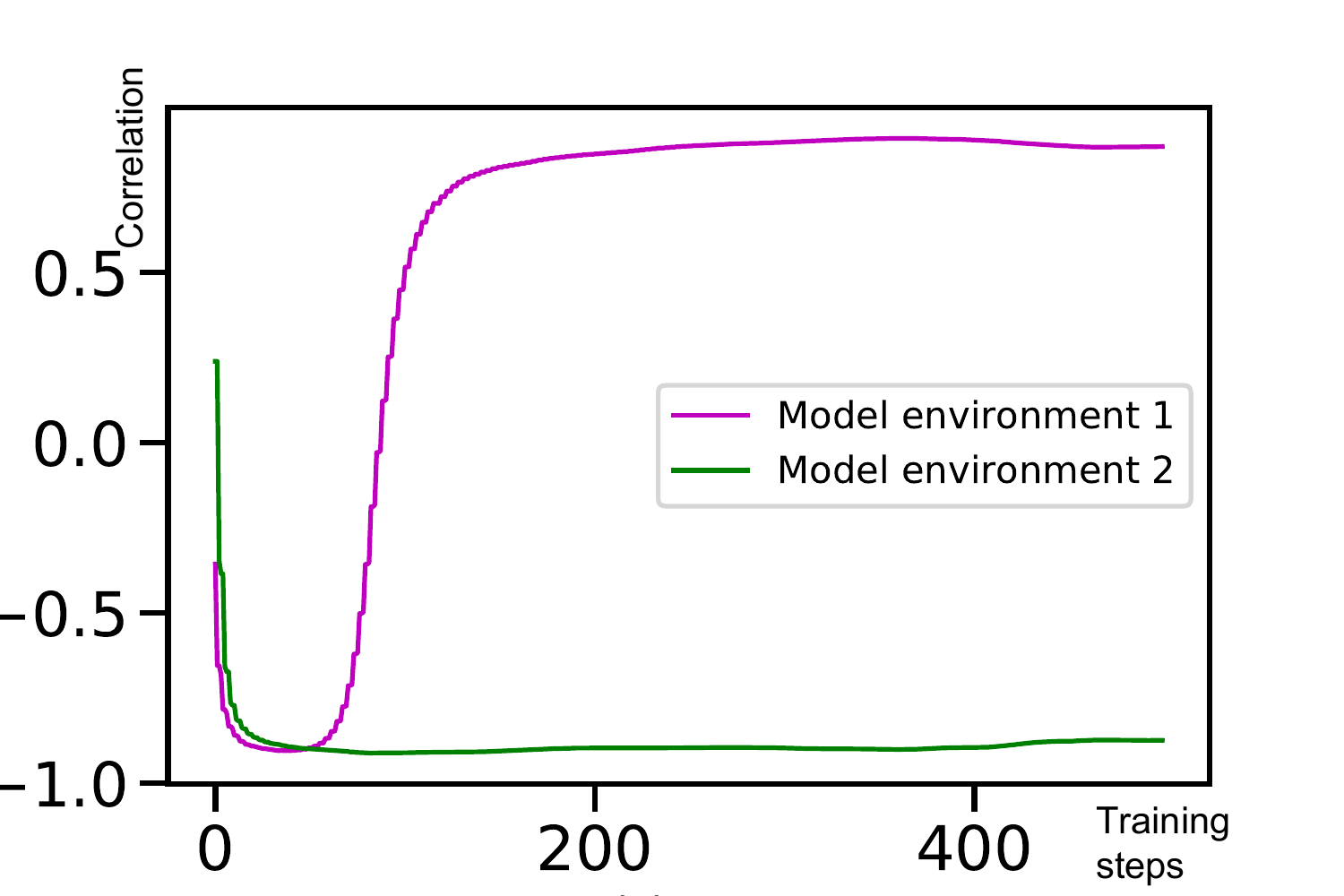}
  \caption{V-IRM Colored Desprites:  Compare individual model correlations}
\label{figs28}
\end{figure}

\begin{figure}
\centering
  \includegraphics[width=2.75in]{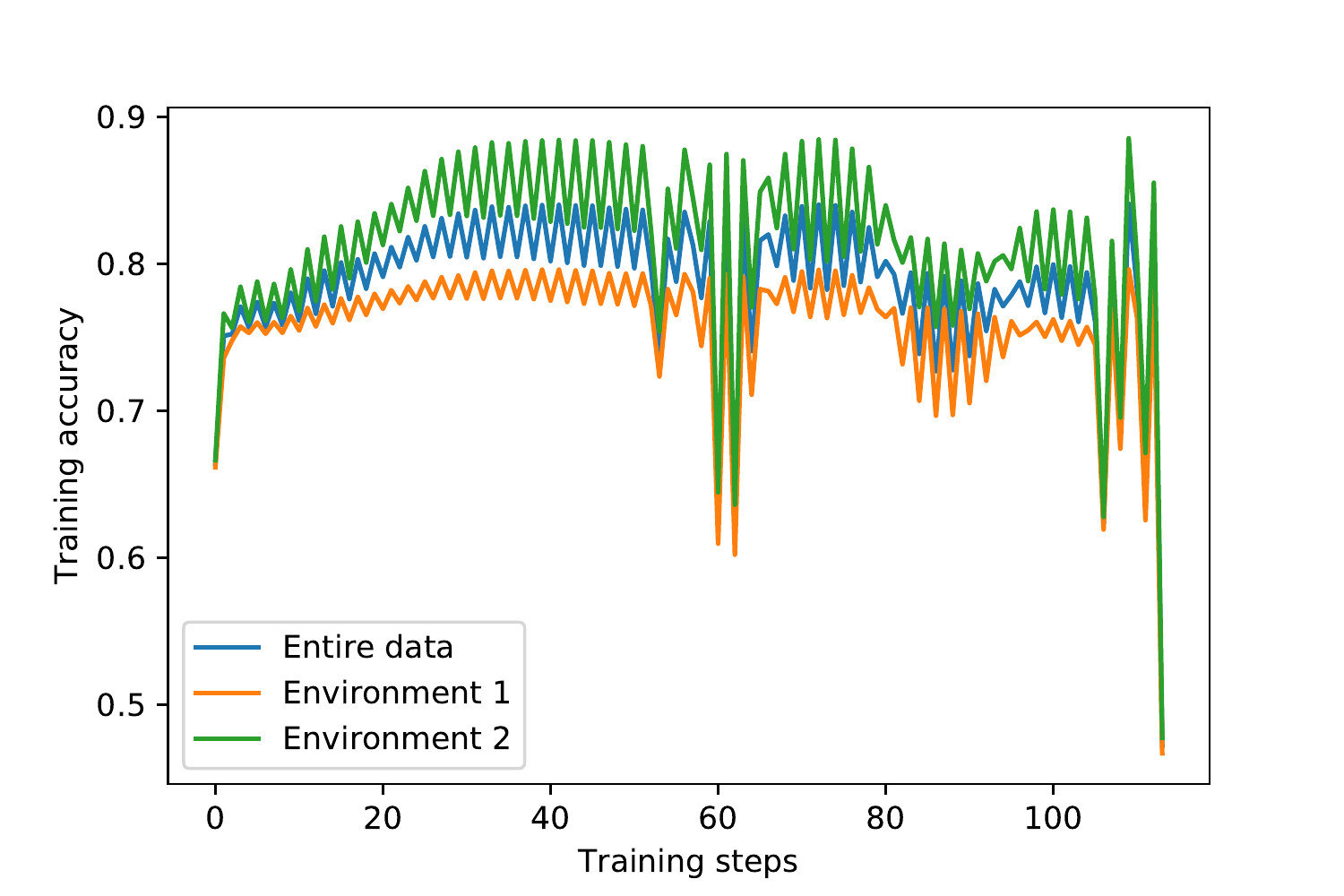}
  \caption{F-IRM Structured Noise Fashion MNIST: Comparing accuracy of ensemble }

\label{figs29}
\end{figure}

\begin{figure}
\centering
  \includegraphics[width=2.75 in]{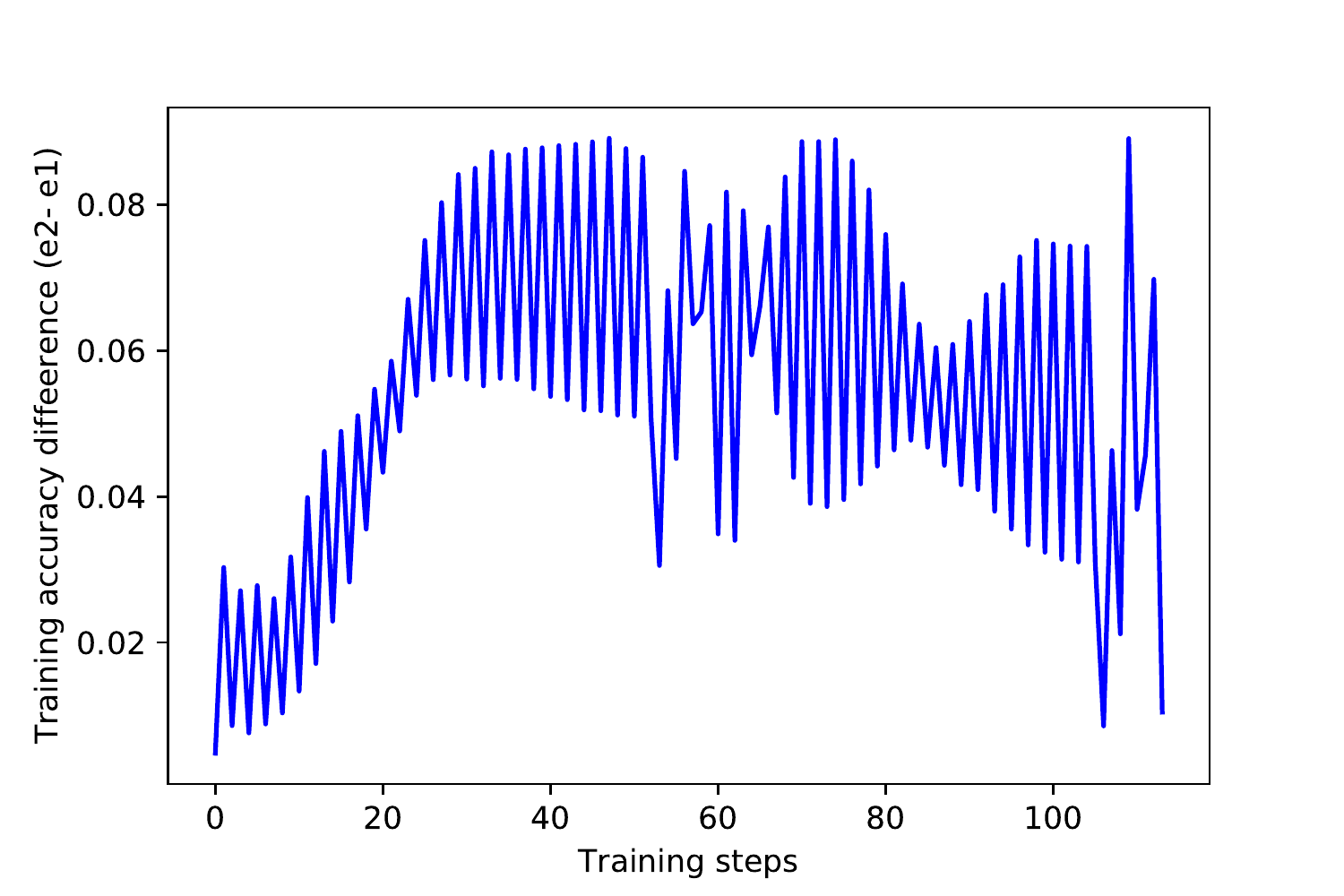}
  \caption{F-IRM Structured Noise Fashion MNIST: Difference in accuracy of the ensemble model between the two environments}
\label{figs30}
\end{figure}

\begin{figure}
\centering
  \includegraphics[ width=2.5in]{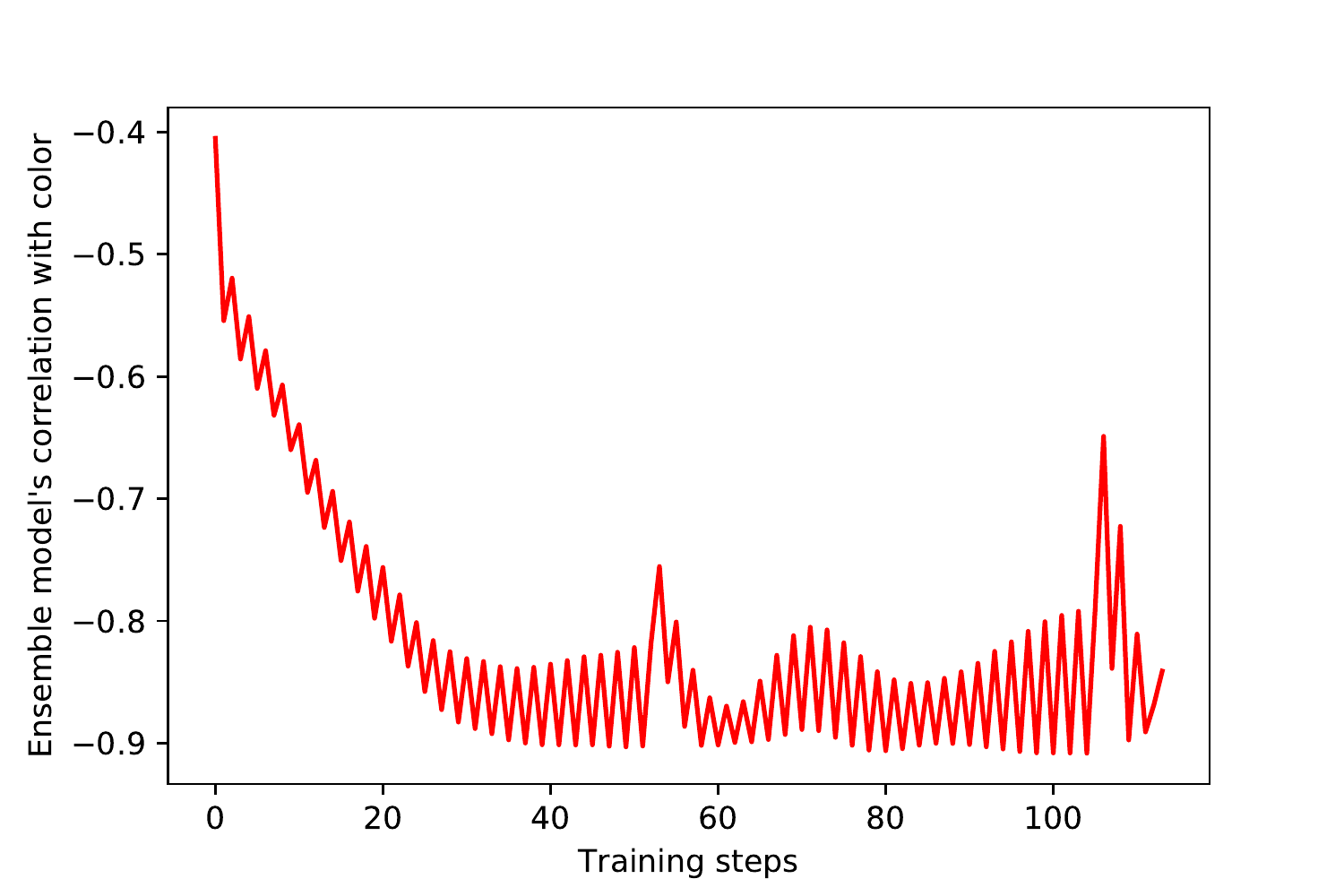}
  \caption{F-IRM Structured Noise Fashion MNIST: Correlation of the ensemble model with color}
\label{figs31}
\end{figure}

\begin{figure}
\centering
  \includegraphics[width=2.5 in]{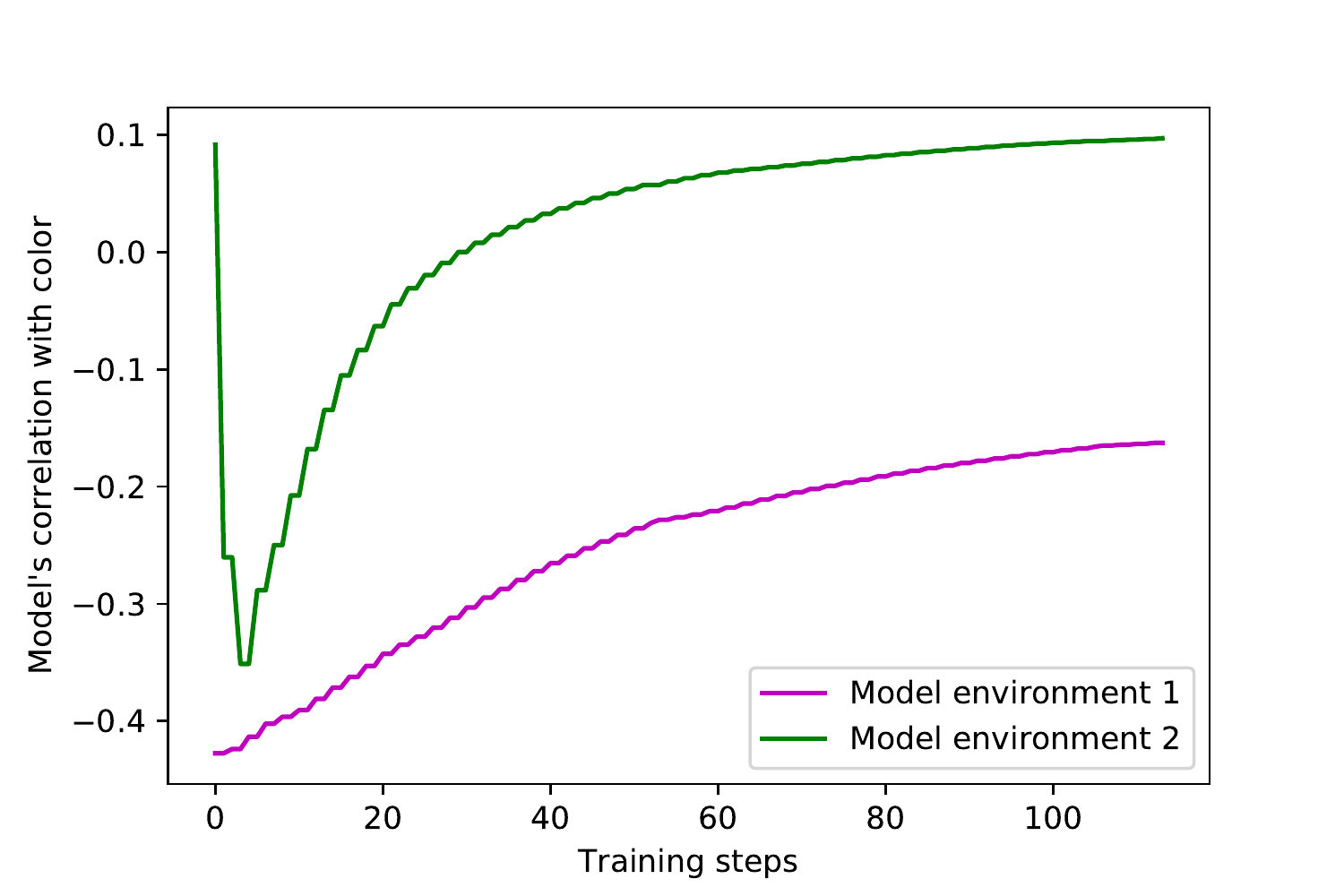}
  \caption{F-IRM Structured Noise Fashion MNIST: Individual model correlation with color}
\label{figs32}
\end{figure}

\begin{figure}
\centering
  \includegraphics[width=2.75in]{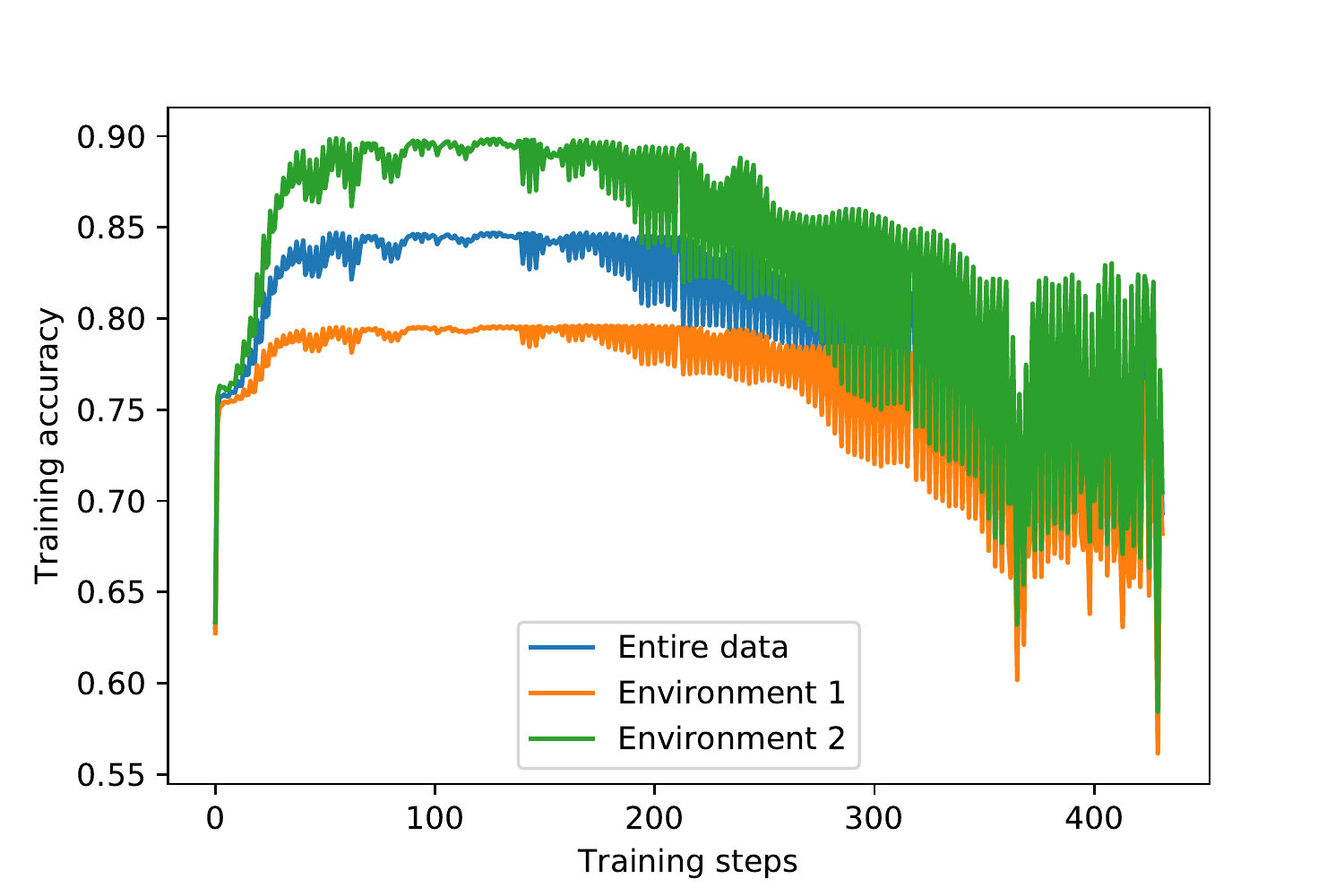}
  \caption{V-IRM Structured Noise Fashion MNIST: Comparing accuracy of ensemble }

\label{figs33}
\end{figure}

\begin{figure}
\centering

  \includegraphics[width=2.75 in]{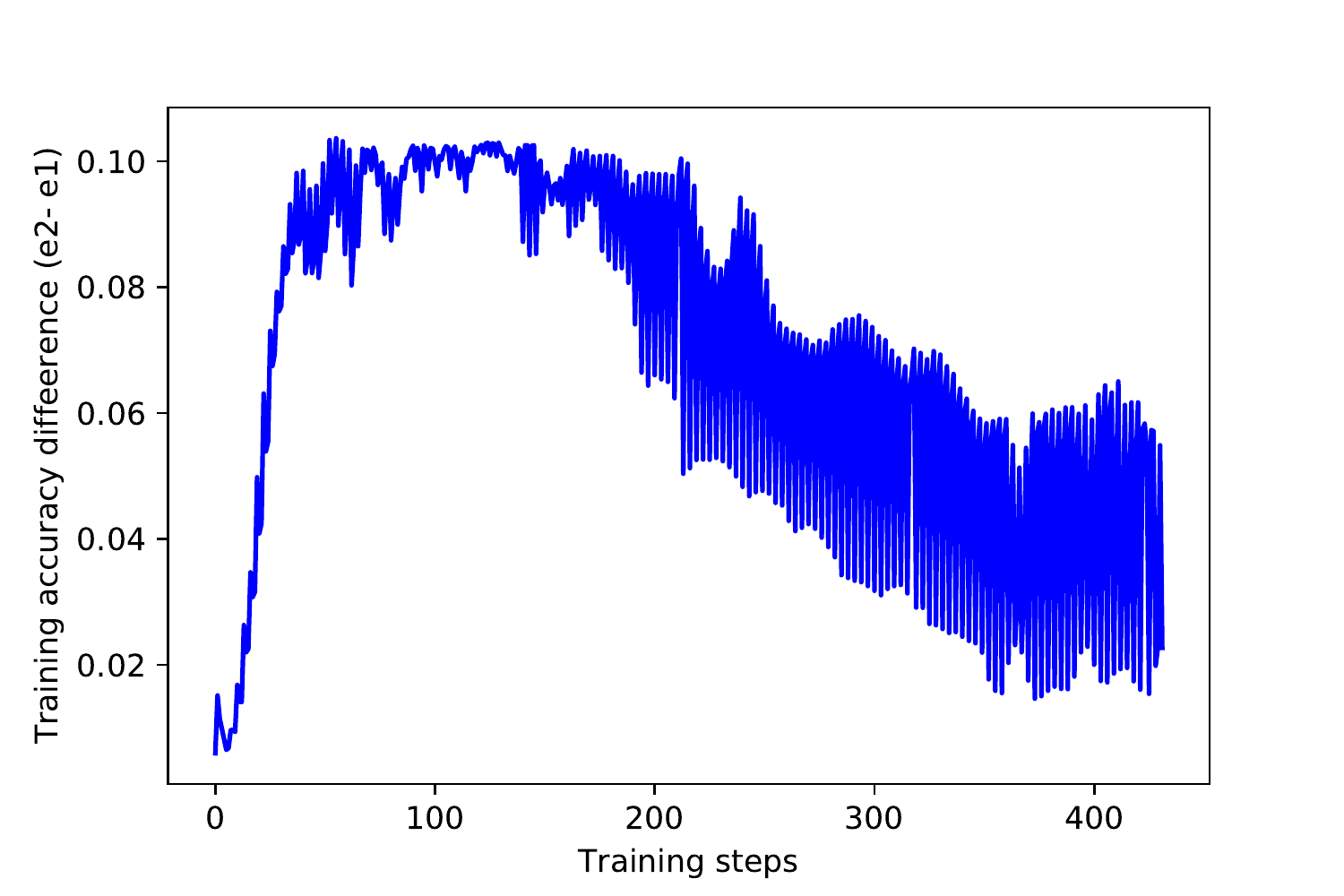}
  \caption{V-IRM Structured Noise Fashion MNIST: Difference in accuracy of the ensemble model between the two environments, }
\label{figs34}
\end{figure}

\begin{figure}
\centering
  \includegraphics[ width=2.75in]{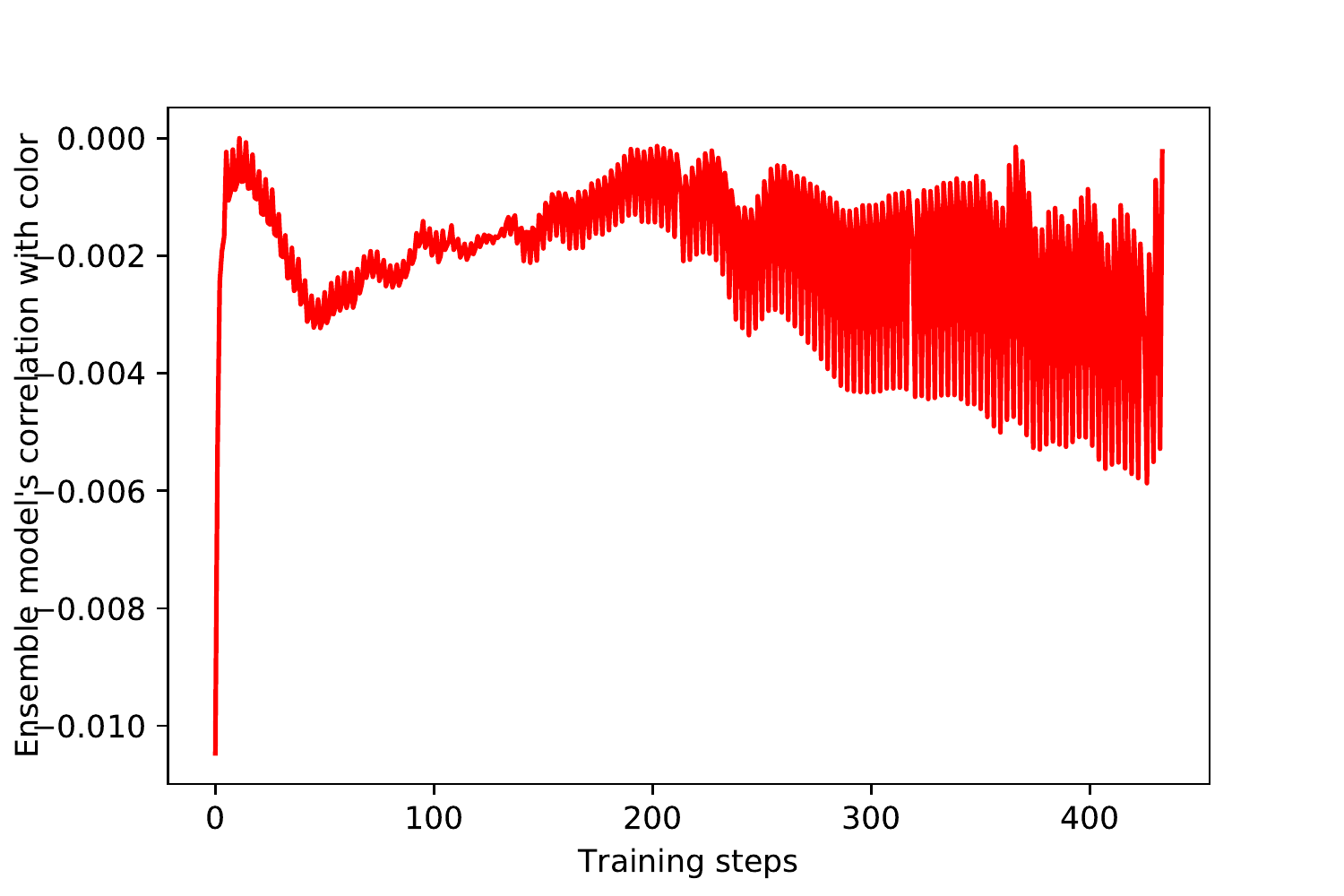}
  \caption{V-IRM Structured Noise Fashion MNIST: Ensemble's correlation with color}

\label{figs35}
\end{figure}

\begin{figure}
\centering

  \includegraphics[ width=2.5 in]{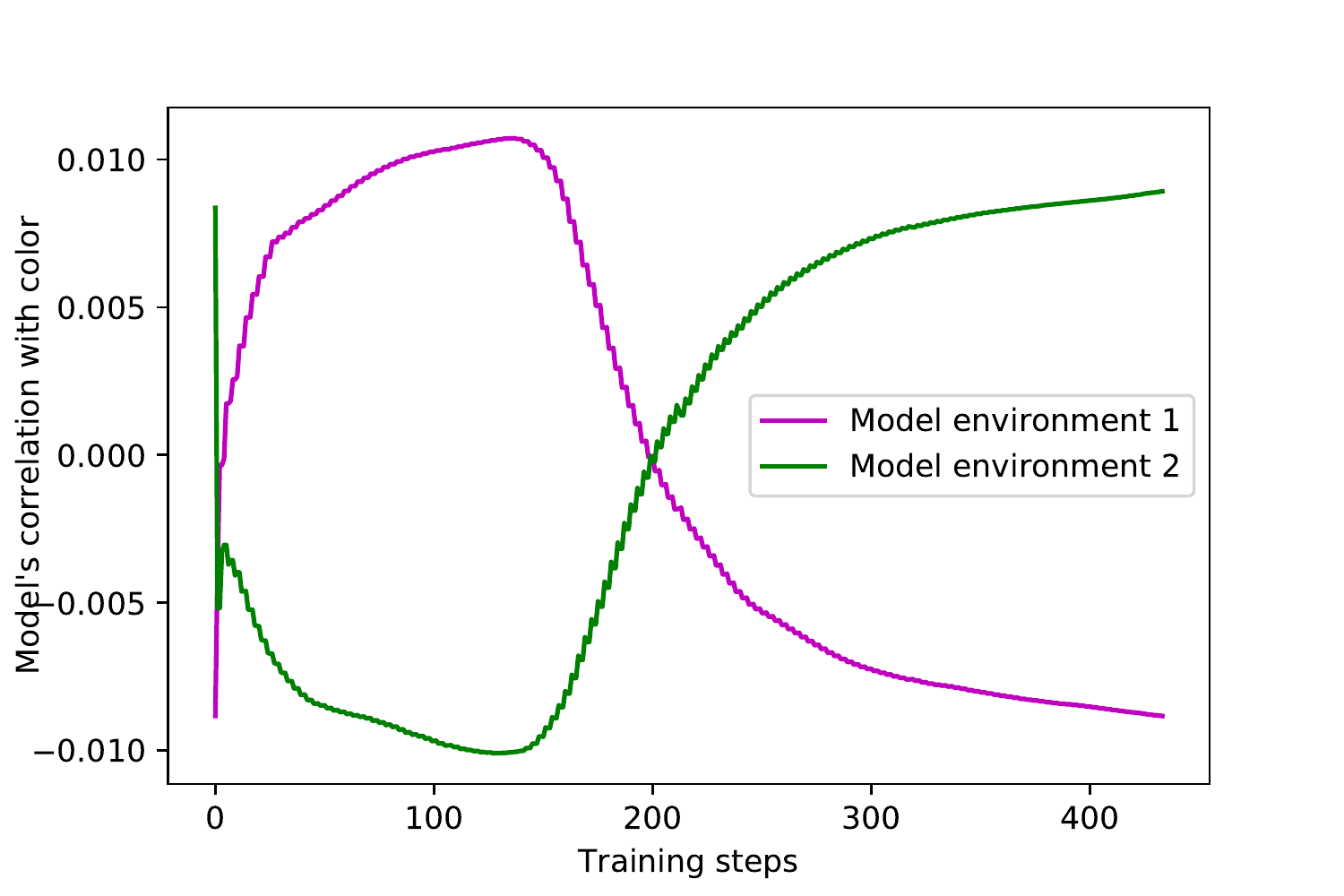}
  \caption{V-IRM Structured Noise Fashion MNIST:  Individual model correlation with color}

\label{figs36}
\end{figure}

In this section, we provide the figures for all the datasets and for both V-IRM and F-IRM game.  In Figure 2-4 in the Experiments Section,  we let each model in its turn use ltr (ltr=5) SGD step updates before the turn of the next model. We show the figure with ltr=5 to visually illustrate the oscillations better. In our experiments (Table 1-4) we set ltr =1; we show the figures corresponding to all our experiments (Table 1-4) in Figure \ref{figs5}-\ref{figs36}. The captions under the plot describe the dataset and the corresponding game (F-IRM/V-IRM). 
All the plots in Figure \ref{figs5}-\ref{figs36} use the termination criteria we described in the Experiments Section. We observe the same trends that we observed and explained in Experiments Section across all the figures. 

\begin{figure}
\centering
  \includegraphics[ width=2.75in]{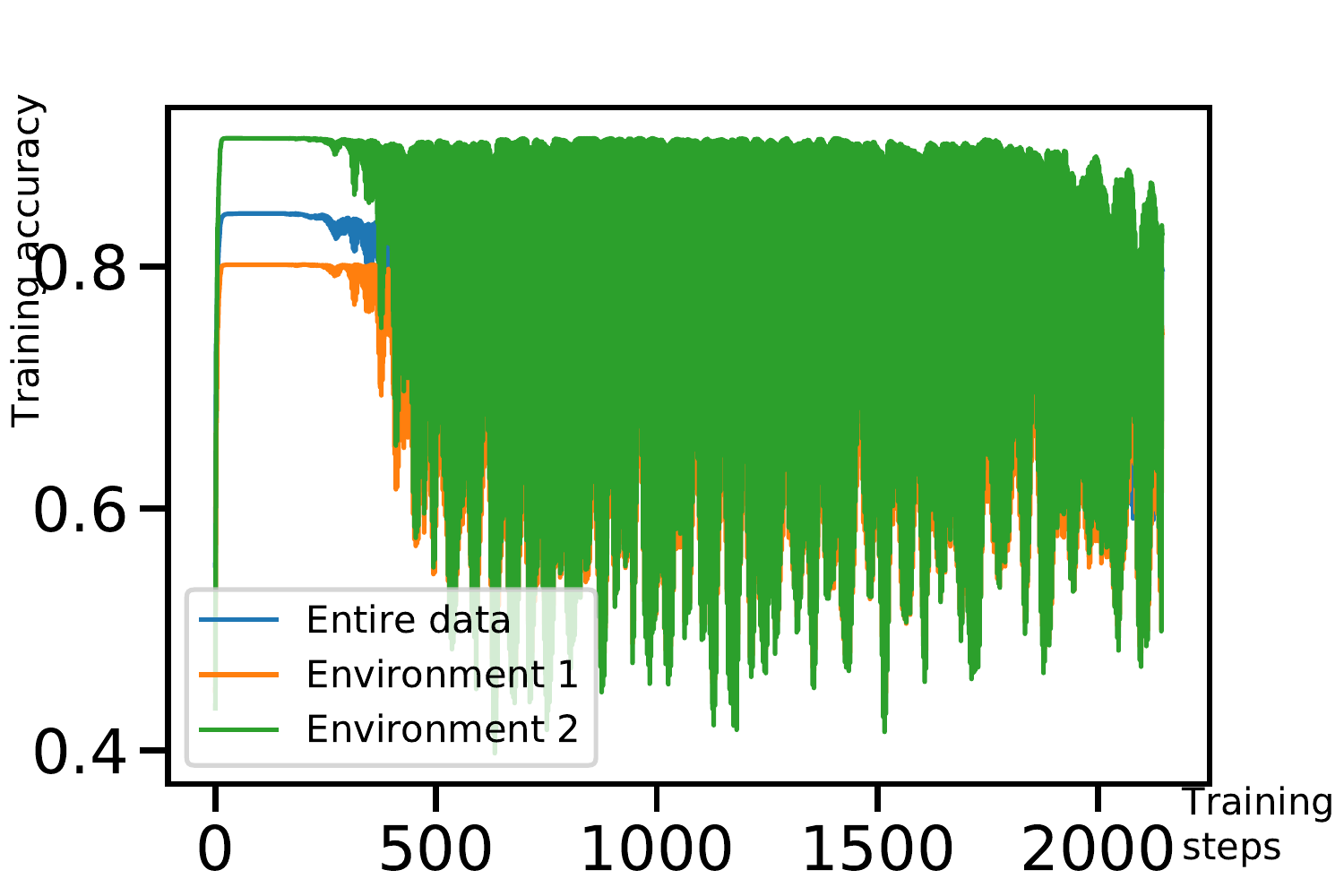}
  \caption{V-IRM Colored Desprites: Comparing accuracy of ensemble (More train steps) }

\label{figs37}
\end{figure}

\begin{figure}
\centering

  \includegraphics[width=2.75 in]{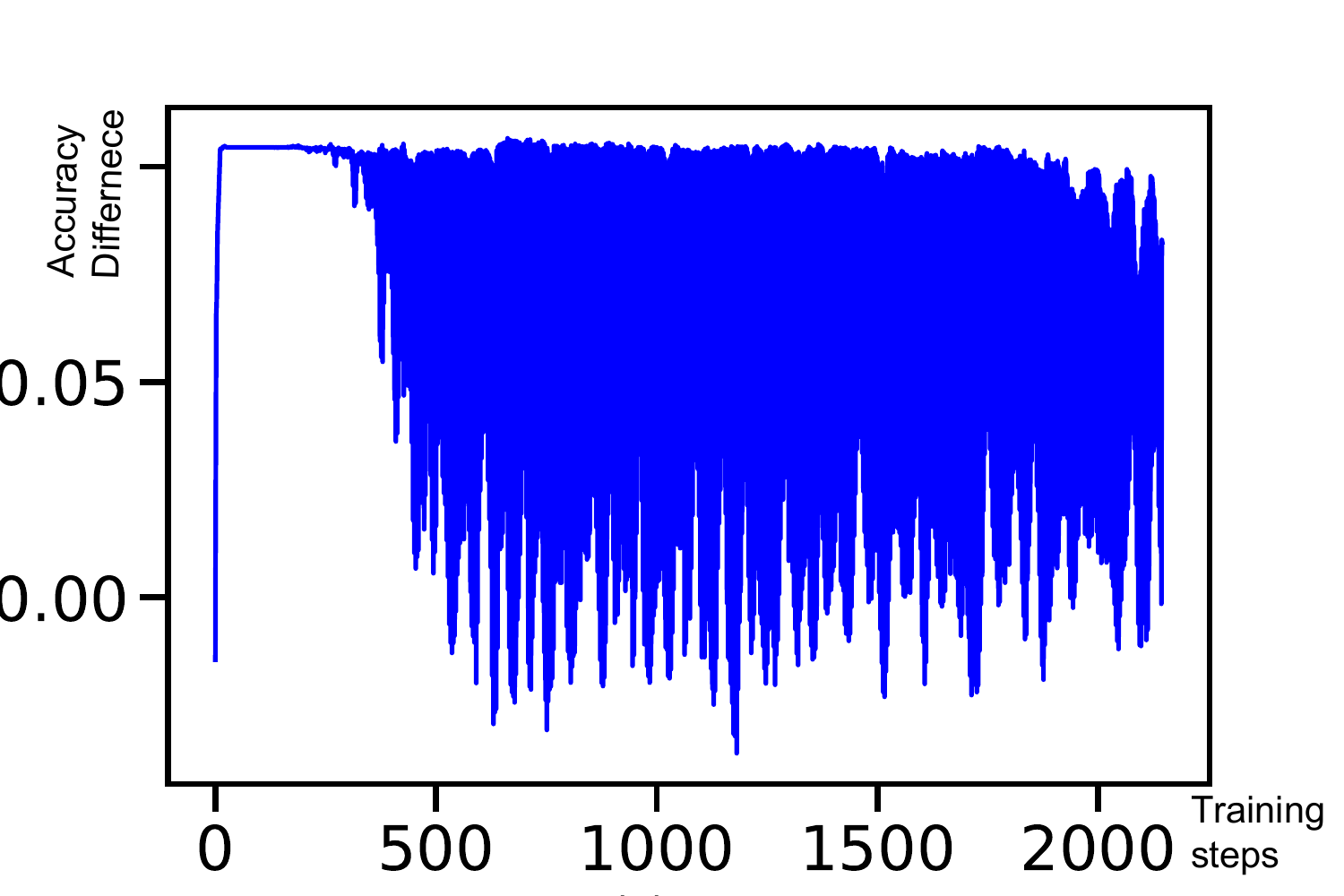}
  \caption{V-IRM Colored Desprites: Difference in accuracy of the ensemble model between the two environments (More train steps)}

\label{figs38}
\end{figure}

\begin{figure}
\centering
  \includegraphics[width=2.5in]{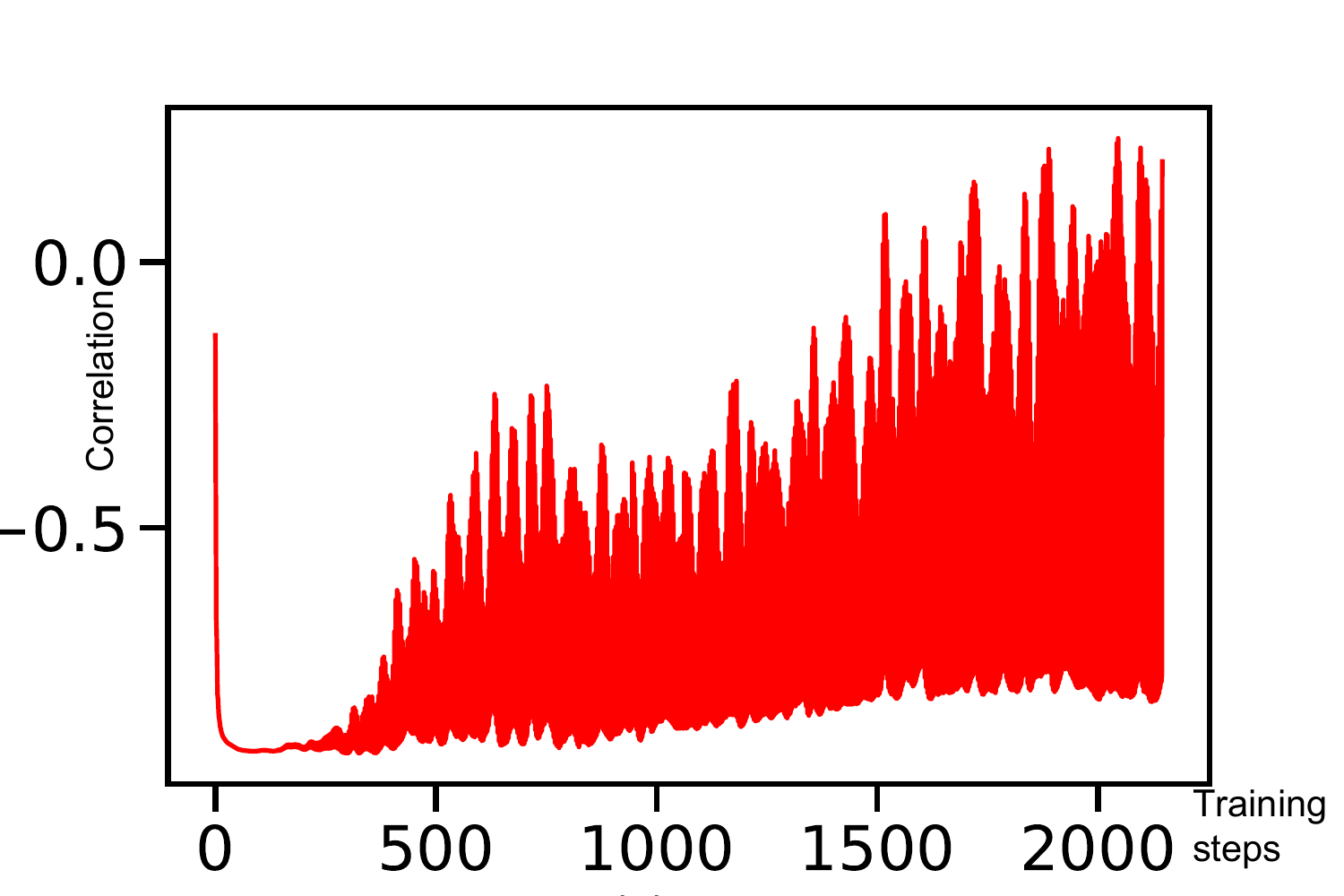}
  \caption{V-IRM Colored Desprites: Ensemble's correlation with color (More train steps)}

\label{figs39}
\end{figure}

\begin{figure}
\centering
  \includegraphics[width=2.5 in]{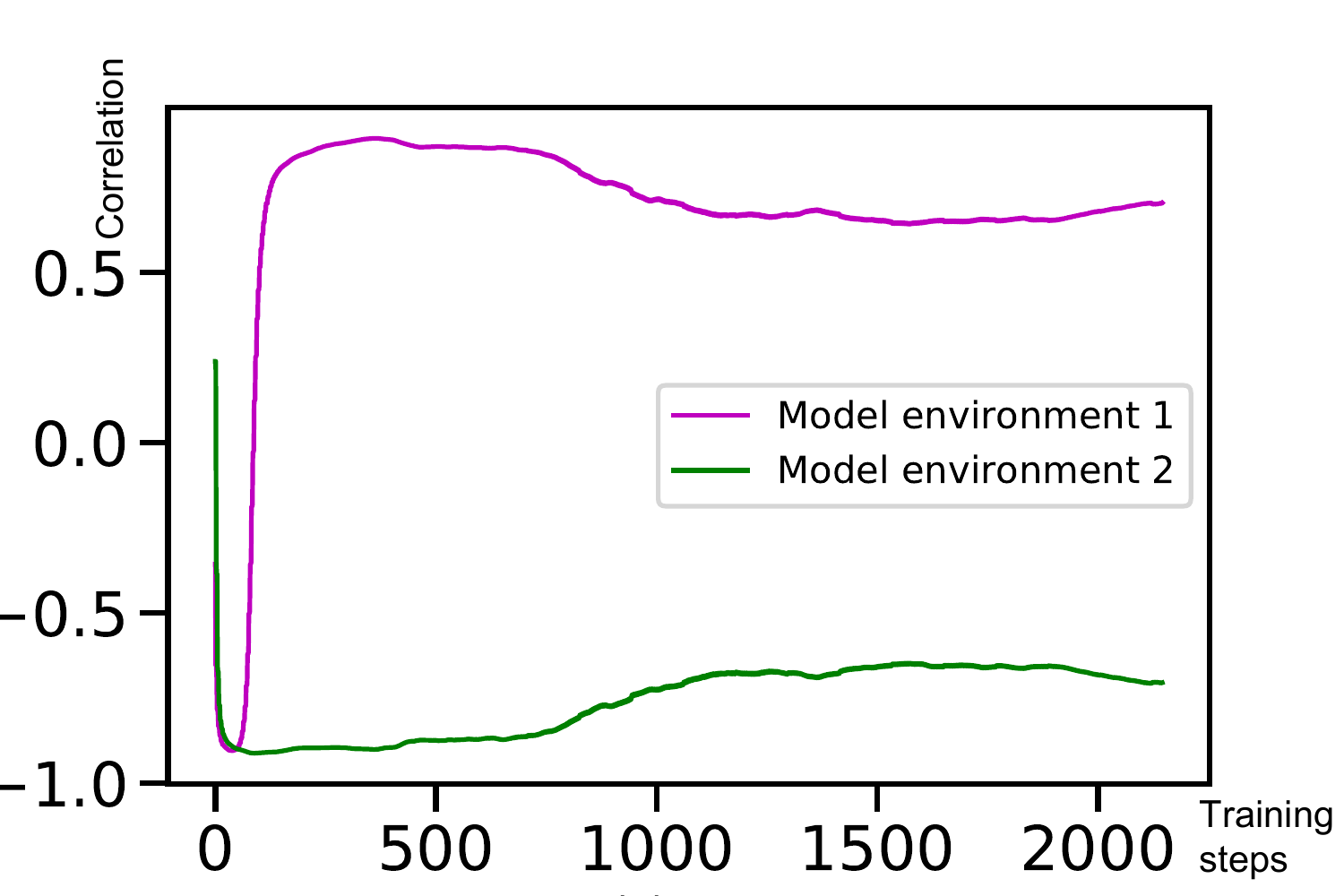}
  \caption{V-IRM Colored Desprites: Individual model correlations (More train steps)}
\label{figs40}
\end{figure}

To illustrate what happens if we let the training go on, in 
Figure \ref{figs36}-\ref{figs40} we let the training for V-IRM on Desprites dataset continue for many more training steps.  Figures \ref{figs36}-\ref{figs40} illustrate that the oscillations are stable and persist. As a result, we continue to encounter the state in which the ensemble does not exploit spurious correlations.







\clearpage
\bibliographystyle{IEEEtran}
\bibliography{ICML_IRM_game_jmtd.bib}
\end{document}